\documentclass[a4paper,fleqn]{cas-dc}

\usepackage[authoryear]{natbib}
\usepackage{amsmath}
\usepackage{amsfonts}
\usepackage{xcolor}
\usepackage{tabularx}
\usepackage{url}
\usepackage{balance}
\usepackage{float}
\usepackage{soul} 
\usepackage{xcolor,colortbl}
\usepackage{stfloats}

\definecolor{Gray}{gray}{0.85}
\newcolumntype{a}{>{\columncolor{Gray}}c}

\def\tsc#1{\csdef{#1}{\textsc{\lowercase{#1}}\xspace}}
\tsc{WGM}
\tsc{QE}

\begin{document}
\let\WriteBookmarks\relax
\def\floatpagepagefraction{1}
\def\textpagefraction{.001}

\shorttitle{}    

\shortauthors{P. Agrafiotis and B. Demir}  

\title [mode = title]{Deep Learning-based Bathymetry Retrieval without In-situ Depths using Remote Sensing Imagery and SfM-MVS DSMs with Data Gaps}

\author[1, 2]{Panagiotis Agrafiotis }
[orcid=0000-0003-4474-5007]
\cormark[1]

\ead{agrafiotis@tu-berlin.de}

\credit{Conceptualization, Data curation, Formal analysis, Funding acquisition, Methodology, Software, Validation, Visualization, Writing – original draft, Writing – review \& editing}

\affiliation[1]{organization={Faculty of Electrical Engineering and Computer Science, Technische Universit{\"a}t Berlin},
            city={Berlin},
            postcode={10587}, 
            country={Germany}}

\author[1, 2]{Beg{\"u}m Demir }
[orcid=0000-0003-2175-7072]

\ead{demir@tu-berlin.de}

\credit{Supervision, Writing – review \& editing}

\affiliation[2]{organization={BIFOLD - Berlin Institute for the Foundations of Learning and Data, Berlin, Germany},
            city={Berlin},
            postcode={10587}, 
            country={Germany}}

\cortext[1]{Corresponding author}

\begin{abstract}
Accurate, detailed, and high-frequent bathymetry is crucial for shallow seabed areas facing intense climatological and anthropogenic pressures. Current methods utilizing airborne or satellite optical imagery to derive bathymetry primarily rely on either Structure-from-Motion and Multi-View Stereo (SfM-MVS) with refraction correction or Spectrally Derived Bathymetry (SDB). However, SDB methods often require extensive manual fieldwork or costly reference data, while SfM-MVS approaches face challenges even after refraction correction. These include depth data gaps and noise in environments with homogeneous visual textures, which hinder the creation of accurate and complete Digital Surface Models (DSMs) of the seabed. To address these challenges, this work introduces a methodology that combines the high-fidelity 3D reconstruction capabilities of the SfM-MVS methods with state-of-the-art refraction correction techniques, along with the spectral analysis capabilities of a new deep learning-based method for bathymetry prediction. This integration enables a synergistic approach where SfM-MVS derived DSMs with data gaps are used as training data to generate complete bathymetric maps. In this context, we propose Swin-BathyUNet that combines U-Net with Swin Transformer self-attention layers and a cross-attention mechanism, specifically tailored for SDB. Swin-BathyUNet is designed to improve bathymetric accuracy by capturing long-range spatial relationships and can also function as a standalone solution for standard SDB with various training depth data, independent of the SfM-MVS output. Experimental results in two completely different test sites in the Mediterranean and Baltic Seas demonstrate the effectiveness of the proposed approach through extensive experiments that demonstrate improvements in bathymetric accuracy, detail, coverage, and noise reduction in the predicted DSM. The code is available at \url{https://github.com/pagraf/Swin-BathyUNet}.

\end{abstract}



\maketitle

\section{Introduction}
About 71\% of the Earth's surface is covered in water, and the oceans hold about 96.5\% of all of the Earth's water. This is about 362 million square kilometers of the total surface area; however, only a small fraction of it has been mapped by direct observation so far. This need for better knowledge of the bottom of water bodies (seabed, riverbed, lakebed) is driven by habitat destruction, marine pollution, cultural heritage at risk, recent tragedies, as well as natural disasters, navigation, and the increasing demand for offshore energy and marine resources. Accurate, detailed, and high-frequent bathymetry is crucial for the under-mapped shallow coastal areas, which are affected by intense climatological and anthropogenic pressures. On the one hand, acoustic methods exploiting echo-sounders are inefficient in shallow waters and subject to waves, reefs, and fail due to multi-path errors, while on the other hand, LiDAR (Light Detection And Ranging) is expensive, both lacking complex semantic content \citep{agrafiotis2020}. To overcome the above issues, aerial and satellite images are widely used nowadays in shallow water mapping \citep{mandlburger2022review, deepblue}. These images are processed with either Structure-from-Motion Multi-view Stereo (SfM-MVS) techniques combined with refraction correction or Spectrally Derived Bathymetry (SDB) methods and can provide a low-cost alternative to the aforementioned acoustic and LiDAR methods. Both have their application areas in seabed mapping studies, and choosing between them depends on the scale of the area to be surveyed, the required resolution of the data, budget constraints, the availability of reference bathymetric data and the intended use of the results.

\subsection{SfM-MVS with refraction correction}
\label{Bathymetric SfM-MVS with Refraction Correction}
SfM is a 3D reconstruction technique used to create 3D structures from overlapping 2D image sequences, which can vary in scales and should be captured from different angles. In the context of bathymetry, SfM combined with MVS is often used to create 3D models of shallow underwater environments based on overlapping airborne or satellite imagery \citep{agrafiotis2019, cao2019shallow, agrafiotis2021, mulsow2020comparison,  lovaas2023coregistration, cao2024making}. Although in terrestrial environments, SfM-MVS reconstruction and texture mapping pipelines can be executed operationally, this does not apply to the marine environment and, especially, for 3D mapping of the seabed. The optical properties of water severely affect this process, with water refraction being the main factor that affects the geometry of remote sensing imagery and consequently of the derived bathymetry, leading to an underestimation of depths up to 40\% of the real depth in the derived Digital Surface Model (DSM) \citep{agrafiotis2020}. More specifically, the amount of the refraction of a light beam is affected by the amount of water that covers the point of origin situated on the bottom and the angle of incidence of the beam in the water/air interface. As shown in \cite{skarlatos2018, agrafiotis2021}, the water depth to flying height ratio is irrelevant in air/ satellite imaging cases, and water refraction correction is always necessary while a common (self-) calibration approach of the sensor fails to model and describe the effects of refraction in the images. Hence, much more sophisticated solutions have been developed to correct for the refraction effects in these cases \citep{dietrich2017, mandlburger2019through, cao2020universal, agrafiotis2021}. When refraction effects are eliminated, SfM-MVS methods can achieve high-resolution and high-accuracy results. However, they depend heavily on the quality of the image data and the complexity of the underwater environment.

In addition to refraction, SfM-MVS-derived bathymetry, like any method for bottom mapping using remote sensing imagery, is influenced by factors related to sea surface and water column conditions such as waves, visibility, turbidity, sun glint, attenuation, scattering in the water column, and, in the case of low-altitude imagery, caustics \citep{agrafiotis2023seafloor}. These challenges can be mitigated through careful planning of data acquisition. However, the effectiveness of SfM-MVS methods, especially through water, also depends on the presence of texture or random patterns on the seafloor to ensure accurate point detection and dense matching. The combination of poor texture and refraction effects violates the epipolar geometry of the stereo pairs: the stereo matching between two images compares the wrong candidate pixels, causing inconsistencies and noise in the pairwise disparity images and depth maps. The impact of inconsistencies in MVS is influenced by the smoothness constraint in semi-global matching \citep{mvs} and the robustness of outlier rejection within the MVS pipeline. In regions with high inconsistency, such as those with poor texture (covered with sand and seagrass), this results in lack of coverage, especially when the rejection of outliers is particularly stringent \citep{lovaas2023coregistration}. Consequently, SfM-MVS-based bathymetry consistently struggles in environments with homogeneous visual textures, such as sandy or heavily seagrass-covered bottoms, environments characterized by homogeneous visual textures, delivering incomplete and noisy bathymetry.

\subsection{Spectrally derived bathymetry}
SDB methods, applied in satellite or aerial imagery, do not depend on the texture of the seabed. Depending on the algorithm applied, a homogeneous seabed may lead to more accurate results, also covering vast areas where high-resolution detail is less critical. These methods are based on the attenuation of radiance as a function of depth and wavelength in the water and can deliver depths over huge shallow areas. In addition to well-established empirical models \citep{lyzenga1978passive, stumpf2003determination} and physics-based and simple regression methods such as random forests and support vector machines \citep{niroumand2020smart, eugenio2021high, thomas2022purely, mudiyanselage2022satellite}, deep learning-based approaches have recently gained great attention in SDB \citep{ai2020convolutional,kaloop2021hybrid,lumban2022extracting,mandlburger2021bathynet,xi2023band, shen2023shallow, zhou2023bathymetry, al2023satellite, magicbathynet, gupta2024improving, musrfm}.
This increased interest stems from their ability to better manage the complex interaction of light with the surface of the water, the water column, and the seabed compared to simpler models \citep{mandlburger2022review}, particularly in complex environments, such as seabeds with varying habitats. 

In comparison to physics-based or SfM-MVS-based approaches, regression- or deep learning-based methods emerge as the most efficient for real-world bathymetric applications. However, most deep learning models currently employed in bathymetry are based on architectures that are not efficient in capturing long-range dependencies, which are inherently beneficial for interpreting the complex and varied characteristics of aquatic landscapes. These models also struggle to adequately capture multi-scale context. Additionally, the effectiveness of these methods is heavily dependent on access to high-quality, correctly georeferenced, and dense in situ bathymetric data for model training, a resource often not available in many operational bathymetric applications. This limitation highlights the need for innovative architectures and methodologies that can address these challenges while expanding the applicability of deep learning in bathymetric mapping.

\subsection{Integrated SfM-MVS and SDB methods}
\label{Integrated}
In the existing literature, there is a clear shortage of studies that integrate photogrammetric and spectral-based techniques for image-based bathymetry mapping, although they are highly complementary \citep{slocum2020combined}. In one of the first approaches towards that, \cite{slocum2020combined} proposed a bathymetry mapping method which combines a refraction-corrected SfM-MVS workflow with spectrally based depth-retrieval algorithms. The complexity and computational demands of this method are considerable, since it does not rely on orthoimages but instead uses all available imagery, requiring the merging of spectrally derived bathymetric data from each aligned image into a single DSM. Furthermore, this method is sensitive to precise camera orientations, as is the refraction correction method used. The test site used was only with water depths up to 4m. In \cite{he2022fully}, authors integrated a photogrammetric and an SDB method, employing a shallow neural network on orthoimages. The maximum depth of the test site used exceeded 16m; however, the missing areas in the DSM represented only a minimal part of the seabed, presenting no challenge. To correct for the refraction effect, they multiplied the apparent SfM-MVS depths by 1.34, the refraction index of fresh water. However, as shown in \cite{agrafiotisphd}, the use of this form of correction is acceptable only in very shallow waters and yields remarkable errors after 2 to 3m depth; In 1m depth, expected errors in depth determination are in the scale of 10\%-20\% of the real depth. Finally, the authors in \cite{cao2024making} exploited satellite photogrammetric depth data derived from WorldView-2 (WV2) satellite stereo images to calibrate an empirical bathymetric model. The results on 3 different sites showcase the feasibility of the method. However, a simple logarithmic band ratio model \citep{stumpf2003determination} was used failing to address the noise inherited by the photogrammetric process. The rest of the proposed methods exhibit methodological flaws, such as overlooking crucial steps like refraction correction, and thus they are not reported here.

The integration of photogrammetric and spectrally derived techniques for bathymetry mapping presents significant promise but also faces several key challenges. These challenges primarily revolve around computational complexity, reliance on precise camera orientations, sensitivity to environmental factors such as water clarity, and attempts to fill only very small voids in DSMs and very shallow depths. Additionally, issues with remaining noise in DSMs and limitations in the refraction correction methods used complicate the process. A significant difficulty also arises in using deep learning models that fail to capture long-range spatial dependencies and multi-scale context, which are essential for accurately modeling complex underwater and seabed environments. Addressing these limitations will be crucial for the development of robust and accurate methodologies for shallow-water mapping, enabling more effective management and conservation of marine ecosystems.

\subsection{Challenges and contribution}
In this work, we merge the advantages of the SfM-MVS and SDB methods to address their individual drawbacks. As already reported, bathymetric SfM-MVS struggles with missing bathymetric data and noisy depths in uniform seabed regions, while SDB is hampered by the need for accurate reference bathymetric data for model training. Our approach seeks to address these issues by minimizing their weaknesses and providing improved, complete, and precise bathymetry. To achieve this, we propose a deep learning-driven approach where the incomplete DSM, derived from SfM-MVS and corrected for refraction effects, is utilized alongside co-registered orthoimagery to train the proposed SDB model. Existing learning-based SDB approaches rely on training data that cover the entire area or a wide range of seabed classes and depths, typically with even distribution. However, the challenge here lies in obtaining learning-based depth information for orthoimage pixels of homogeneous areas that lack depth annotations or have only sparse depth annotations in the training data, as these areas are often overlooked by the SfM-MVS depth extraction process. Thus, our approach must contend with numerous unnecessary depth labels where bathymetric data already exists from the SfM-MVS, along with sparse depth labels available for the missing areas. The main contributions of this work are summarized as follows:
\begin{itemize}
    \item We propose a novel approach in which the DSM derived from SfM-MVS, including data gaps, is corrected for refraction effects and used as input for a deep learning-based bathymetry prediction model, along with the co-registered orthoimagery. This approach leverages orthoimages generated from refraction-corrected DSMs to eliminate horizontal errors \citep{agrafiotis2020}.
    \item We employ a state-of-the-art refraction correction method to improve the accuracy of SfM-MVS training data, which outperforms previously applied methods.
    \item We introduce Swin-BathyUNet, a deep learning network designed for SDB that combines U-Net \citep{unet} with Swin Transformer \citep{swin} self-attention layers and cross-attention mechanisms to effectively capture long-range spatial dependencies and multi-scale context. These dependencies are crucial for overcoming the unique challenges posed in a wide range of bathymetric mapping scenarios, where reference data may be fully available or limited. Swin-BathyUNet can also function as a standalone solution for standard bathymetric mapping with various training depth data, independent of SfM-MVS output. To deal with the numerous unnecessary depth labels where bathymetric data from SfM-MVS already exist, a boundary-sensitive weighted Root Mean Square Error (RMSE) loss function is introduced.
    \item We provide extensive experimental results and an ablation study of the introduced learning-based SDB network, demonstrating its superiority over the baseline in terms of bathymetric accuracy, noise reduction, and coverage, as well as significant improvements compared to the original SfM-MVS DSM.
\end{itemize}

The rest of the article is organized as follows. Section \ref{section:Methodology} presents the proposed methodology for the prediction of shallow water bathymetry with a self- and cross-attention network and refraction corrected SfM-MVS reference data, while Section \ref{section:Data} presents and analyzes the exploited datasets and provides details on the experimental setup. Section \ref{section:Results} presents and analyzes the experimental results while Section \ref{section:Discussion} discusses the achievements and limitations of the proposed approach. Section \ref{section:Conclusion} concludes the article.

\section{Proposed Methodology}
\label{section:Methodology}
\begin{figure*}[h!]
\centering
\includegraphics[width=17cm]{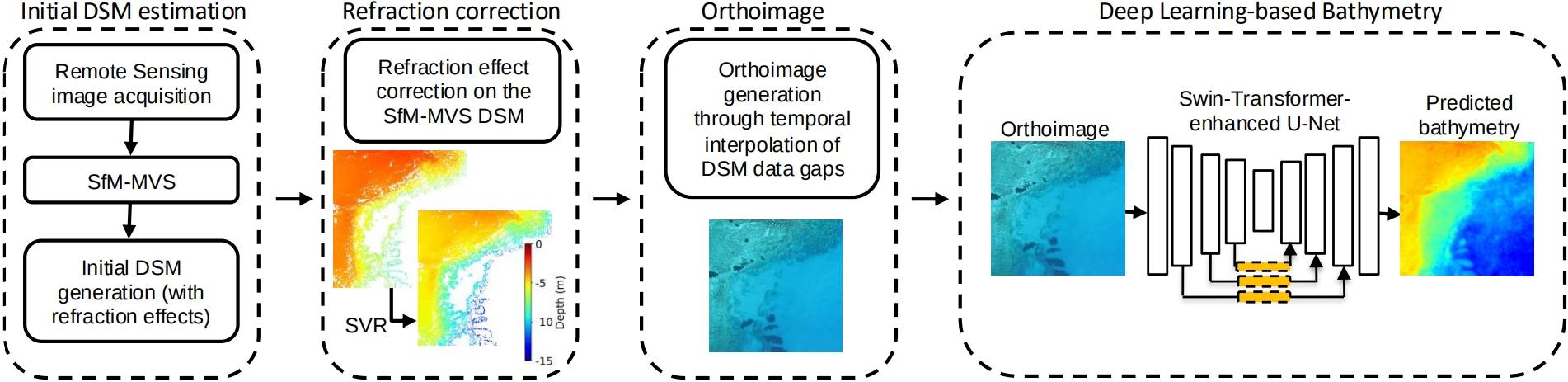}

\caption{The four major pillars of the proposed methodology.}
\label{methodology}
\vspace{-0.15in}
\end{figure*}

The proposed methodology incorporates several key pillars, as illustrated in Figure \ref{methodology}. First, SfM-MVS using the refracted imagery is performed, facilitating the initial (refracted) generation of DSM. Secondly, the initial DSM is corrected for the refraction effects using a state-of-the-art linear support vector regression (SVR) model, and subsequently the orthoimage of the area is generated using the corrected DSM. Third, the final bathymetry is predicted by the Swin-BathyUNet, leveraging pairs of RGB orthoimage patches and co-registered refraction corrected SfM-MVS-derived DSMs with data gaps. Specific information on those steps is provided below.

\subsection{SfM-MVS and initial DSM generation}
Following the remote sensing image data collection and ground control point (GCP) measurements, an initial SfM-MVS is executed ignoring the refraction effects on the images. In the SfM step, the relative 3D positions and orientations of the cameras are estimated, along with the 3D coordinates of distinctive features in the images, known as keypoints. These keypoints are tracked across multiple images to establish correspondences, which are then used to calculate the camera poses and the 3D structure of the scene (sparse 3D point cloud). SfM outputs are georeferenced using the GCPs. Once the camera poses and the sparse 3D point cloud are obtained through SfM, MVS densely reconstructs the scene geometry. MVS algorithms \citep{mvs} estimate the depth of every pixel in the images by analyzing the photometric and geometric consistency in multiple views. This dense depth information can be further refined to generate high-quality textured 3D models and DSMs. For the purpose of this work, we performed SfM-MVS using the commercial photogrammetry software Agisoft Metashape. However, applying SfM-MVS on a submerged scene imaged from air- or satellite-borne platforms always results in a 3D dense point cloud and consequently a DSM with apparent water depths. Therefore, each depth must be consequently corrected for the refraction effects.

\subsection{Refraction correction in SfM-MVS}
\subsubsection{Water refraction principles}
Imagery captured over the water that depicts the seafloor, commonly referred to as through-water imagery, is primarily affected by geometric distortions caused by refraction, as described by Snell's Law. The most well-known form of this law for light traveling through different media is given by Equation \ref{eq:equation2.1}:

\begin{equation}
\mathrm{n}_{1} \sin \left(\mathrm{a}_{1}\right)=\mathrm{n}_{2} \sin 
\label{eq:equation2.1}
\end{equation}

\begin{figure}[h]
\centering
\includegraphics[width=7.5cm]{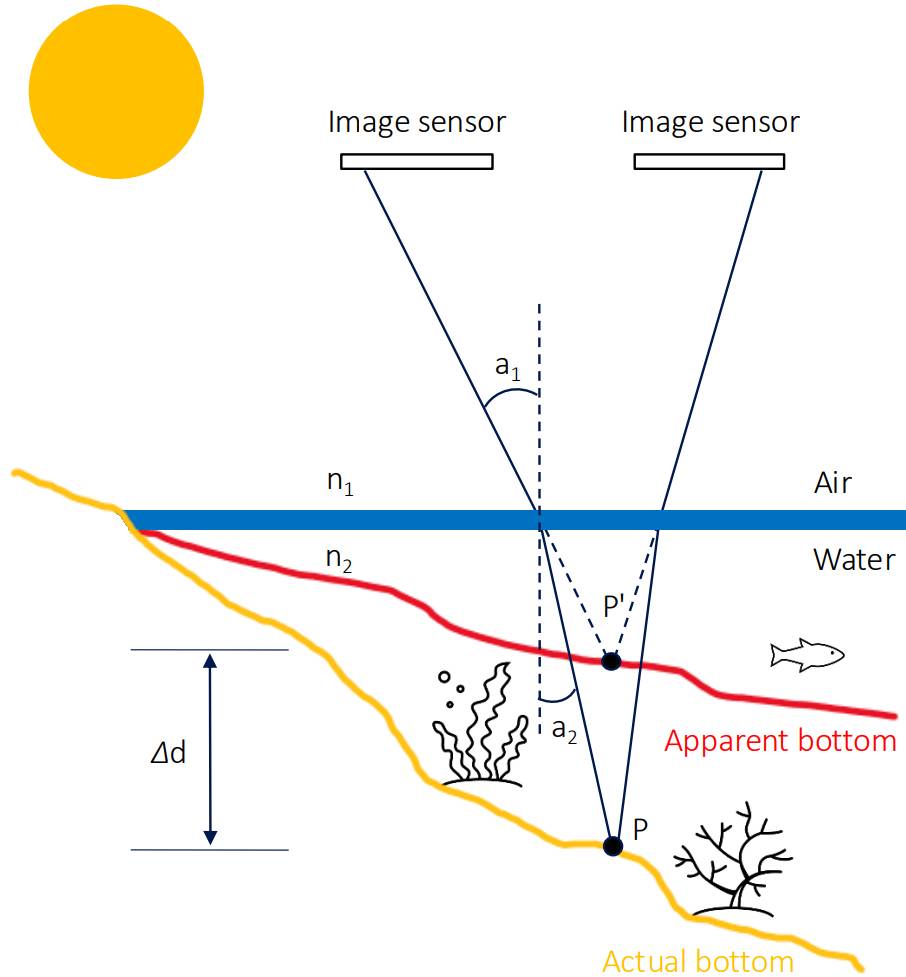}
\caption{Water refraction effects on the geometry of the apparent bottom DSM generation using SfM-MVS techniques.}
\label{refr}
\end{figure}

This equation is employed to describe the connection between the angles of incidence and refraction, particularly concerning the passage of light or other waves through a boundary separating two distinct isotropic mediums, such as water, glass, or air. 
Snell's Law states that the ratio of the sines of the angles of incidence ($\mathrm{a}_{1}$) and refraction ($\mathrm{a}_{2}$), shown also in Figure \ref{refr}, is equivalent to the ratio of phase velocities in the two media, or equivalent to the reciprocal of the ratio of the indices of refraction ($n_1$ and $n_2$ respectively, where $n_1 > n_2$). The law follows from Fermat's principle of least time, which in turn follows from the propagation of light as waves.

The challenge intensifies when stereo view or multiple view geometry is employed. In Figure \ref{refr}, the $XY$ plane of the world coordinate system $X, Y, Z$ is defined as the boundary plane between air and water (blue horizontal line) and the $Z$ axis is perpendicular to this plane with its positive direction upwards. If water was not present, the depth of a point $P$ on the bottom would be calculated by SfM-MVS correctly. However, when water is present, due to the refraction effect, the point's $P$ depth is calculated in it's apparent position $P'$ lying always in a shallower depth than the real one. The same applies for all the points of a 3D point cloud which are situated under the surface of the water and consequently all the submerged part of the generated DSM. Consequently, the main aim of the refraction correction method to be used here is to compensate for the systematic underestimation of depths in the DSM of the bottom by learning the correlation of the apparent depth $Z{_0}$ of a point in $P'$ with its ground truth depth $Z$ in it's true position $P$ on the bottom.

\subsubsection{Refraction correction approach}
In order to correct for the water refraction in the generated SfM-MVS-based DSM, we exploit a state-of-the-art methodology based on a linear SVR model, able to estimate the correct depth in shallow waters by providing only the apparent one \citep{agrafiotis2019, agrafiotis2021}. As shown in \cite{agrafiotis2021}, the method used outperforms those previously applied in the integrated methods discussed in Section \ref{Integrated}. The true and apparent depths have been shown to exhibit a clear linear relationship in SfM-MVS cases \citep{agrafiotis2021}, making a linear SVR model sufficient. Considering the problem of approximating the set of depths:
\begin{equation}
D=\left\{\left(Z_{0}^{1}, Z^{1}\right), \ldots,\left(Z_{0}^{l}, Z^{l}\right)\right\} \quad Z_{0} \in R^{n}, \quad Z \in R
\label{eq:equation4.1}
\end{equation}
with a linear function
\begin{equation}
f\left(Z_{0}\right)=\left\langle w, Z_{0}\right\rangle+b
\label{eq:equation4.2}
\end{equation}

the work presented in \citep{agrafiotis2019, agrafiotis2021} formulates this function using an SVR model.  There, the predictions for new apparent depths $Z_{0}$ can be made using:

\begin{equation}
f\left(Z_{0}\right)=\sum_{n=1}^{N}\left(a_{n}+\hat{a}_{n}\right) k\left(Z_{0}, Z_{0 \mathrm{n}}\right)+b
\label{eq:equation4.5}
\end{equation}

which is expressed in terms of the kernel function \textit{k}. The support vectors refer to the data points that contribute to predictions according to Equation \ref{eq:equation4.5}, while the parameter $b$ can be determined using a data point for which $0 < a_n, \hat{a}_{n}  <C$ \citep{87}. 

\subsection{DSM update and orthoimage generation}
By applying the SVR model described to the refracted DSM, the refraction-free DSM is obtained. To generate orthoimages of the study areas and ensure comprehensive coverage, we adopt a systematic approach to address the no-data areas in the updated DSM using Agisoft Metashape. These data gaps, typically arising from unextracted 3D point cloud due to matching failures in the SfM-MVS process, particularly in homogeneous regions like sand or seagrass, undergo interpolation to ensure seamless integration, albeit without the incorporation of actual depth data. This temporary gap filling will not result in horizontal errors in the orthoimages, as the uniformity of the rectified/ reprojected pixels depicting mostly homogeneous sandy and seagrass-covered areas, eliminates potential visual discrepancies \citep{agrafiotis2020}. Once the updated DSM is created, the texture from the original imagery is projected onto the refraction corrected DSM. To create the orthoimage, the textured DSM is reprojected onto a flat plane, rectifying distortions and ensuring that all pixels have uniform scale and true geographic coordinates, resulting in a geometrically accurate, map-like image, free from perspective distortions. 

\subsection{Learning-based bathymetry}
\subsubsection{SDB principles}
\begin{figure}[h]
\centering
\includegraphics[width=7.5cm]{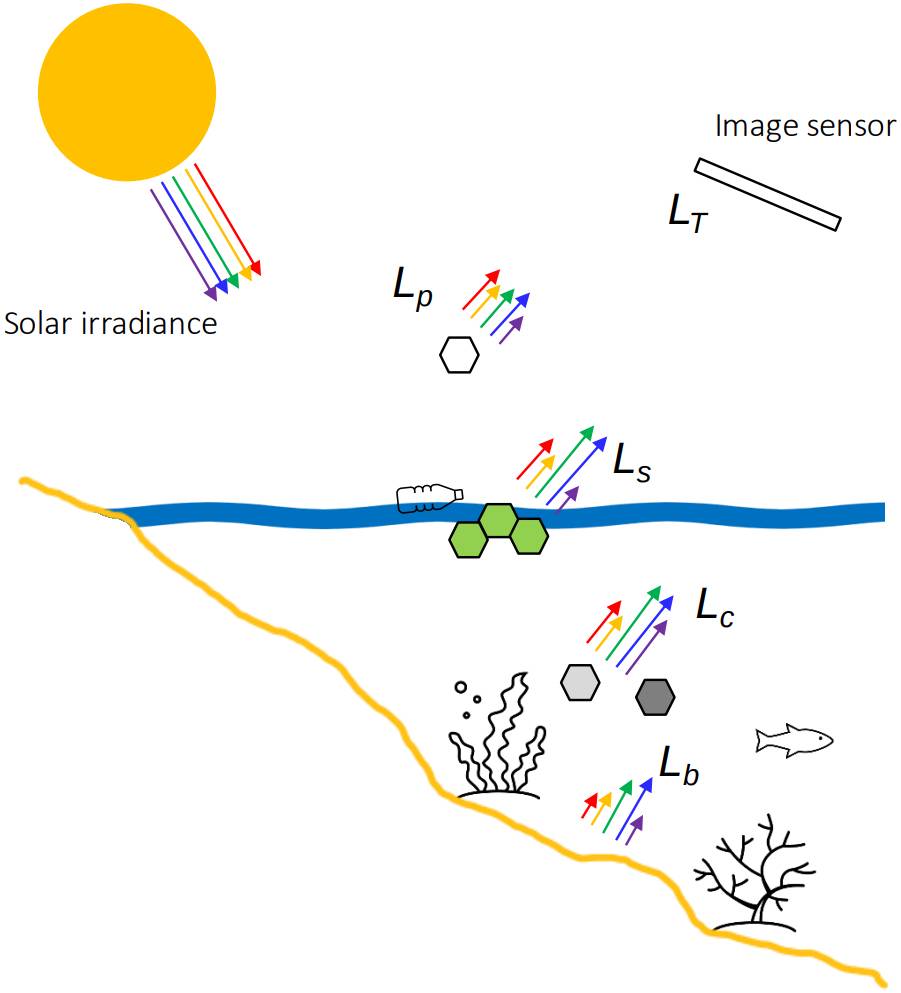}
\caption{Spectrally Derived Bathymetry. With the sun as an illumination source, the formed image content comprises back-scatter components from atmosphere, water surface, water column and water bottom.}
\label{sdb}
\end{figure}

In this subsection, the basics of SDB are given. In this context, the main goal is to model the relationship between the radiometry of the imagery used and the reference water depths. To undertake this task, a deep understanding of how solar radiation interacts with different elements such as the atmosphere, water surface, water column, and even the bottom of the water body (Figure \ref{sdb}) should be achieved. This understanding should encompass how these interactions vary depending on the wavelength of the radiation.

The total up-welling radiance $L_T(\lambda)$ possibly captured by the image sensor can be expressed as the sum of individual partial contributions \citep{legleiter2009spectrally}:

\begin{equation}
L_T(\lambda)=L_p(\lambda)+L_s(\lambda)+L_c(\lambda)+L_b(\lambda)
\label{lt}
\end{equation}

where $L_p(\lambda)$ are the contributions from the atmosphere, $L_s(\lambda)$ is the radiance reflected from the water surface, $L_c(\lambda)$ is the radiance from the water column, and $L_b(\lambda)$ is the bottom-reflected radiance. $L_s(\lambda)$ depends on the roughness of the water surface and sun position (sun glint) and $L_c(\lambda)$ is related to the water’s optical properties  and contributing factors are absorption and scattering by pure water, but also turbidity caused by suspended sediment and organic matter.

By solving Equation \ref{lt} for $L_b(\lambda)$ we obtain the radiance reflected from the water bottom, as expressed in Equation \ref{lb}: 

\begin{equation}
L_b(\lambda)=L_T(\lambda)-L_p(\lambda)-L_s(\lambda)-L_c(\lambda)
\label{lb}
\end{equation}

This radiance serves as the basis for modeling the color-to-depth relationship by using independent reference data to train either a simple or complex model.

\subsubsection{Proposed learning-based bathymetry network}
The proposed Swin-BathyUNet architecture builds on the conventional U-Net \citep{unet}, which is the most widely used model in tasks related to ocean remote sensing \citep{deepblue}, known for its encoder-decoder structure. Although continuous enhancements have been made to the U-Net model to improve accuracy in the context of ocean mapping \citep{deepblue} and image-based bathymetry specifically \citep{mandlburger2021bathynet, magicbathynet}, the skip-connection scheme still struggles with capturing the multi-scale context, limiting it's performance \citep{ates2023dual}. However, improvements to the skip-connection mechanism to address this limitation, particularly in the context of ocean mapping and image-based bathymetry, have not yet been explored.

\begin{figure*}[h!]
\centering
\includegraphics[width=17.4cm]{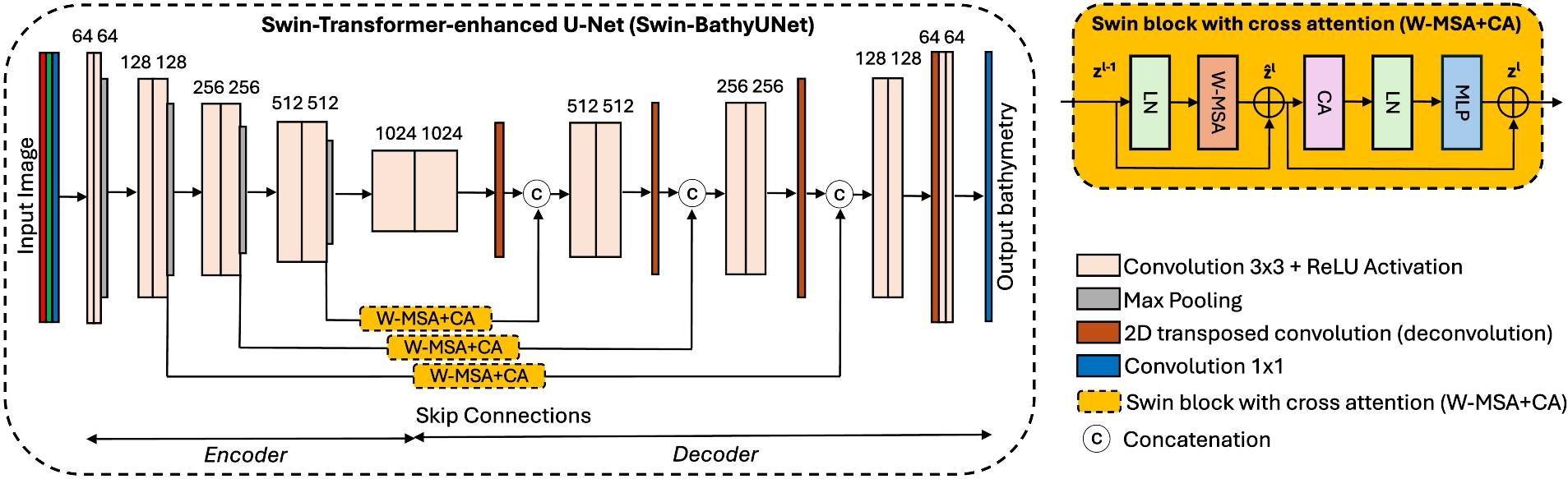}
\vspace{-0.1in}
\caption{The proposed Swin-BathyUNet architecture and a Swin Transformer Block. W-MSA is the multi-head self attention module with regular windowing configurations. Notation presented with Equation \ref{equation_swin}.}
\vspace{-0.1in}
\label{unetswin}
\end{figure*}

To model the color-to-depth relationship and address this issue, we extend the U-Net architecture by integrating self- and cross-attention mechanisms through three Swin Transformer \citep{swin} blocks. This integration includes both window-based self-attention and cross-attention mechanisms (Figure \ref{unetswin}). Unlike typical U-shaped networks that use Transformer-based attention \citep{vit}, where attention is either applied within the encoder \citep{cao2022swin} or a Transformer is integrated into the U-shaped architecture \citep{chen2021transunet}, we aim to combine the detailed high-resolution spatial information from CNN features with the global context provided by Transformers in a more lightweight design. To achieve this, we enhance the skip connections and address the original U-Net's limitations in capturing long-range and multi-scale dependencies. The architecture consists of an encoder, a center block, and a decoder with skip connections, enriched with Swin blocks at various stages to improve the model's ability to learn complex features across different scales. The main components of the network are described as follows:

\textbf{Encoder blocks}: Four convolutional blocks are defined to progressively extract features from the input images. Each block utilizes a series of two convolutional layers followed by ReLU activation functions. In each block, the number of filters doubles with each successive layer: from 64 to 128, 128 to 256, and 256 to 512 \citep{unet}.

\textbf{Swin Transformer blocks:} The encoder outputs are then combined with the outputs from the Swin Transformer blocks with cross-attention mechanisms, specifically designed to operate on the encoder’s outputs. The first Swin block processes the 512 input channels from the fourth encoder layer in an embedding dimension of 512, handling a spatial dimension of $\frac{H}{16} \times \frac{W}{16}$. The second block operates on 256 input channels from the third encoder layer, projecting them to an embedding dimension of 256 with a spatial dimension of $\frac{H}{8} \times \frac{W}{8}$. Finally, the third block takes 128 input channels from the second encoder layer, maintaining an embedding dimension of 128 and a spatial dimension of $\frac{H}{4} \times \frac{W}{4}$. The outputs from the Swin blocks are interpolated to match the decoder size and then concatenated with the decoder blocks. The Swin Transformer block \citep{swin} consists of a window self-attention layer followed by a cross attention layer and a multilayer perceptron (MLP) with ReLU non-linearity. The input is first normalized using Layer Normalization (LN). The block employs a window-based attention mechanism, wherein the input tensor is partitioned into non-overlapping windows, allowing for local self-attention computations. The cross-attention (CA) layer enables interactions between multiple input sequences. Finally, the output undergoes a second normalization and is processed through an MLP, which employs residual connections to combine outputs. The Swin Transformer blocks are computed as:

\begin{equation}
\begin{aligned}
\hat{z}^l &= \text{W-MSA}(\text{LN}(z^{l-1})) + z^{l-1},\\
z^l &= \text{MLP}(\text{LN}(\text{CA}(\hat{z}^l))) + \hat{z}^l
\end{aligned}
\label{equation_swin}
\end{equation}

where $\hat{z}^l$ and $z^l$ denote the output features of the W-MSA module and the CA and MLP module for block $l$, respectively; W-MSA denotes window based multi-head self-attention using regular window partitioning configuration. The MLP within the Swin Transformer block is defined as:

\begin{equation}
\text{MLP}(X) = W_2 \cdot \text{ReLU}(W_1 \cdot X + b_1) + b_2
\end{equation}

where, \( W_1 \in \mathbb{R}^{D \times D_{\text{hidden}}} \) is the weight matrix for the first linear transformation, \( W_2 \in \mathbb{R}^{D_{\text{hidden}} \times D} \) is the weight matrix for the second linear transformation, \( b_1 \in \mathbb{R}^{D_{\text{hidden}}} \) and \( b_2 \in \mathbb{R}^{D} \) are the bias terms,  \( D \) is the input feature dimension, \( D_{\text{hidden}} = 4D \) is the dimensionality of the hidden layer. Here, \( X \) is the input tensor, and after the first linear transformation \( W_1 \cdot X + b_1 \), the output is passed through the ReLU activation function. The resulting tensor is then transformed by \( W_2 \) to produce the final output of the MLP.

The self-attention mechanism computes the relationships between queries, keys, and values within each window. The query, key, and value matrices are calculated as:

\begin{equation}
Q = X_{\text{input}} W_Q, \quad K = X_{\text{input}} W_K, \quad V = X_{\text{input}} W_V
\end{equation}

where \( W_Q, W_K, W_V \in \mathbb{R}^{D \times D} \) are the learned weight matrices. The attention scores determine the importance of different parts of the input within each window and are computed as:

\begin{equation}
\text{Attention}(Q, K, V) = \text{Softmax}\left(\frac{Q K^\top}{\sqrt{d_k}}\right) V
\end{equation}

with \( d_k = \frac{D}{N_\text{heads}} \) being the dimensionality of each head's output, serving as a scaling factor to stabilize the gradients during training. The output after applying the attention mechanism with Layer Normalization is given by:

\begin{equation}
X' = X_{\text{input}} + \text{Attention}(\text{LN}(X_{\text{input}}))
\end{equation}

The final output after passing through the MLP is computed as:

\begin{equation}
X'' = X' + \text{MLP}(\text{LN}(X'))
\end{equation}

In addition to self-attention, the network integrates a cross-attention mechanism to capture interactions between multi-level features and refine the upsampling process in the decoder. Let \( X_{\text{low}} \in \mathbb{R}^{B \times N_{\text{low}} \times D} \) represent low-level features from earlier layers, and \( X_{\text{high}} \in \mathbb{R}^{B \times N_{\text{high}} \times D} \) represent high-level features from deeper layers. The cross-attention mechanism is formulated as follows:

\begin{equation}
\begin{aligned}
Q_{\text{high}} &= X_{\text{high}} W_{Q_{\text{high}}} \\
K_{\text{low}} &= X_{\text{low}} W_{K_{\text{low}}} \\
V_{\text{low}} &= X_{\text{low}} W_{V_{\text{low}}}
\end{aligned}
\end{equation}

where \( W_{Q_{\text{high}}}, W_{K_{\text{low}}}, W_{V_{\text{low}}} \in \mathbb{R}^{D \times D} \) are the learned weight matrices. The cross-attention (CA) output is computed as:

\begin{equation}
\text{CA}(Q_{\text{high}}, K_{\text{low}}, V_{\text{low}}) =
\text{Softmax}\left(\frac{Q_{\text{high}} K_{\text{low}}^\top}{\sqrt{d_k}}\right) V_{\text{low}}
\end{equation}

with \( d_k = \frac{D}{N_\text{heads}} \) serving as the scaling factor.
\\

\textbf{Center block}: It takes 512 channels as input from the last encoder layer and increases the output to 1024 channels through two sequential convolutional layers, each using a kernel size of 3 and a padding of 1.

\textbf{Decoder blocks}: The decoder consists of four upsampling stages, each designed to reconstruct the spatial resolution of the feature maps. The first upsampling stage takes 1024 input channels from the center block and outputs 512 channels, while the final upsampling stage takes 128 channels and outputs 64 channels. For each stage, the outputs from the Swin blocks are concatenated with the decoder blocks. Finally, the network concludes with a single convolutional layer (1x1 convolution) that compresses the 64 channels to a single output channel, representing the corresponding bathymetry map.

\subsubsection{Boundary-sensitive weighted RMSE loss} 
We have created a custom boundary-sensitive weighted (BSW) RMSE loss function that incorporates spatial information through the use of a distance transform. The process starts by calculating the Euclidean distance transform (EDT) for each batch element's mask, measuring the distance of each pixel to the nearest zero pixel, which represents the missing areas of the SfM-MVS DSM. The distances are then clipped to lie within a specified range, defined by a minimum and maximum distance. This clipping ensures that the distances influencing the loss function are controlled and bounded, preventing extreme values from disproportionately affecting the model's learning process. We then compute weights based on these distances using either a linear or exponential decay function, where the weights inversely relate to the distance. Pixels closer to the areas of interest get higher weights, helping the model focus on learning from the more critical regions. After applying these weights to the loss calculation, we normalize the result by the number of relevant pixels to ensure that the model learns in a balanced way. Given the output of the neural network $Z \in \mathbb{R}^N$, the ground truth depth $Z_0 \in \mathbb{R}^N$, and a binary mask $m \in \{0, 1\}^N$, we define the following components:

For each element in the batch, we compute the Euclidean distance transform of a binary mask, which is defined as:

\begin{equation}
\text{D}(m)_i = \sqrt{\sum_{n} (x[i] - m[i])^2} 
\label{loss1}
\end{equation}

for $\text{m}(i) = 1$, where \( m[i] \) is the mask point with the smallest Euclidean distance to input points \( x[i] \), and \( n \) is the number of dimensions. Distances are clipped to be within the range $[D_{min}, D_{max}]$. Based on the decay type; linear or exponential, we compute the weights \( w \) as follows:

For linear decay:
\begin{equation}
w = 1 - \frac{D_{\text{clipped}} - D_{\text{min}}}{D_{\text{max}} - D_{\text{min}}}
\label{loss2}
\end{equation}   

For exponential decay:
\begin{equation}
w = \exp\left( - \frac{D_{\text{clipped}} - D_{\text{min}}}{D_{\text{max}} - D_{\text{min}}} \right)
\label{loss3}
\end{equation}  

where $\text{D}_{clipped}$ is the calculated clipped distance. Consequently we apply the mask to the weights. Then we calculate the Mean Squared Error (MSE) loss and apply the masked weights to the loss and normalize. The final boundary-sensitive weighted Root Mean Squared Error (RMSE) loss function is defined as follows:

\begin{equation}
\mathcal{L}_{\text{BSW}} = \sqrt{ \frac{\sum_{i=1}^{N} \text{w}_i \cdot (\text{Z}_i - \text{Z}_{0i})^2}{\sum_{i=1}^{N} \text{m}_i} }
\label{loss7}
\end{equation}

\section{Dataset and experimental setup}
\label{section:Data}
\subsection{Study areas}
\begin{figure*}[h!]
  \setlength{\tabcolsep}{1.5pt}
  \renewcommand{\arraystretch}{1}
  \footnotesize

\centering
  \begin{tabular}{m{0.5cm}ccc}

   \textbf{(a)} &\begin{minipage}[c]{0.44\columnwidth}
        \centering
        \includegraphics[width=\linewidth]{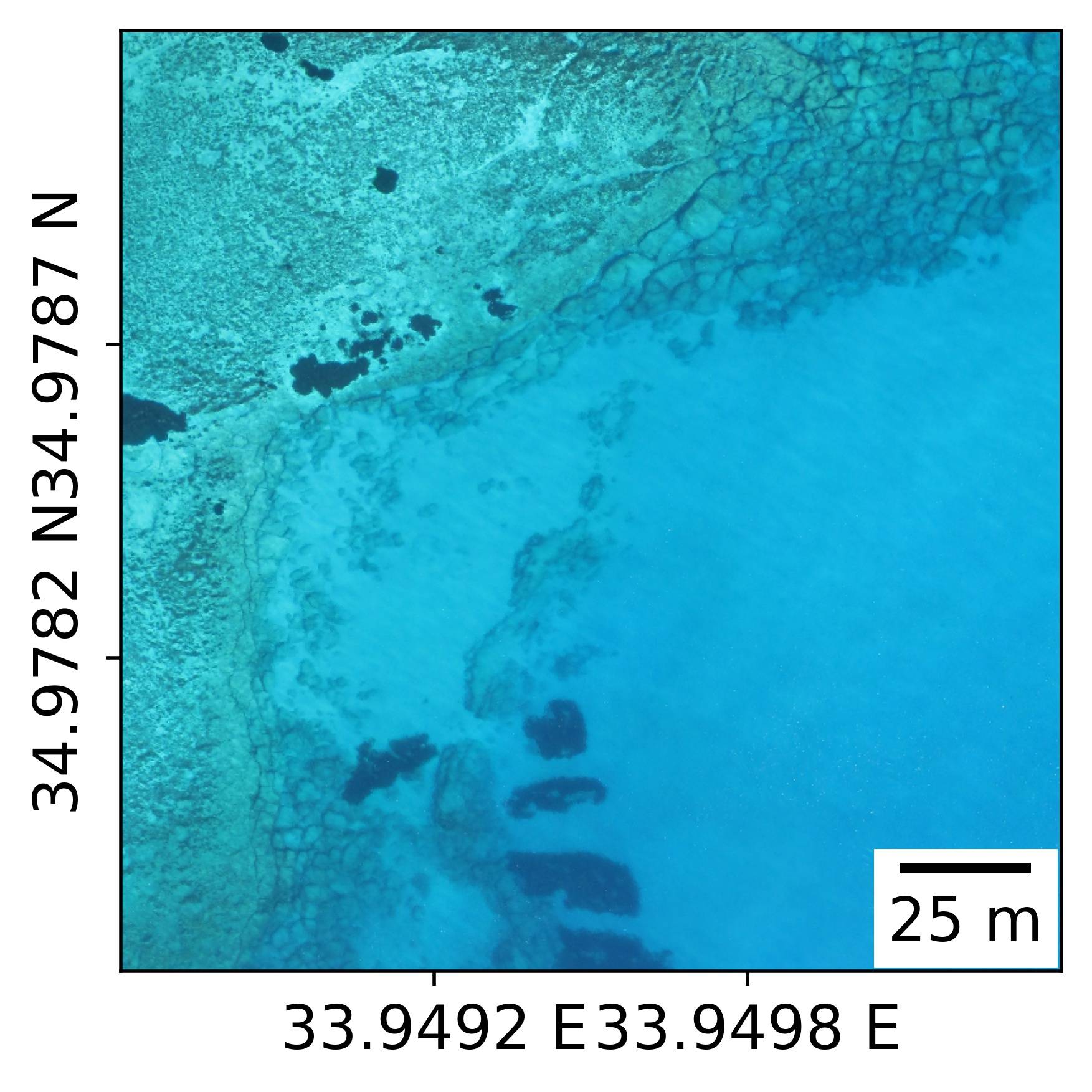}
    \end{minipage}&

    \begin{minipage}[c]{0.44\columnwidth}
        \centering
        \includegraphics[width=\linewidth]{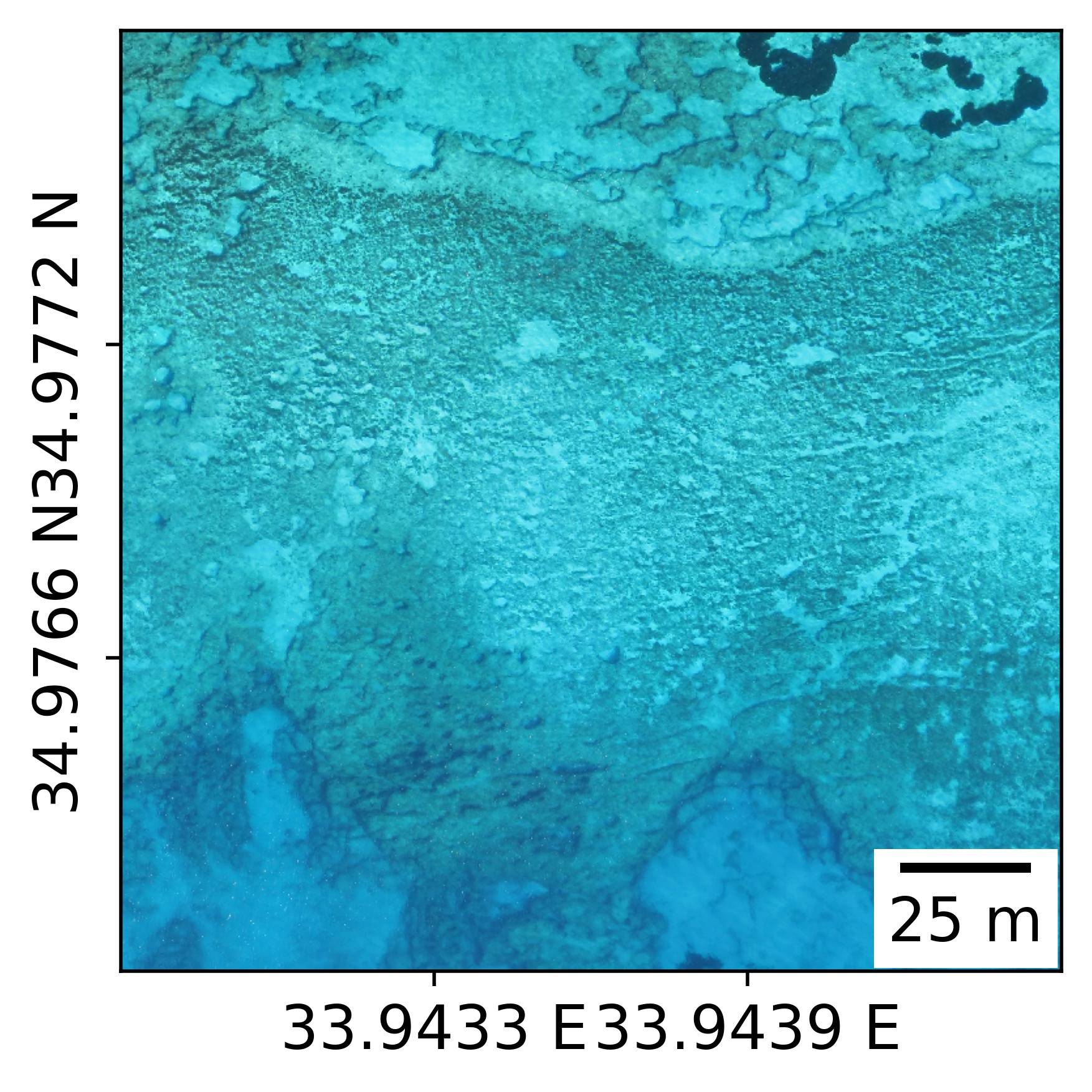}
    \end{minipage}&    
\hspace{-1.2cm}

    \begin{minipage}[c]{0.44\columnwidth}

        \includegraphics[width=\linewidth]{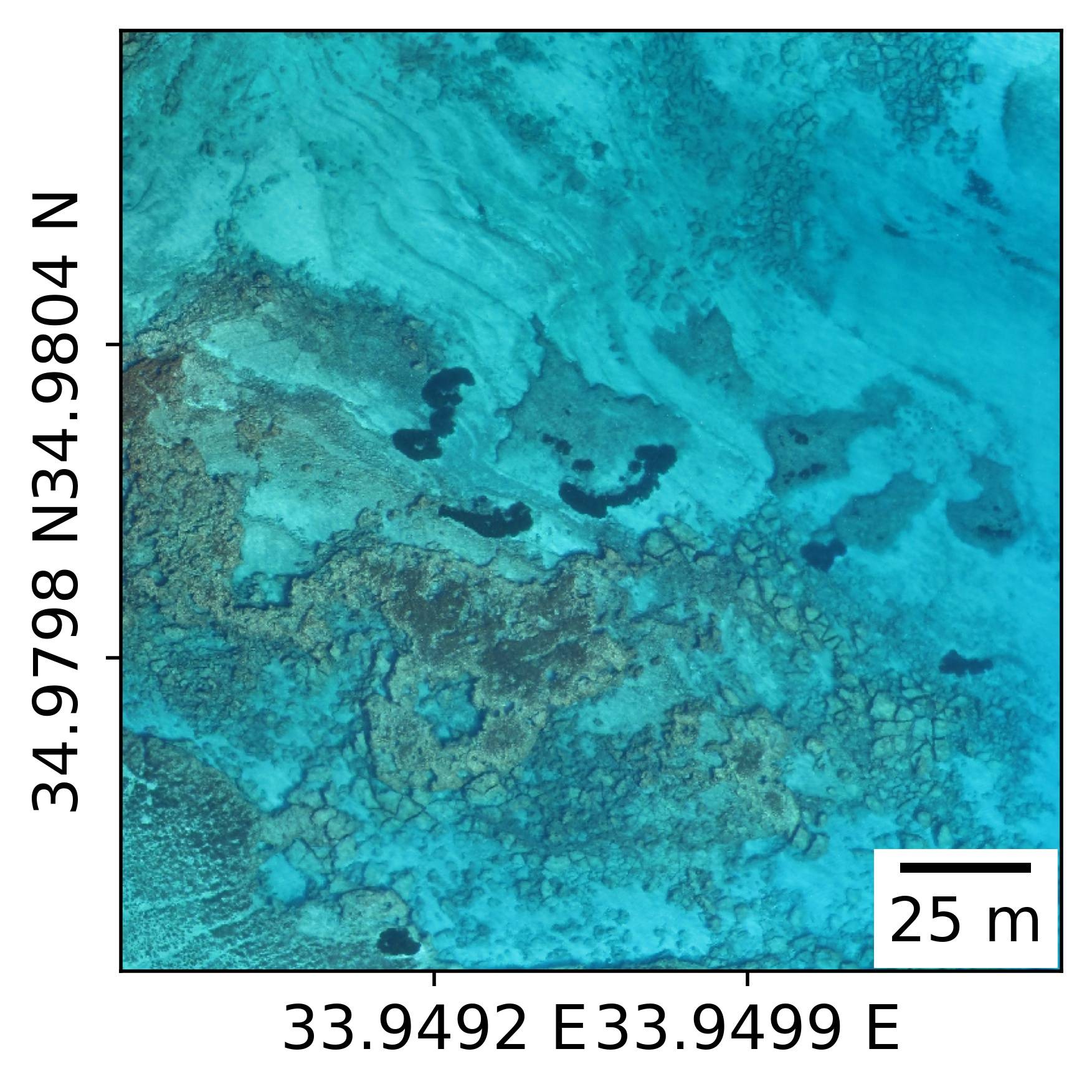}
    \end{minipage}\\ 

    \textbf{(b)} &\begin{minipage}[c]{0.44\columnwidth}
        \centering
        \includegraphics[width=\linewidth]{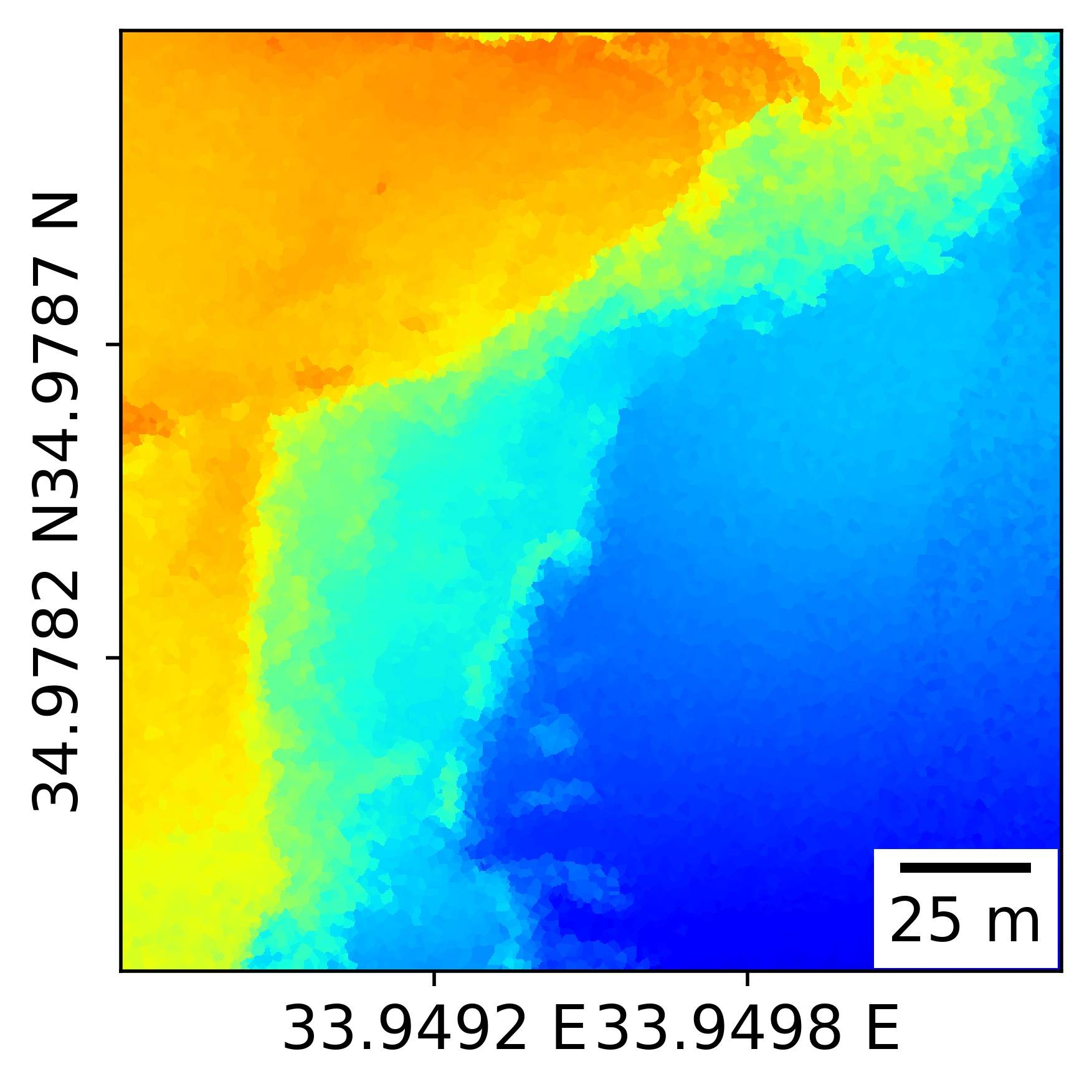}
    \end{minipage}&

    \begin{minipage}[c]{0.44\columnwidth}
        \centering
        \includegraphics[width=\linewidth]{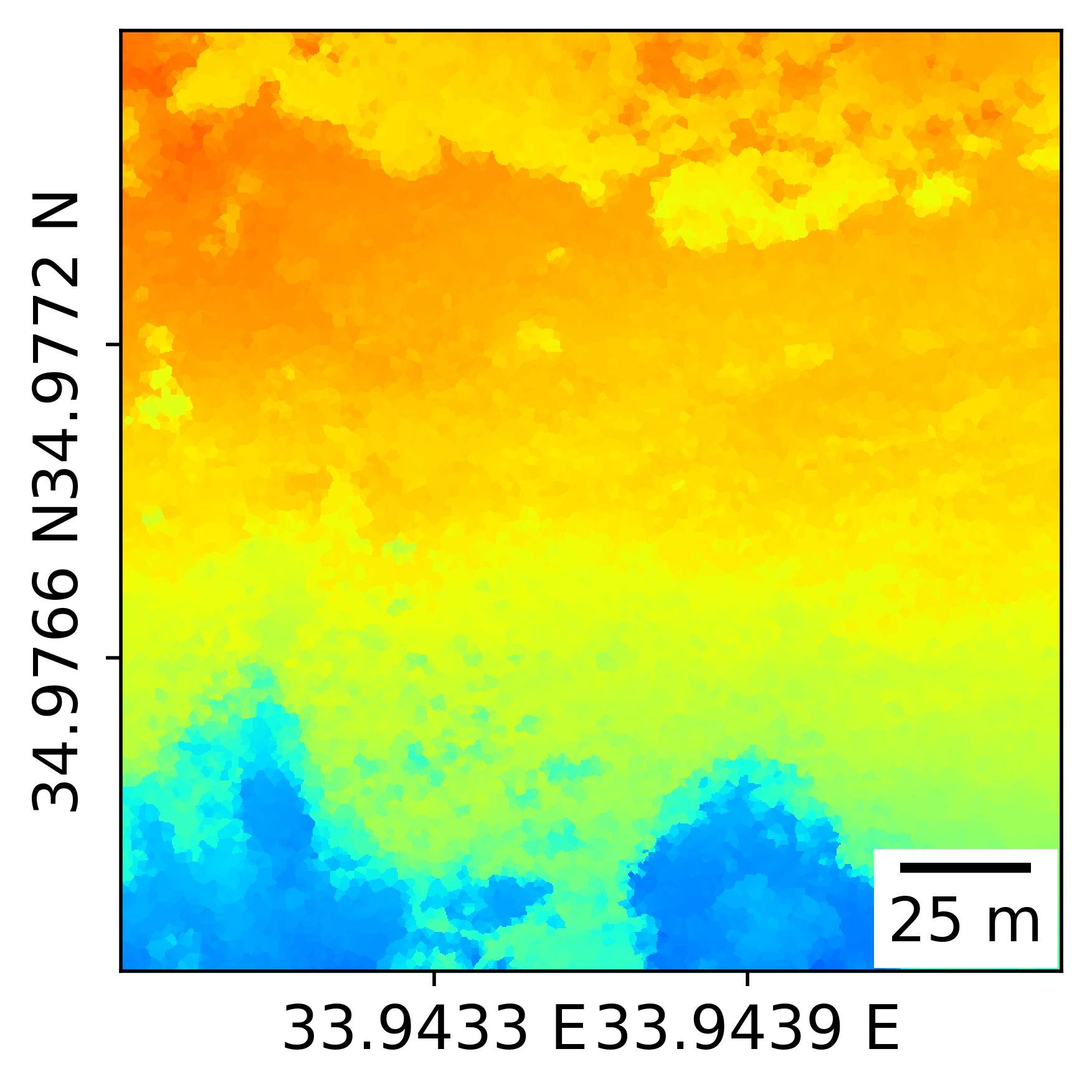}
    \end{minipage}&

    \begin{minipage}[c]{0.55\columnwidth}
        \centering 
        \includegraphics[width=\linewidth]{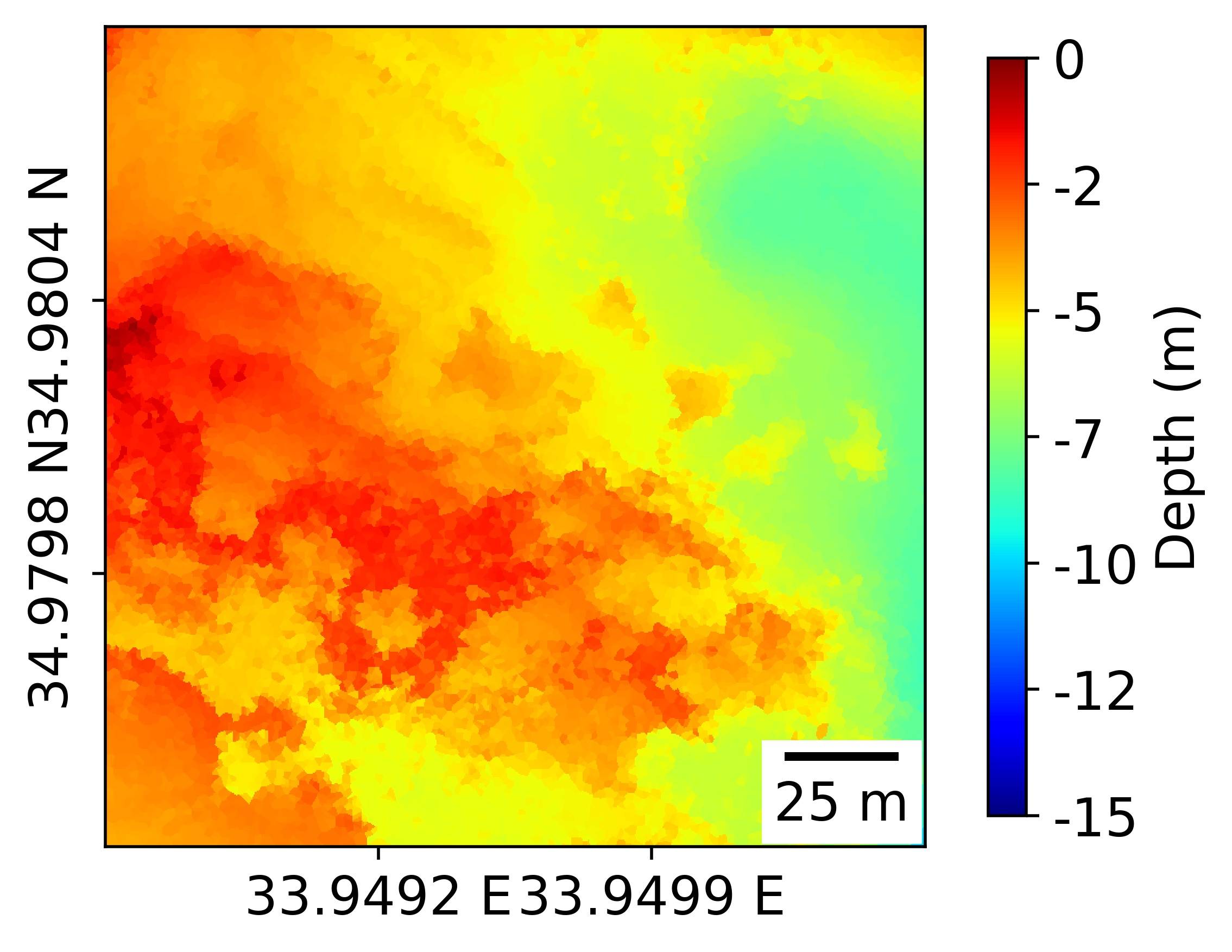}
    \end{minipage} \vspace{1pt}\\

    \textbf{(c)} & \begin{minipage}[c]{0.44\columnwidth}
        \centering
        \includegraphics[width=\linewidth]{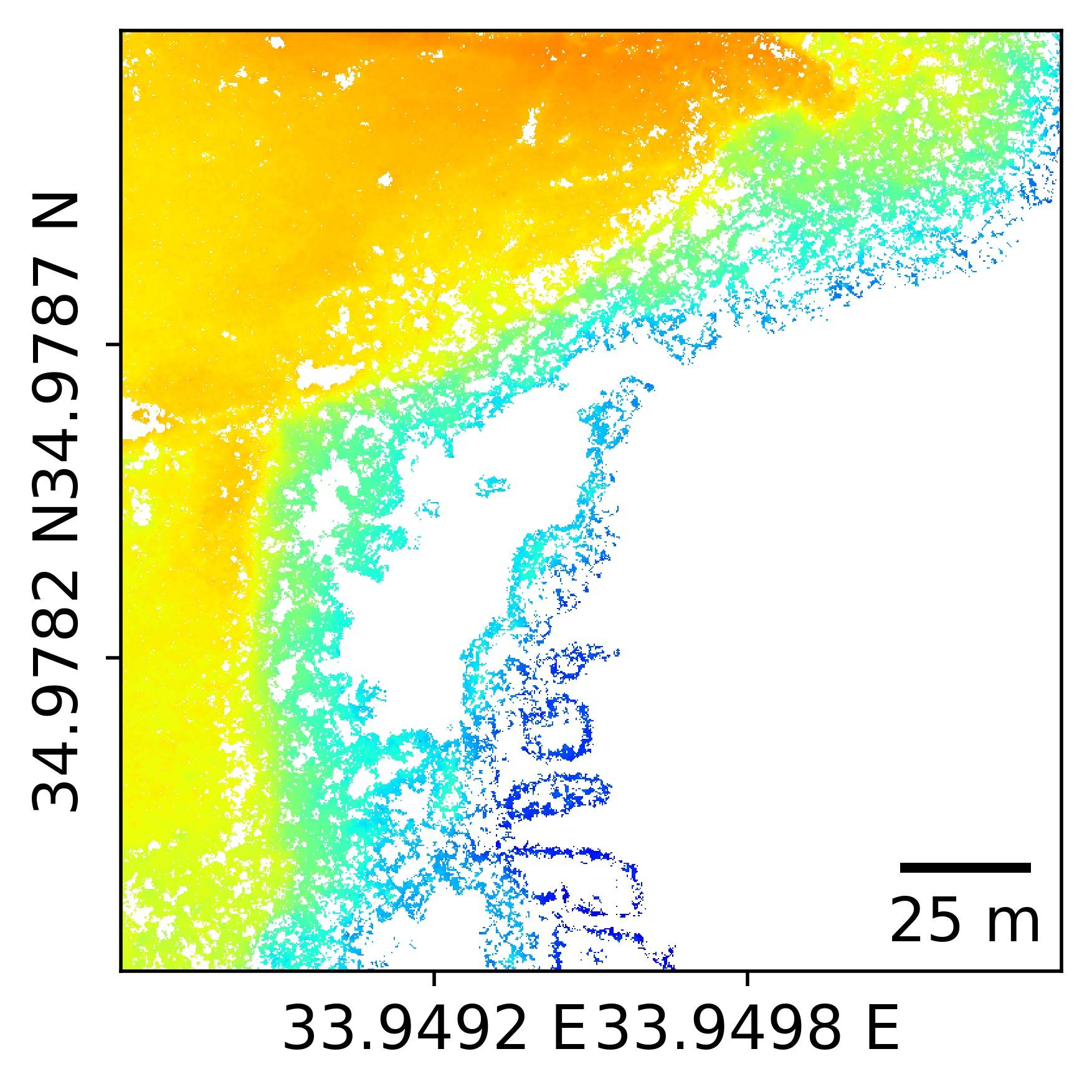}
    \end{minipage}&

    \begin{minipage}[c]{0.44\columnwidth}
        \centering
        \includegraphics[width=\linewidth]{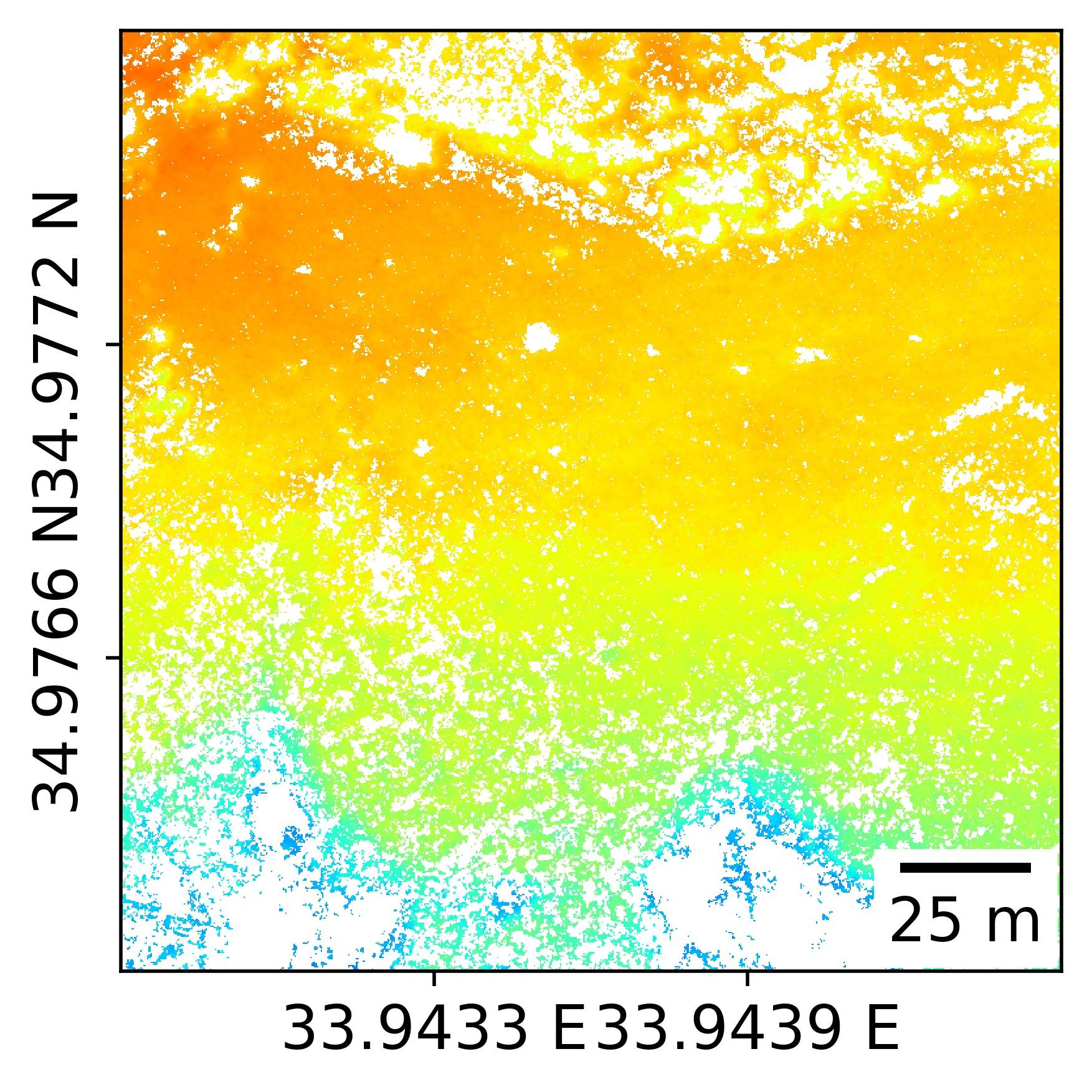}
    \end{minipage}&

    \begin{minipage}[c]{0.55\columnwidth}
        \centering 
        \includegraphics[width=\linewidth]{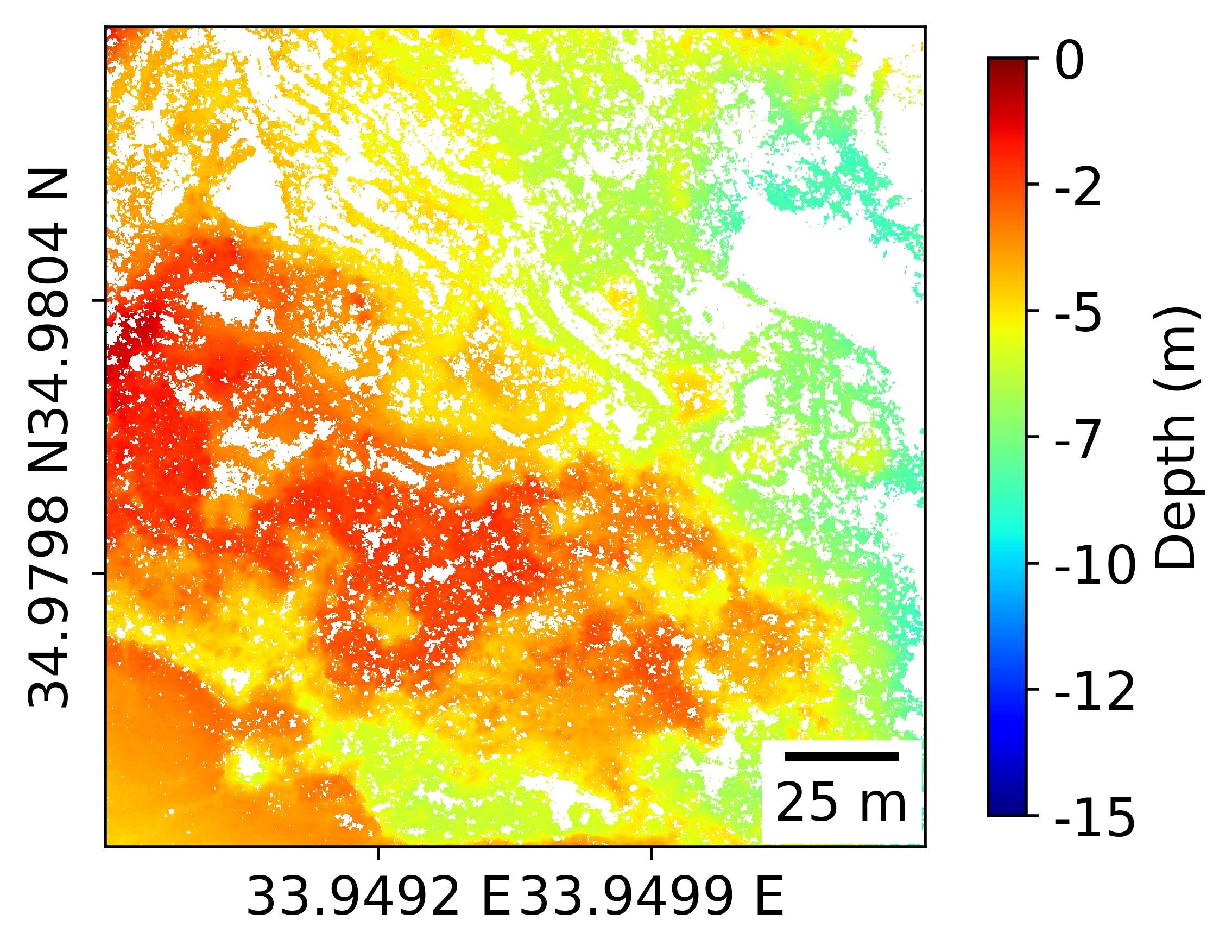}
    \end{minipage}\\

\end{tabular}
\vspace{-0.07in}
\caption{(a) Example orthoimage patches of the Agia Napa area, (b) reference bathymetry, and (c) SfM-MVS refraction corrected bathymetry, relative to the WGS '84. In bathymetry images, white color represents the missing data (gaps). }
\label{fig:fig1}
\vspace{-0.07in}
\end{figure*}

\begin{figure*}[h!]
  \setlength{\tabcolsep}{1.5pt}
  \renewcommand{\arraystretch}{1}
  \footnotesize
  
  \centering

  \begin{tabular}{m{0.5cm}cccc}
 \textbf{(a)} &   \begin{minipage}[c]{0.44\columnwidth}
        \centering
        \includegraphics[width=\linewidth]{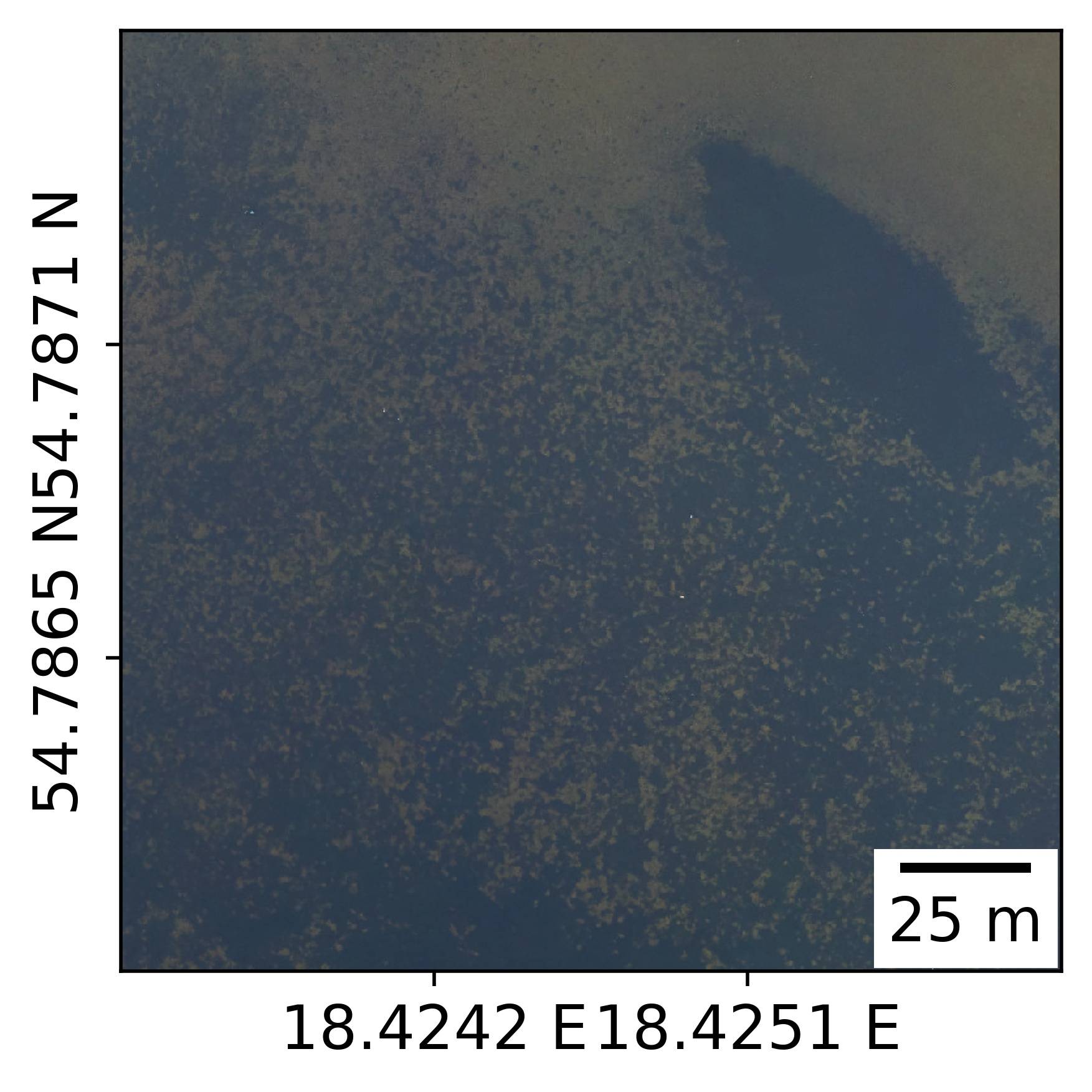}
    \end{minipage}& 
    \hfill

    \begin{minipage}[c]{0.44\columnwidth}
        \centering
        \includegraphics[width=\linewidth]{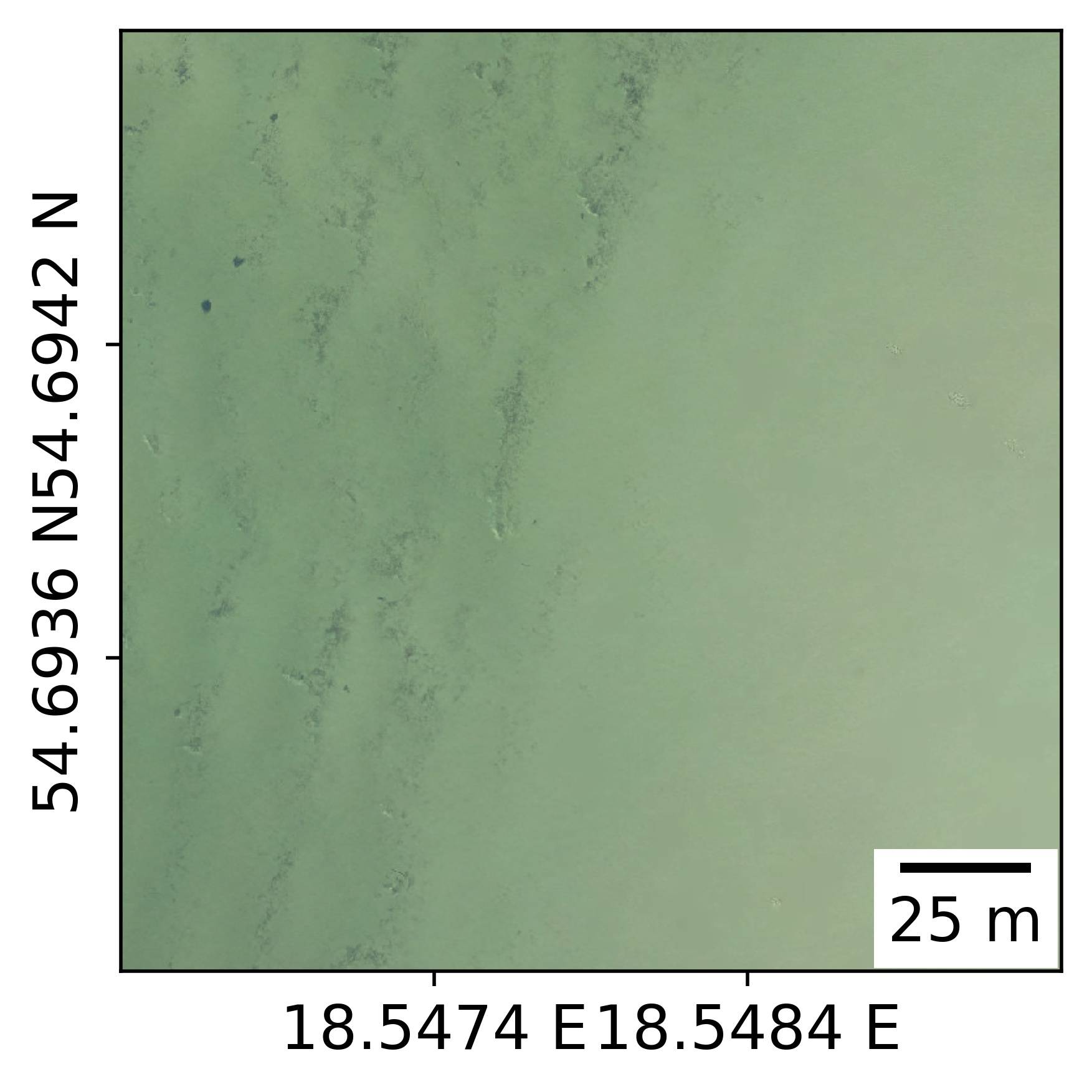}
    \end{minipage}&    
\hspace{-1cm}

    \begin{minipage}[c]{0.44\columnwidth}
    \centering
        \includegraphics[width=\linewidth]{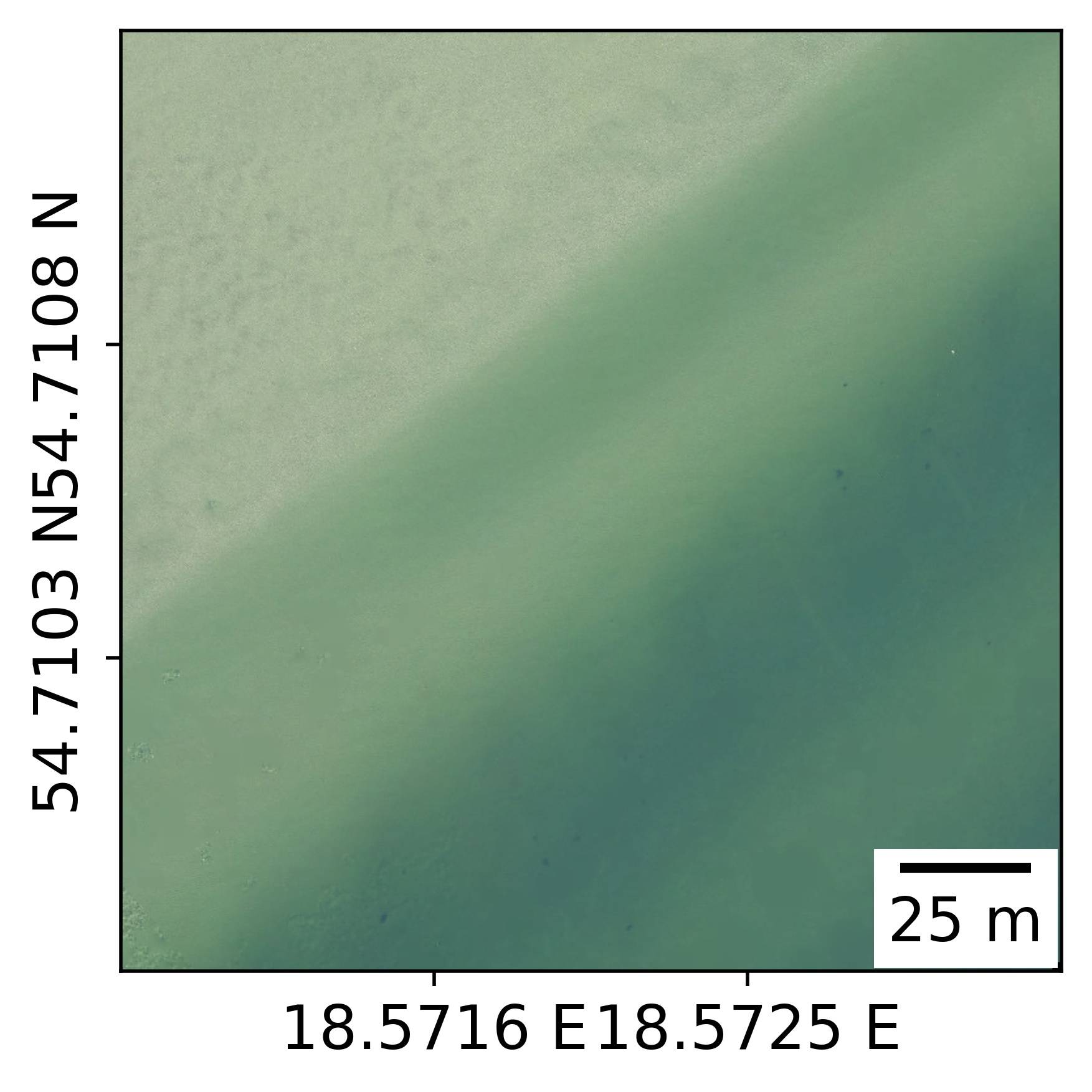}
    \end{minipage} \vspace{1pt}\\

  \textbf{(b)} &   \begin{minipage}[c]{0.44\columnwidth}
        \centering
        \includegraphics[width=\linewidth]{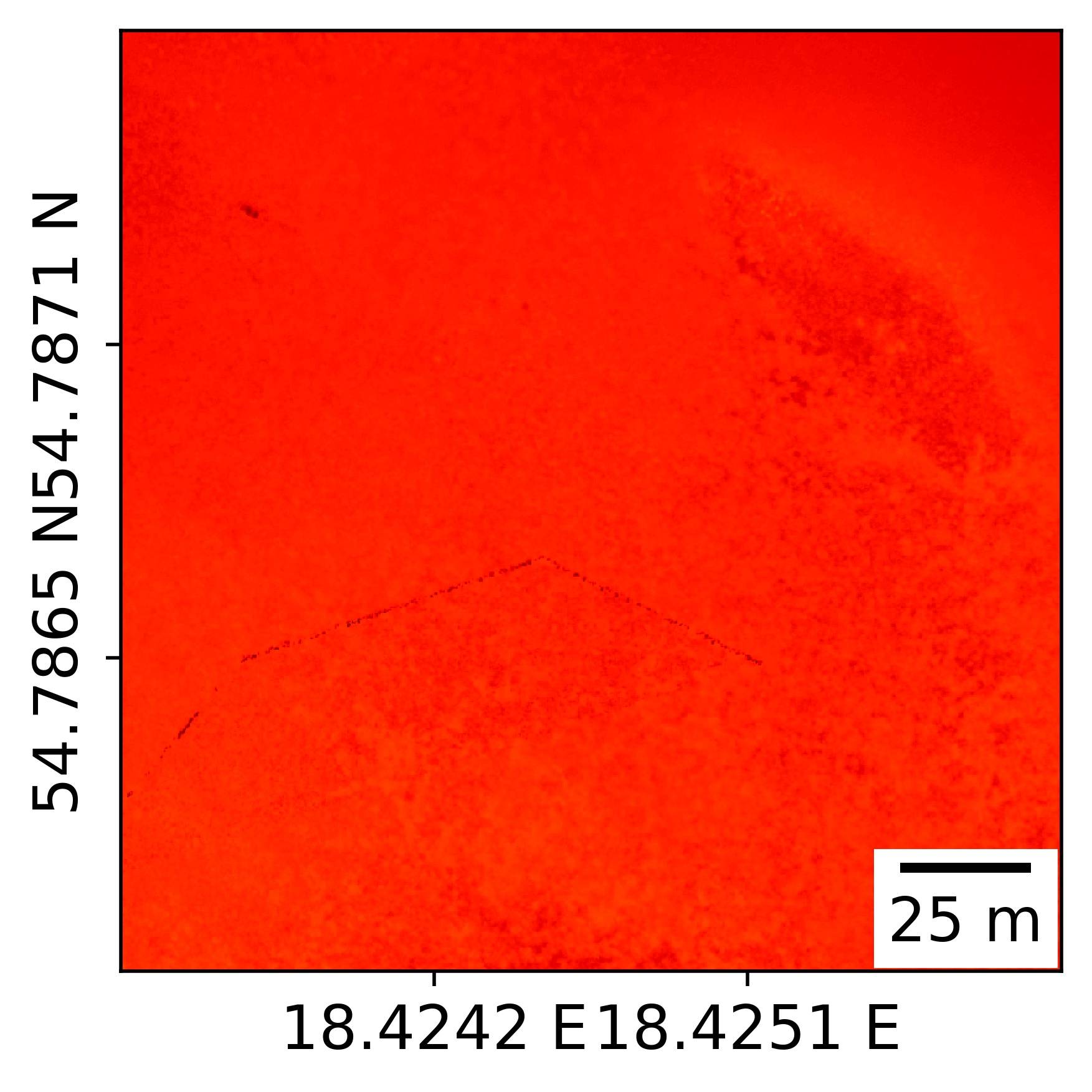}
    \end{minipage}& 
    \hfill

    \begin{minipage}[c]{0.44\columnwidth}
        \centering
        \includegraphics[width=\linewidth]{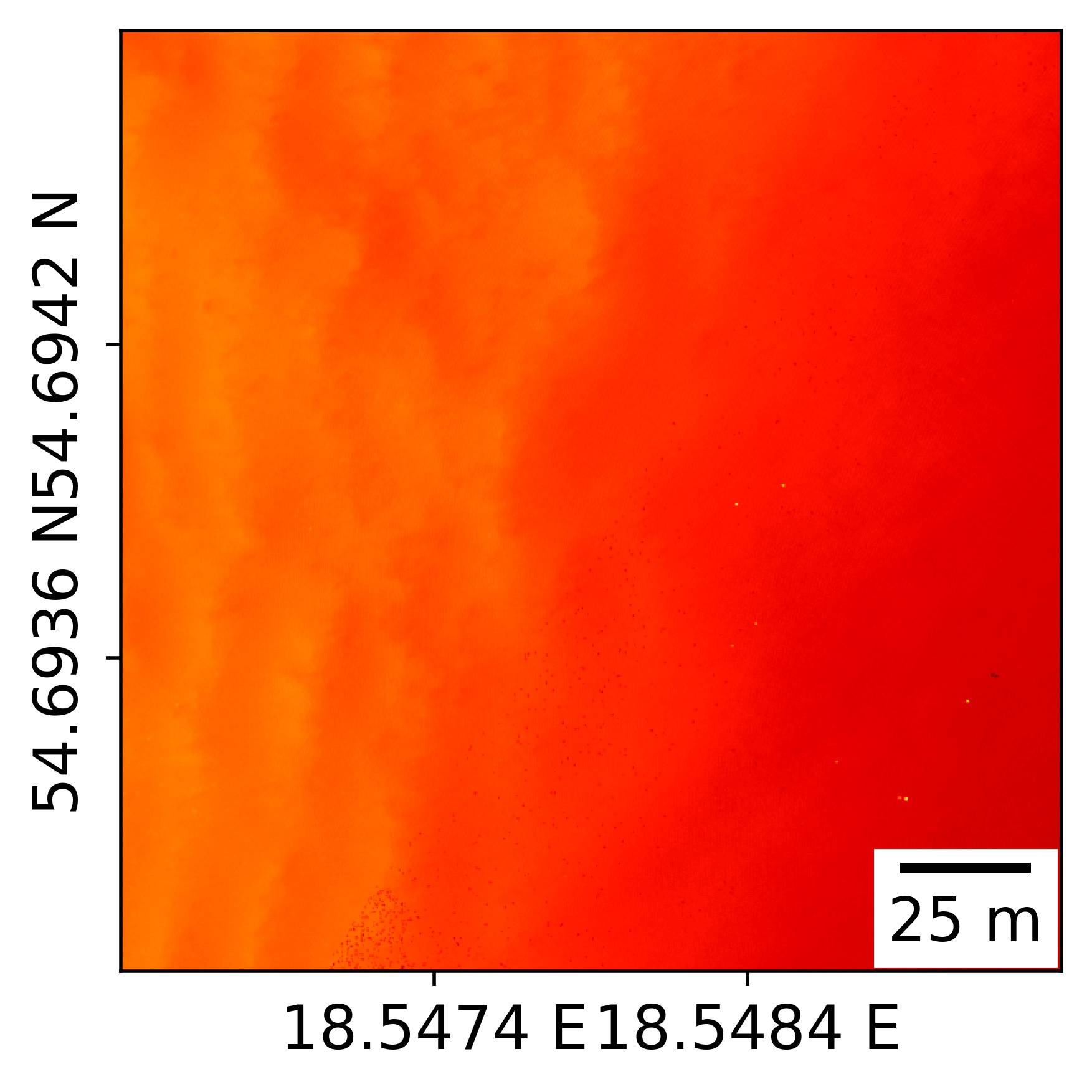}
    \end{minipage}&    
    \hfill

    \begin{minipage}[c]{0.55\columnwidth}
        \centering
        \includegraphics[width=\linewidth]{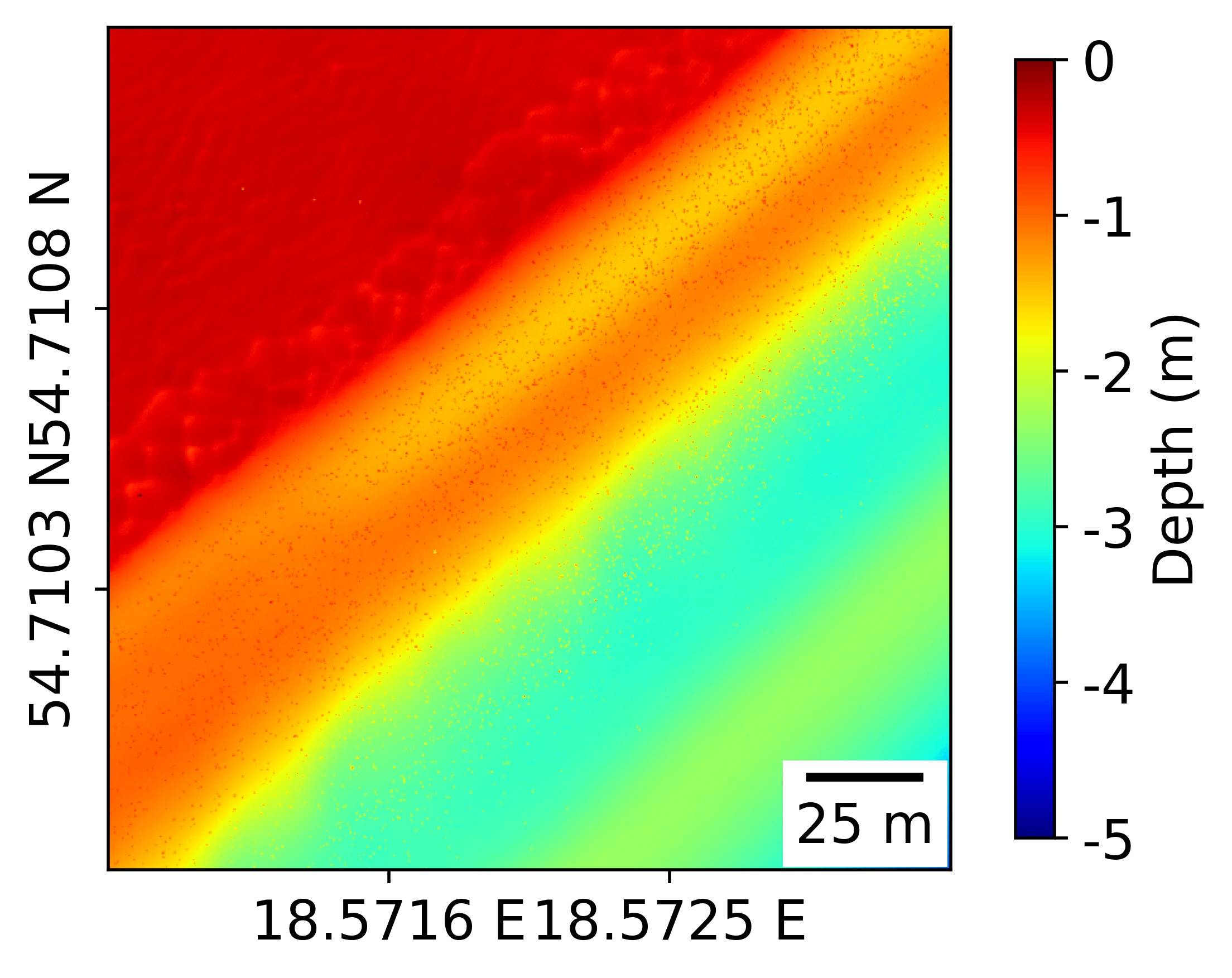}
    \end{minipage} \vspace{1pt}\\

  \textbf{(c)} &   \begin{minipage}[c]{0.44\columnwidth}
        \centering
        \includegraphics[width=\linewidth]{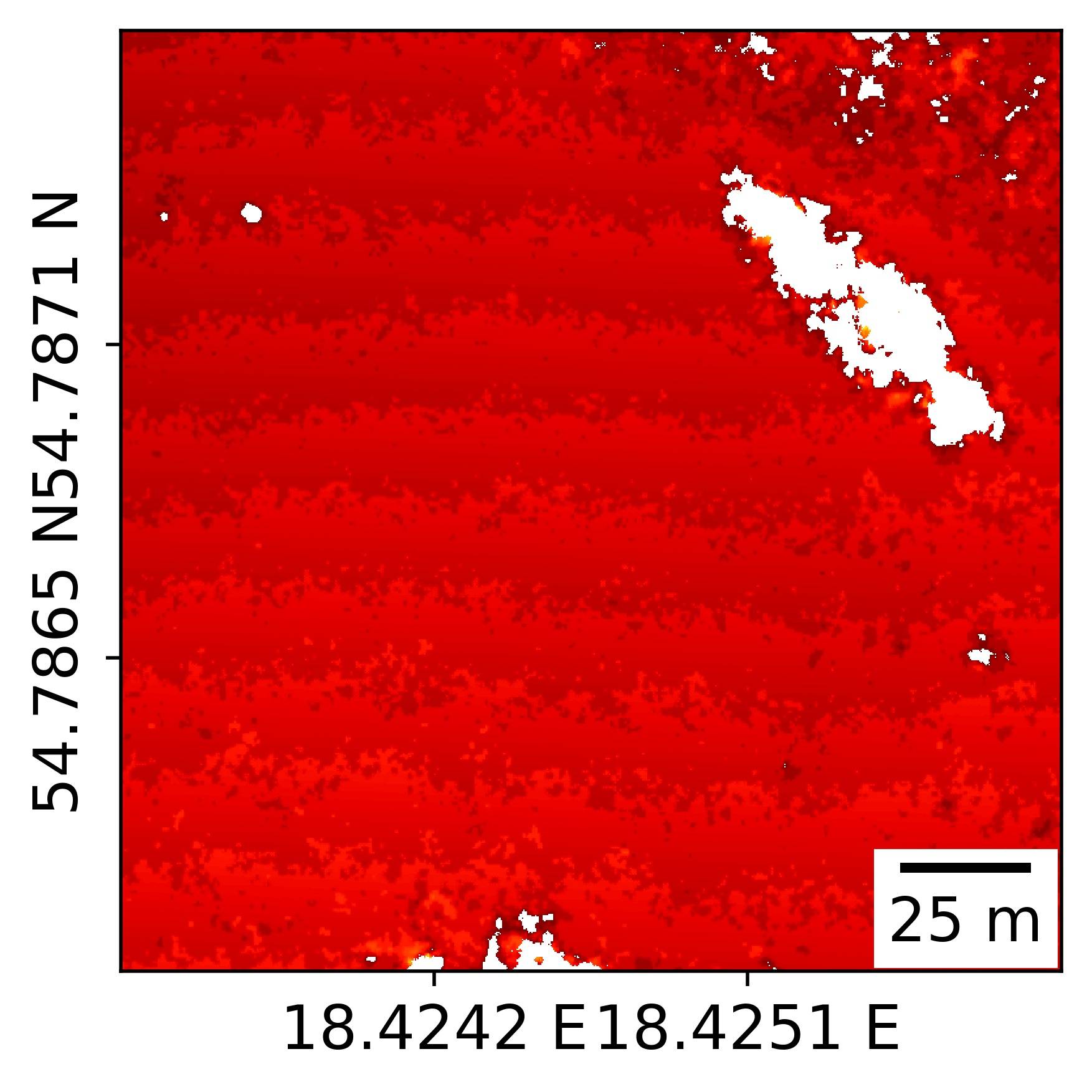}
    \end{minipage}& 
    \hfill

    \begin{minipage}[c]{0.44\columnwidth}
        \centering
        \includegraphics[width=\linewidth]{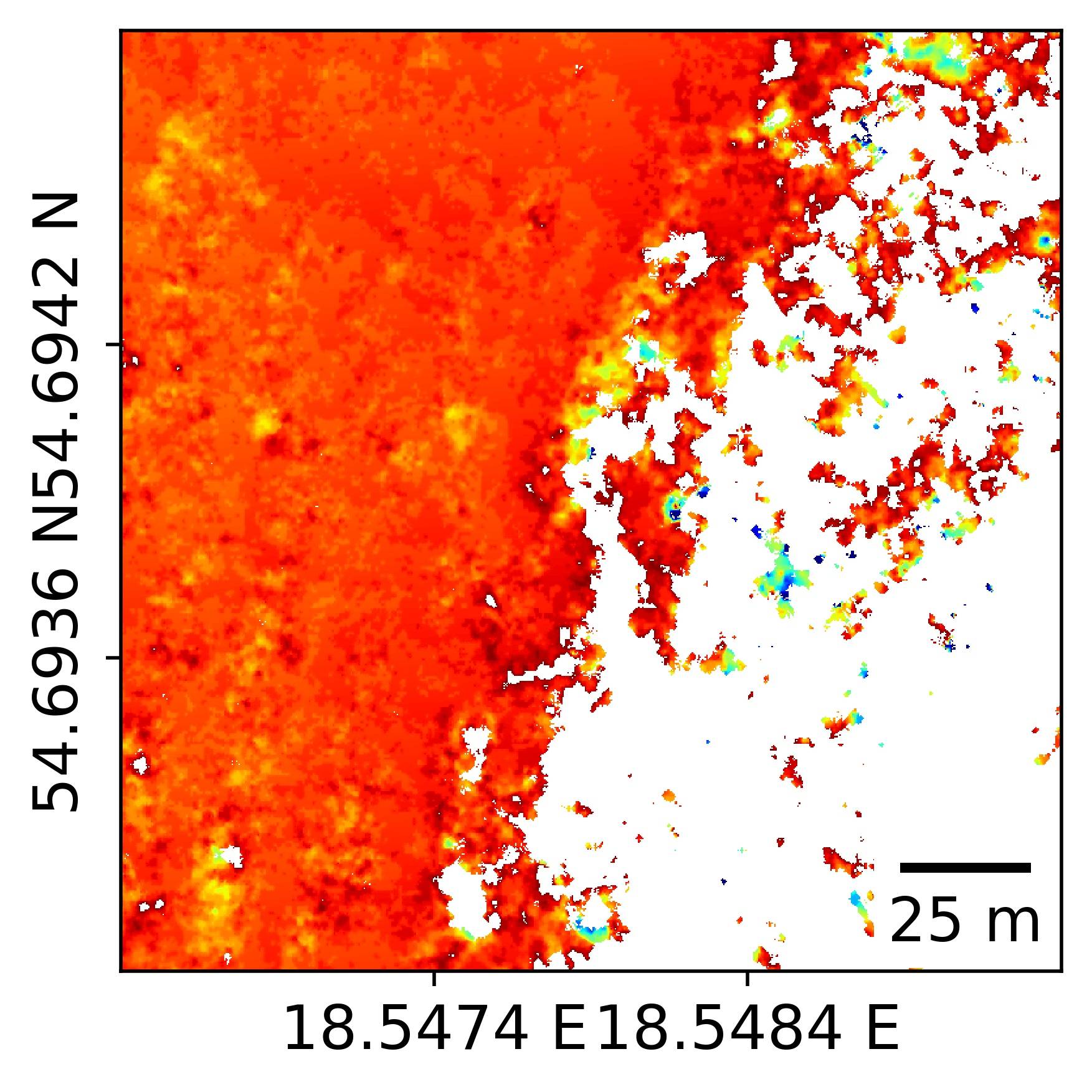}
    \end{minipage}&    
    \hfill

    \begin{minipage}[c]{0.55\columnwidth}
        \centering
        \includegraphics[width=\linewidth]{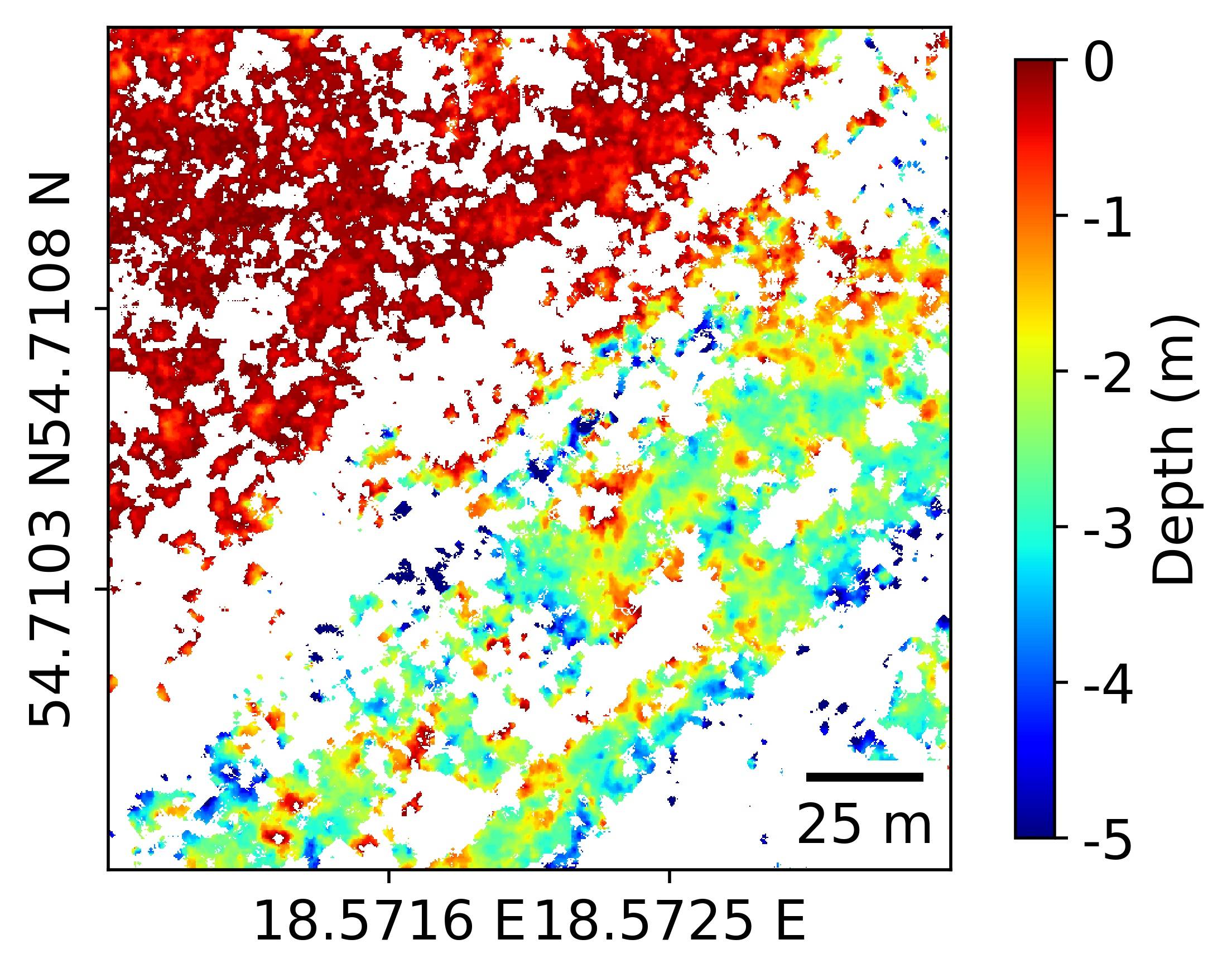}
    \end{minipage} 
 \vspace{1pt}\\
\end{tabular}
\vspace{-0.07in}
\caption{(a) Example orthoimage patches of the Puck Lagoon area, (b) reference bathymetry, and (c) SfM-MVS refraction corrected bathymetry, relative to the WGS '84 datum. In bathymetry images, white color represents the missing data (gaps).}
\label{fig:fig2}
\vspace{-0.07in}
\end{figure*}

To evaluate the proposed approach, we used aerial imagery and DSM modalities from the MagicBathyNet dataset \citep{magicbathynet}. We also enriched this dataset with the respective co-registered SfM-MVS DSM patches derived from \cite{skarlatos2018, puck, magicbathynet}. Details about the SfM-MVS steps, number of GCPs used, remaining errors in bundle adjustment, DSM generation process as well as data acquisition dates are provided in \citep{magicbathynet}. MagicBathyNet covers two distinct coastal regions with contrasting water column characteristics and seabed types: (i) the Agia Napa area in Cyprus (see Figure \ref{fig:fig1}), which represents a diverse range of typical Mediterranean coastal waters and seabed compositions, including rock, sand, seagrass (Posidonia oceanica), and macroalgae (Filamentous/turf algae); and (ii) the Puck Lagoon in Poland (see Figure \ref{fig:fig2}), which is a characteristic lagoon of the Baltic Sea, predominantly featuring eelgrass/pondweed (Zostera marina,
Stuckenia pectinata, Potamogeton perfoliatus) and sand, while also encompassing ports, marinas, estuaries, and dredged areas. In Agia Napa, the seabed of the area used in this work reaches a depth of -15.50m. Bathymetric LiDAR data were collected using the Leica HawkEye III system. The aerial image collection was conducted using a fixed-wing UAV equipped with a Canon IXUS 220HS camera, featuring a 4.30mm focal length, 1.55$\mu$m pixel size, and a 4000×3000 pixels format. The selected area of the Puck Lagoon reaches a depth of -5.80m. Bathymetric data obtained using the Riegl VQ-880-GII LiDAR and Teledyne Reson T50/T20 multibeam echo-sounders. Aerial imagery was obtained using a Phase One iXM-100 camera with a 35mm focal length, 3.7$\mu$m pixel size, and a 11664x8750 pixels format. SfM-MVS and orthoimage generation was performed using the commercial photogrammetry software Agisoft Metashape. The orthoimages were then divided into non-overlapping patches. Each orthoimage and bathymetry patch covers an area of 180x180m, depicted in 720x720 pixels (ground sampling distance (GSD) of 0.25m) in the aerial modalities. A total of 21 patches, covering 0.7 km\textsuperscript{2}, were used for Agia Napa, while 2019 patches, spanning 65.4 km\textsuperscript{2}, were used for Puck Lagoon.

Figure \ref{fig:fig1}a shows example orthoimage patches of the Agia Napa area while Figure \ref{fig:fig1}b shows their respective LiDAR/multi-beam echosounder bathymetry from \citep{magicbathynet}, which will be referred to as reference bathymetry used for the evaluation of the predicted bathymetry in this work. Figure \ref{fig:fig1}c shows the respective SfM-MVS refraction corrected bathymetry used to train the learning-based bathymetry model, all relative to the World Geodetic System (WGS) '84. Figures \ref{fig:fig2}a, b, and c show the respective example patches for the Puck Lagoon Area. SfM-MVS bathymetry in both areas is incomplete mainly due to the large sandy and seagrass-covered areas at the bottom as well as the characteristics of the water column.

\subsection{Causes of data gaps in SfM-MVS DSM}
\label{III.B}
Figures \ref{fig:fig1} and \ref{fig:fig2} provide evidence of a clear correlation between depth and the results obtained from SfM-MVS, particularly when considering the combination of depth and the homogeneous nature of the mostly sandy seabed. As can be seen there, at both sites, when depth is increased, the SfM-MVS DSM has more data gaps.

For the Agia Napa area, a significant factor contributing to this phenomenon is the degradation of image quality resulting from a combination of strong refraction effects and reduced visibility caused by absorption and scattering, particularly at depths exceeding 9 to 10m. As the density of seawater varies with factors such as temperature, salinity, and pressure, the deeper the depth, the more layers of varying refractive indices the light rays encounter. Consequently, due to refraction, pixels that could otherwise be matched may end up positioned far from the epipolar line, resulting in either their exclusion or the creation of inaccurate correspondences. This depth limitation aligns with \cite{legleiter2019defining}, where the maximum reliable detectable depth in the conducted tests was determined to be 9.50m. 

Analysis of SfM results for the Agia Napa test site revealed that beyond the depth of 9-10m, although a dense point cloud was generated, it lacked the necessary matches in the SfM process to ensure precise and reliable 3D reconstruction. Although for the specific case of the Agia Napa test site, the seafloor at depths greater than 9-10m is primarily characterized by large sandy areas with scattered patches of seagrass and rocks-this might explain some of the mismatches. However, it has also been observed that some seagrasses create distinctive blobs for pyramid matching because of their strong gradients, which result from the high contrast between the seagrass and the surrounding sand. In these depths, even if more detailed features such as the scattered rocky formations exist, which in the shallower areas serve as distinctive and unique points for feature detection and matching, they are not suitable for key points, obviously due to the blurriness and the general deterioration of the image quality because of the depth. Same phenomena are also apparent in the Puck Lagoon area in much shallower depths; however, in these cases it is related to limited water visibility and clarity due to suspended matter and chlorophyll concentration. Furthermore, as can also be seen in Figure \ref{fig:fig2}c, due to the conditions of the water column described above, the DSM derived from SfM-MVS is noisier and limited to relatively shallow depths compared to Agia Napa. Accordingly, in this work we aim to exploit this sparse bathymetric information generated by the SfM-MVS, being in the boundaries of the distinctive blobs of seagrass and rock. That way, depths will also be predicted for the deeper parts of the sites or the ones without texture, overcoming issues like image blurring and overall degradation in quality due to increased depth. 

\subsection{Experimental setup}
\label{experimentalsetup}
\subsubsection{SVR training and inference}
The linear SVR model deployed in this study is trained on synthetic pairs of $Z$ and $Z_0$ from \cite{agrafiotis2021}. The advantage of using synthetic data for training this model lies in the precision and reliability of the depth information, as synthetic data can provide perfectly accurate depth labels. Additionally, there is complete knowledge of both the exterior and interior orientations of the cameras employed, resulting in a flawless SfM-MVS process and DSM. In the context of seabed imaging, errors and limitations in image matching due to visibility constraints at depths greater than 10-15m \citep{agrafiotisphd}, as well as inaccuracies introduced by a wavy surface, are eliminated, making refraction the only unknown variable. Furthermore, by employing a mathematical function to generate and describe the DSMs, potential incompatibilities and errors that could arise from using real-world bathymetric data are avoided. This results in high-quality training data. As demonstrated in \cite{agrafiotis2021}, SVR models trained for refraction correction using synthetic data effectively address the effects of refraction, outperforming the state-of-the-art.

To train and infer using the SVR model, we utilized the scikit-learn implementation \citep{scikit-learn} within a Python environment. The estimator kernel was set to linear, and to prevent overfitting during the training process, we fine-tuned the hyper-parameter $C$ of the estimator through a grid search approach \citep{bergstra2012random}. In our experiments, the hyper-parameter $C$ was typically determined to be around 0.1, while the remaining parameters were configured to their default values \citep{pedregosa2011scikit}. The SVR model (Equation \ref{eq:equation4.5}) was trained using synthetic real $Z$ and apparent $Z_0$ depths and is capable of predicting the real depth in scenarios where only the apparent depth is available, such as in the provided SfM-MVS DSM data. For both test areas, the same SVR model was used, specifically the one trained on synthetic data acquired from 150m height above sea level, as described in \cite{agrafiotis2021}.

\subsubsection{Swin-BathyUNet training and inference}
We implemented our method in PyTorch using one NVIDIA A100 80GB GPU. For training, we employed a boundary-sensitive weighted (BSW) RMSE loss function alongside the Adam optimizer \citep{adam}. For Agia Napa data, the initial learning rate was set to 2.5 x $10^{-4}$, and a cosine annealing scheduler \citep{cosine} was utilized over 30 epochs. For Puck Lagoon area, the initial learning rate was set to 1.25 x $10^{-4}$ using the same scheduler for 60 epochs. The BSW RMSE loss function featured a linear decay of weights, ranging from a minimum of 1 to a maximum of 2. The ViT uses an embedding dimension of 512, 256, and 128 for each block respectively, with a depth of 1 and 8 attention heads, alongside a window size of 64 , an MLP ratio of 4, and a patch embedding size of 32. The U-Net includes four encoder and decoder layers with 64, 128, 256, and 512 filters, respectively. A dropout rate of 0.1 is applied throughout the model to mitigate overfitting. The network takes in 3 input channels (RGB) and produces 1 output depth channel. Data normalized to [0, 1] range by dividing the pixels values of the RGB orthoimages with 255 and the depths with -15m for the Agia Napa area and with -6m for the Puck Lagoon area. Images and depths were used in their original size i.e. 720x720 pixels. As for the data augmentation, we performed random rotations, as well as random vertical and horizontal flips. Sun glint was removed from the orthoimages during training and inference. Since a near-infrared band was not available, sun glint removal was performed by identifying and replacing pixel values that exceed a specified brightness threshold, corresponding to the glint caused by sunlight reflecting off the water surface. By replacing these high pixel values with the average values of their neighboring pixels, the function helped mitigate the sun glint, resulting in a more accurate bathymetry. 

For training we used pairs of RGB orthoimage patches, along with georeferenced and refraction-corrected SfM-MVS DSM patches. Importantly, the same image RGB patches were employed for both training and testing because the areas requiring prediction (i.e., the gaps) were not visible in the training DSM data. Independent reference data acquired via LiDAR and multibeam echo-sounder (MBES) systems, not used in the training or testing process, were utilized to evaluate those areas. Evaluation metrics included RMSE: {\scriptsize $\sqrt{\frac{1}{n}\sum_{i=1}^{n} (y_{i} - \hat{y}_{i})^2}$}, mean absolute error (MAE): {\scriptsize $\frac{1}{n}\sum_{i=1}^{n} |y_{i} - \hat{y}_{i}|$}, and standard deviation (Std.): {\scriptsize $\sqrt{\frac{1}{n} \sum_{i=1}^{n} (y_{i} - \mu)^2}$}.

\section{Experimental results and analysis}
\label{section:Results}

\subsection{Refraction correction method performance}
\begin{figure*}[h!]
\vspace{-0.1in}
  \setlength{\tabcolsep}{1.5pt}
  \renewcommand{\arraystretch}{1}
  \footnotesize
\centering
  \begin{tabular}{m{0.5cm}ccc}

   \textbf{(a)} &\begin{minipage}[c]{0.44\columnwidth}
        \centering
        \includegraphics[width=\linewidth]{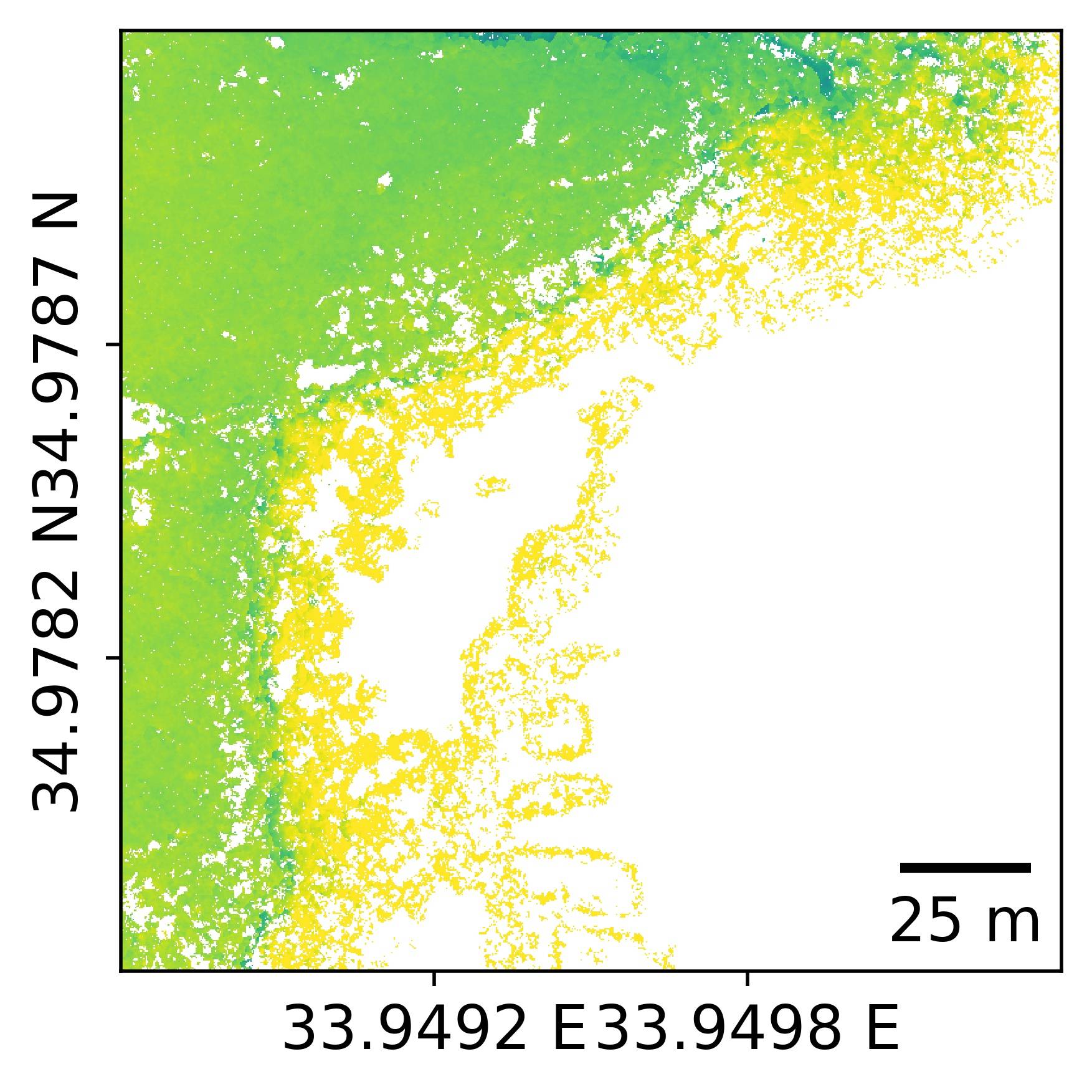}
    \end{minipage}& 
    \hfill

    \begin{minipage}[c]{0.44\columnwidth}
        \centering
        \includegraphics[width=\linewidth]{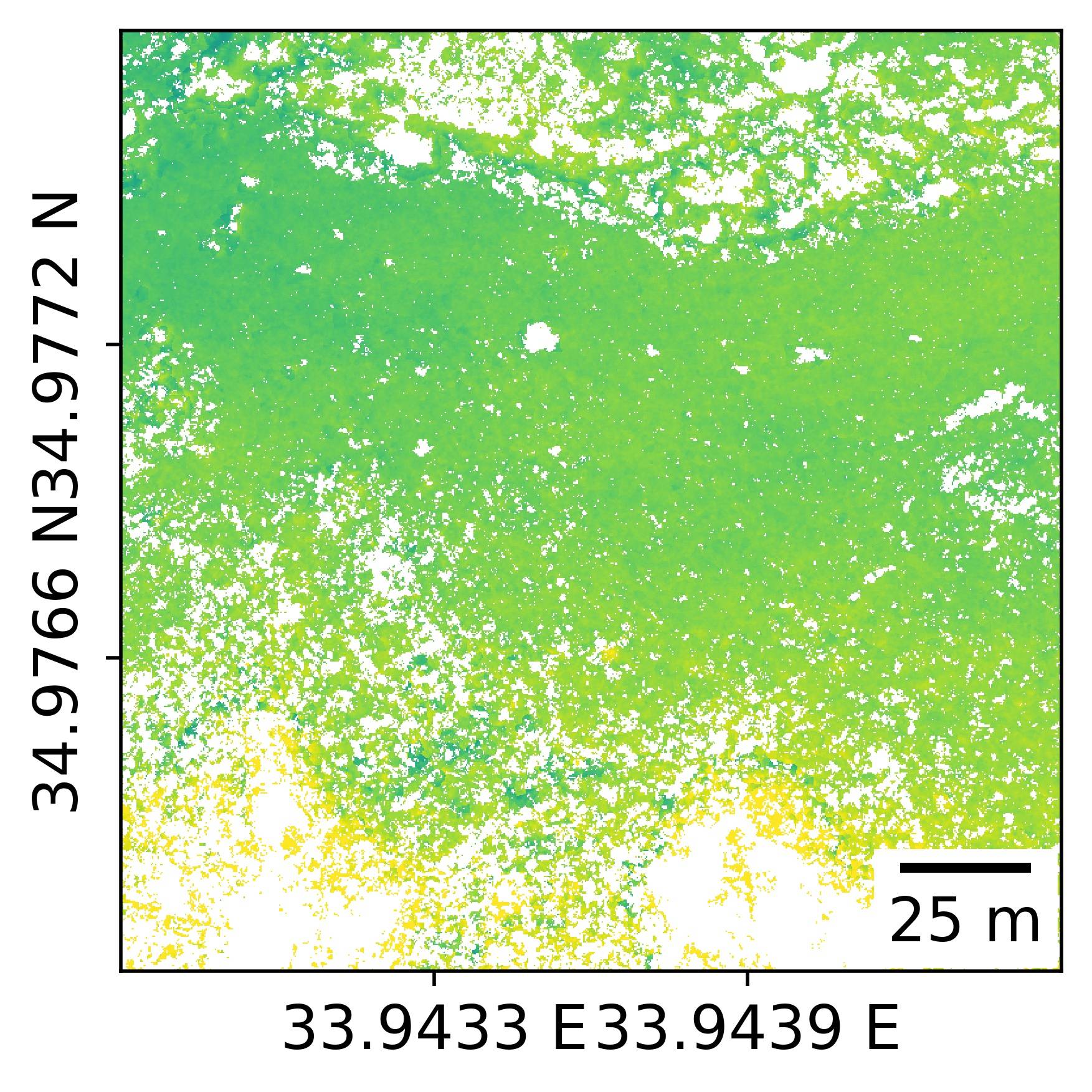}
    \end{minipage}&    
    \hfill

    \begin{minipage}[c]{0.55\columnwidth}
        \centering
        \includegraphics[width=\linewidth]{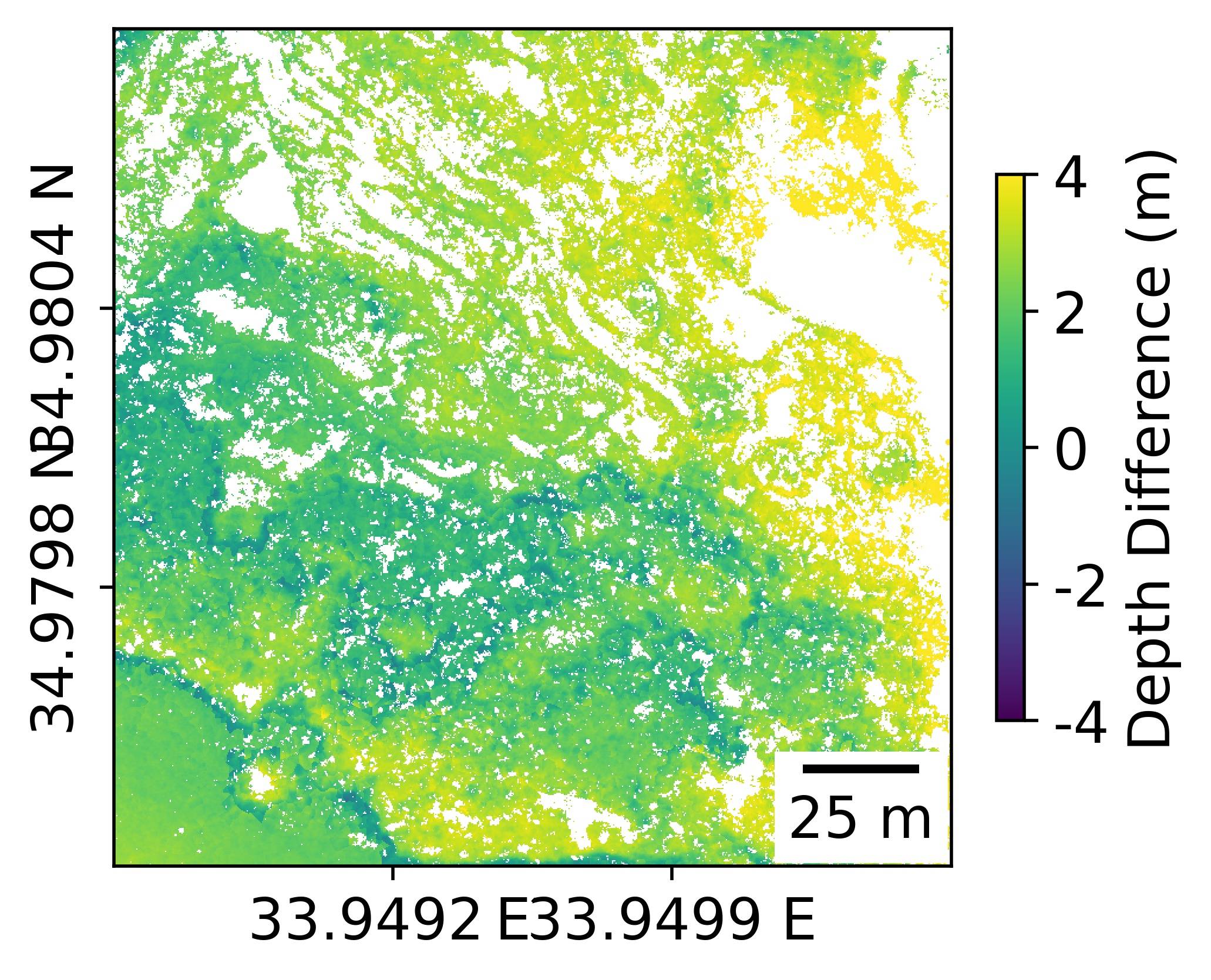}
    \end{minipage}  \vspace{1pt}\\

       \textbf{(b)} &\begin{minipage}[c]{0.44\columnwidth}
        \centering
        \includegraphics[width=\linewidth]{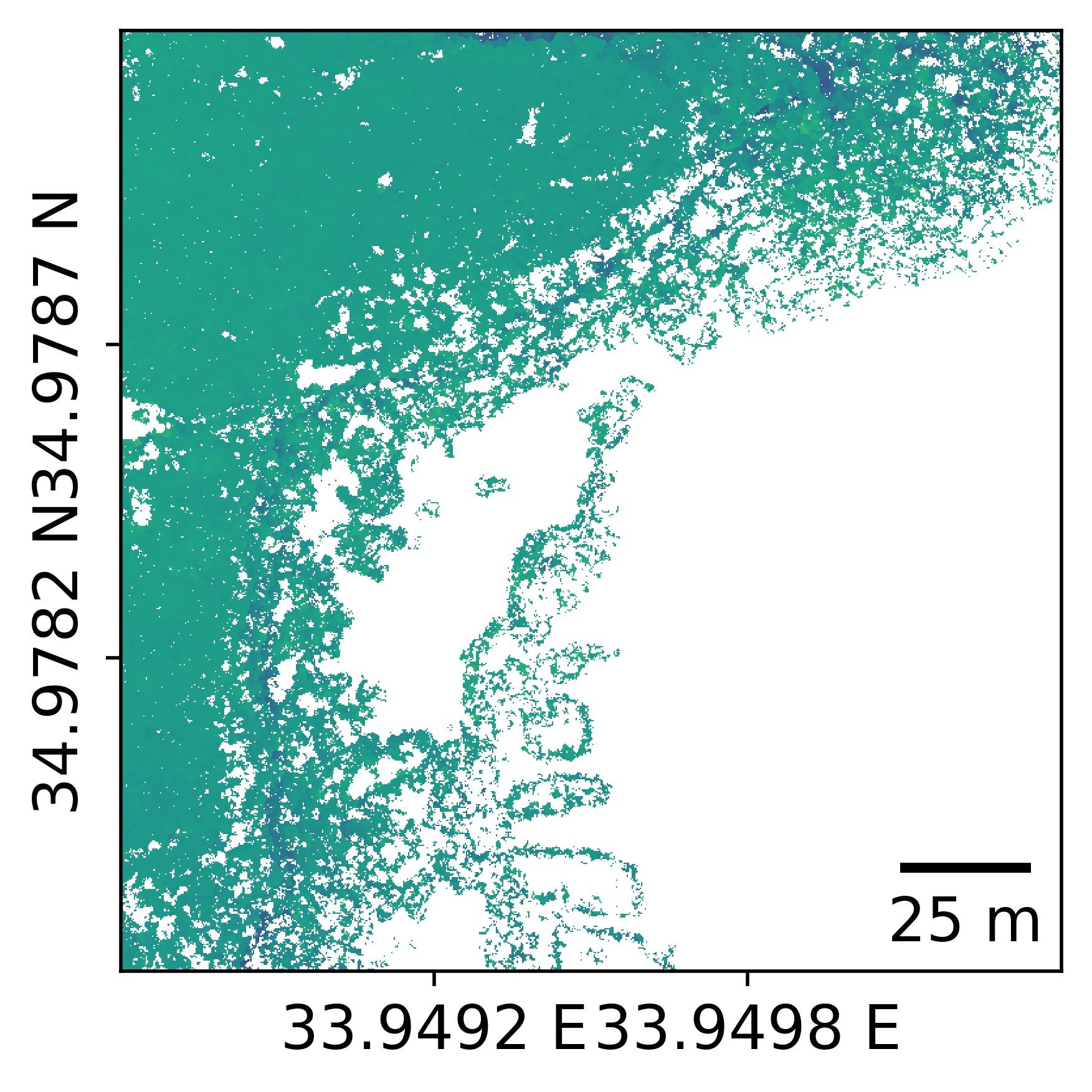}
    \end{minipage}& 
    \hfill

    \begin{minipage}[c]{0.44\columnwidth}
        \centering
        \includegraphics[width=\linewidth]{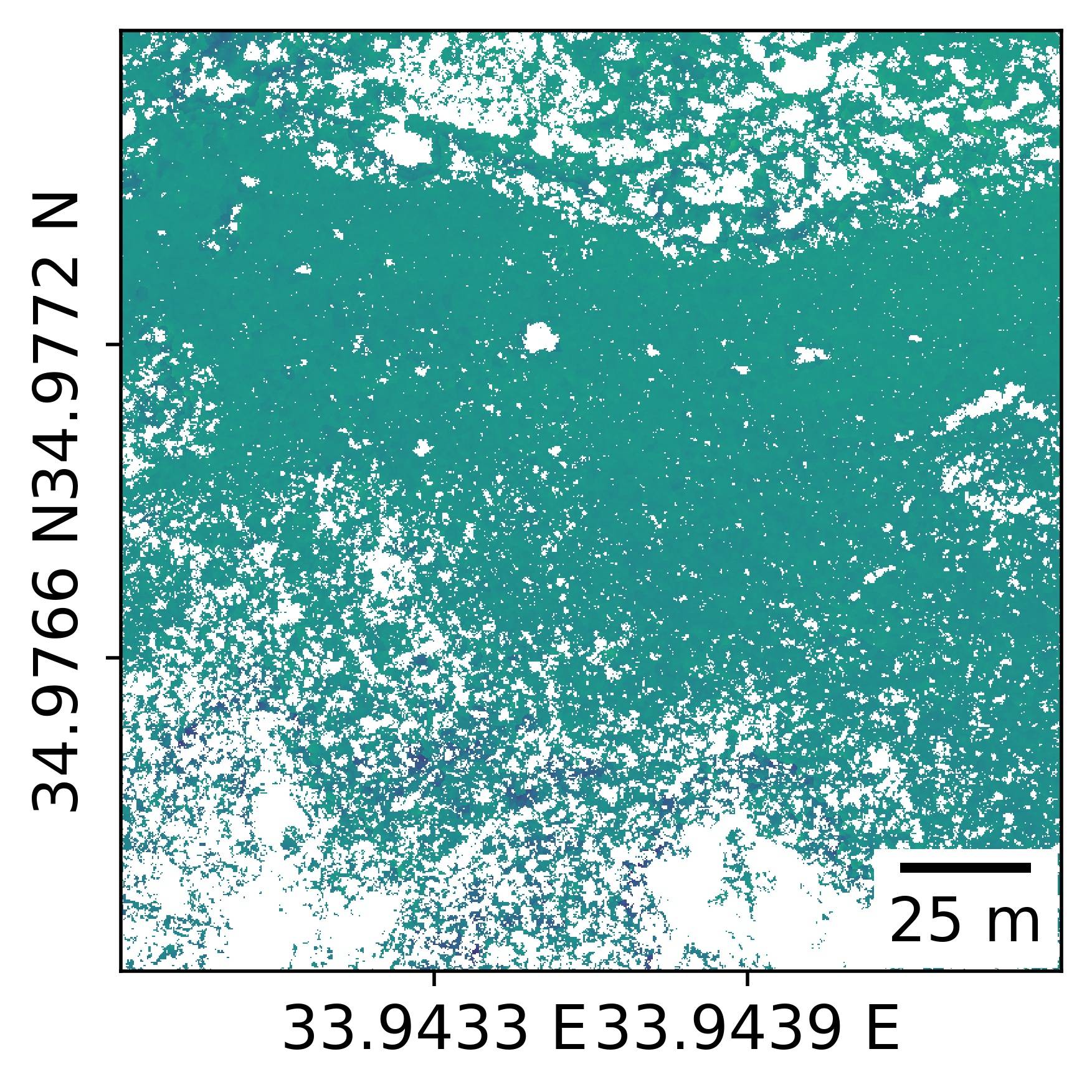}
    \end{minipage}&    
    \hfill

    \begin{minipage}[c]{0.55\columnwidth}
        \centering
        \includegraphics[width=\linewidth]{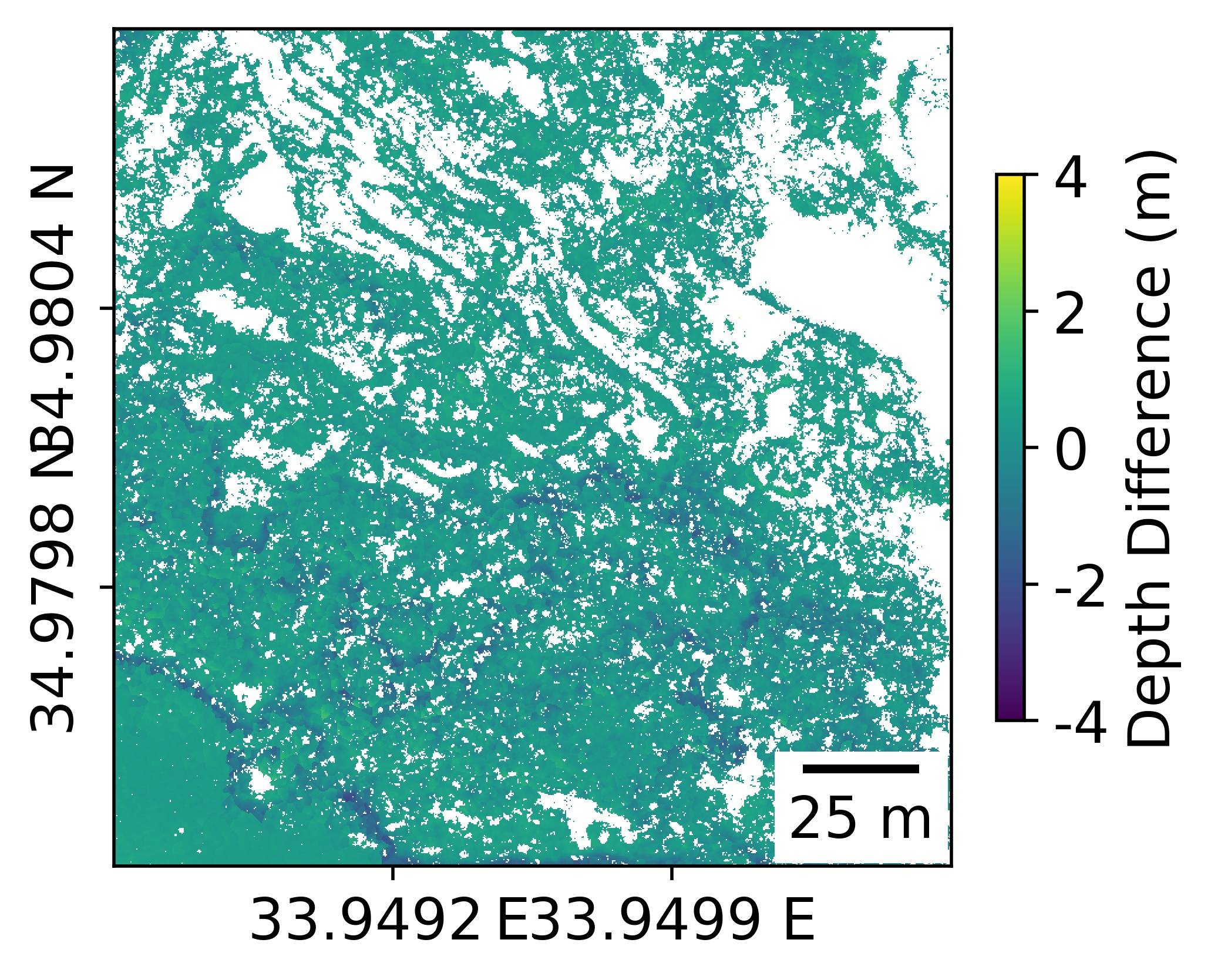}
    \end{minipage}  \vspace{1pt}\\

    \textbf{(c)} &\begin{minipage}[c]{0.44\columnwidth}
        \centering
        \includegraphics[width=\linewidth]{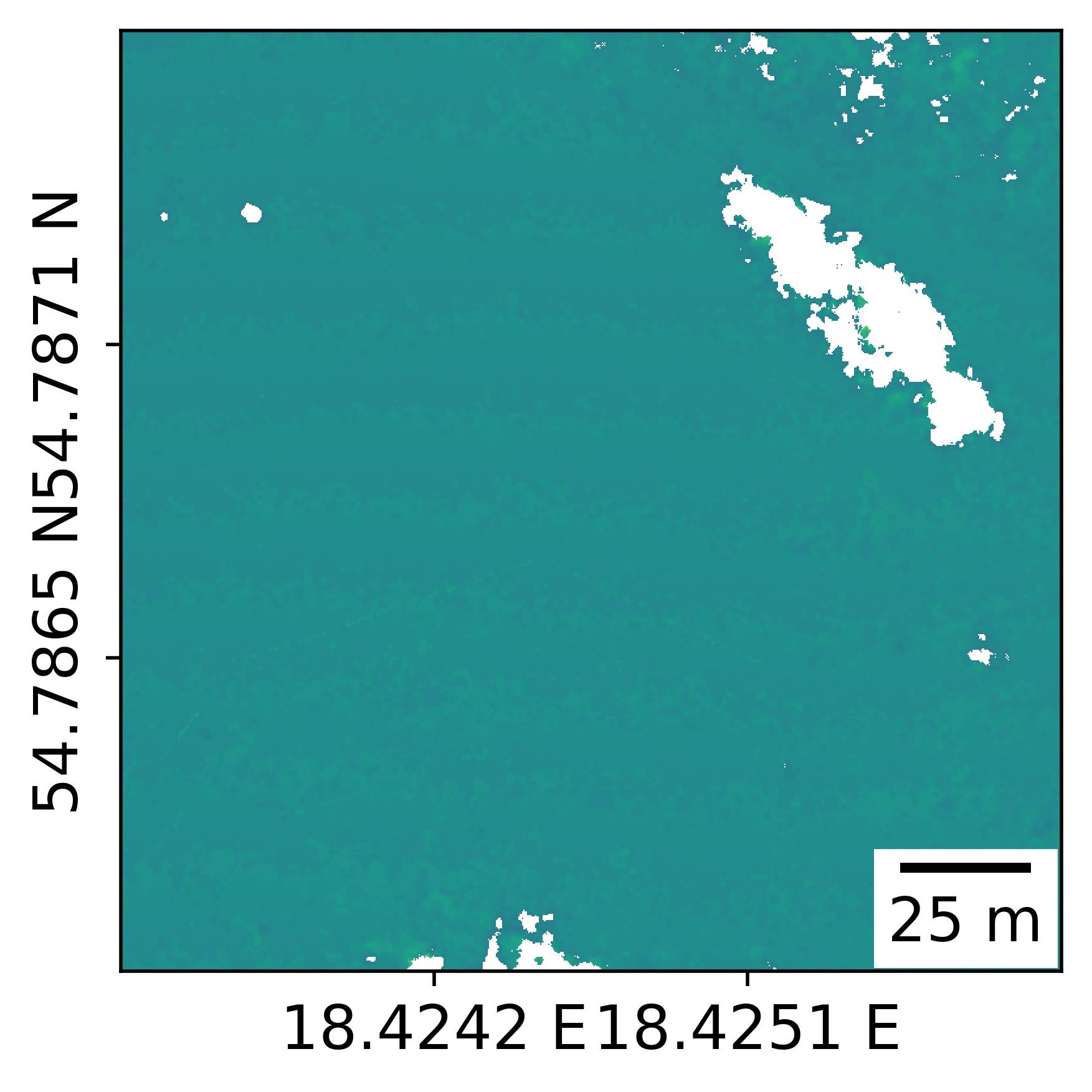}
    \end{minipage}& 
    \hfill

    \begin{minipage}[c]{0.44\columnwidth}
        \centering
        \includegraphics[width=\linewidth]{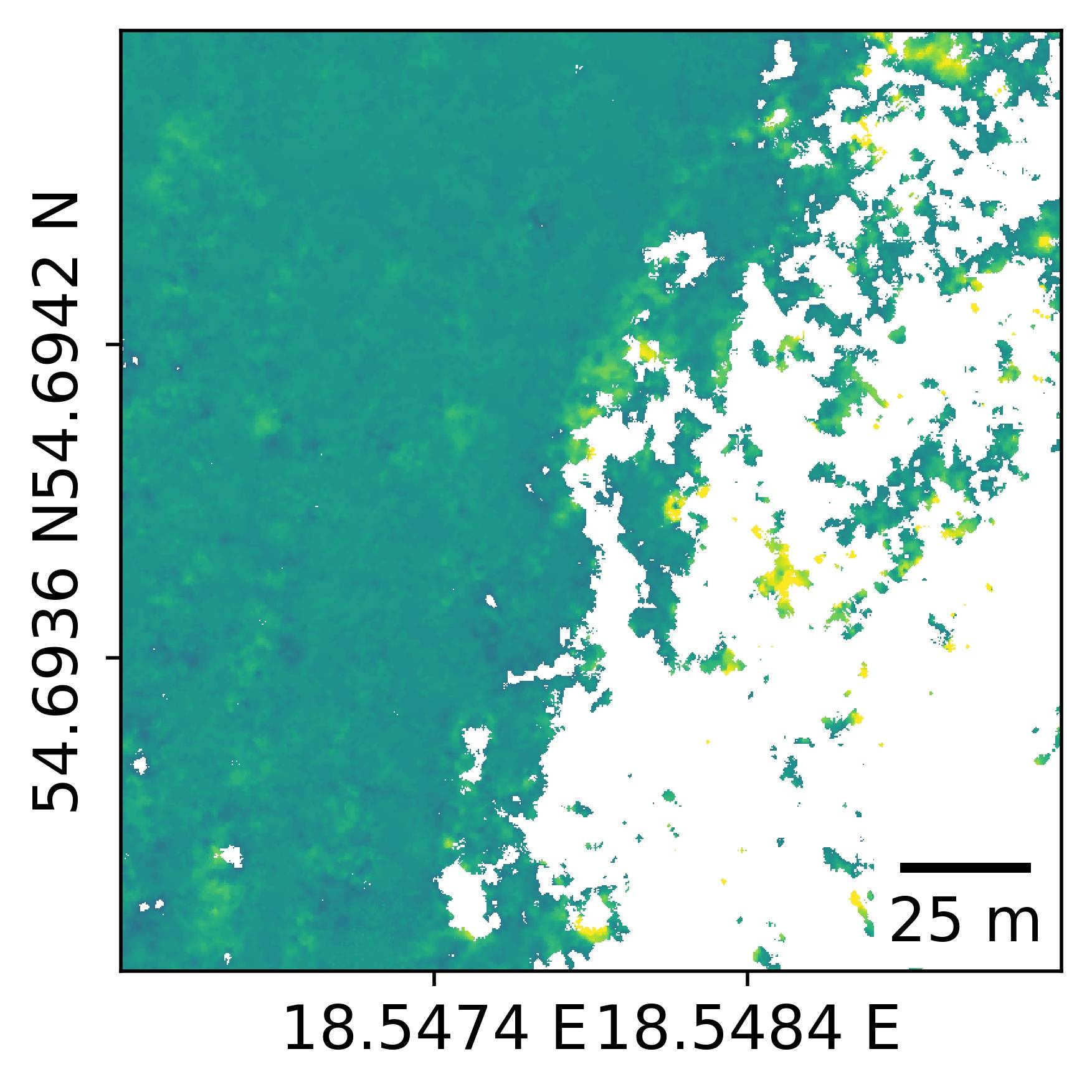}
    \end{minipage}&    
    \hfill

    \begin{minipage}[c]{0.55\columnwidth}
        \centering
        \includegraphics[width=\linewidth]{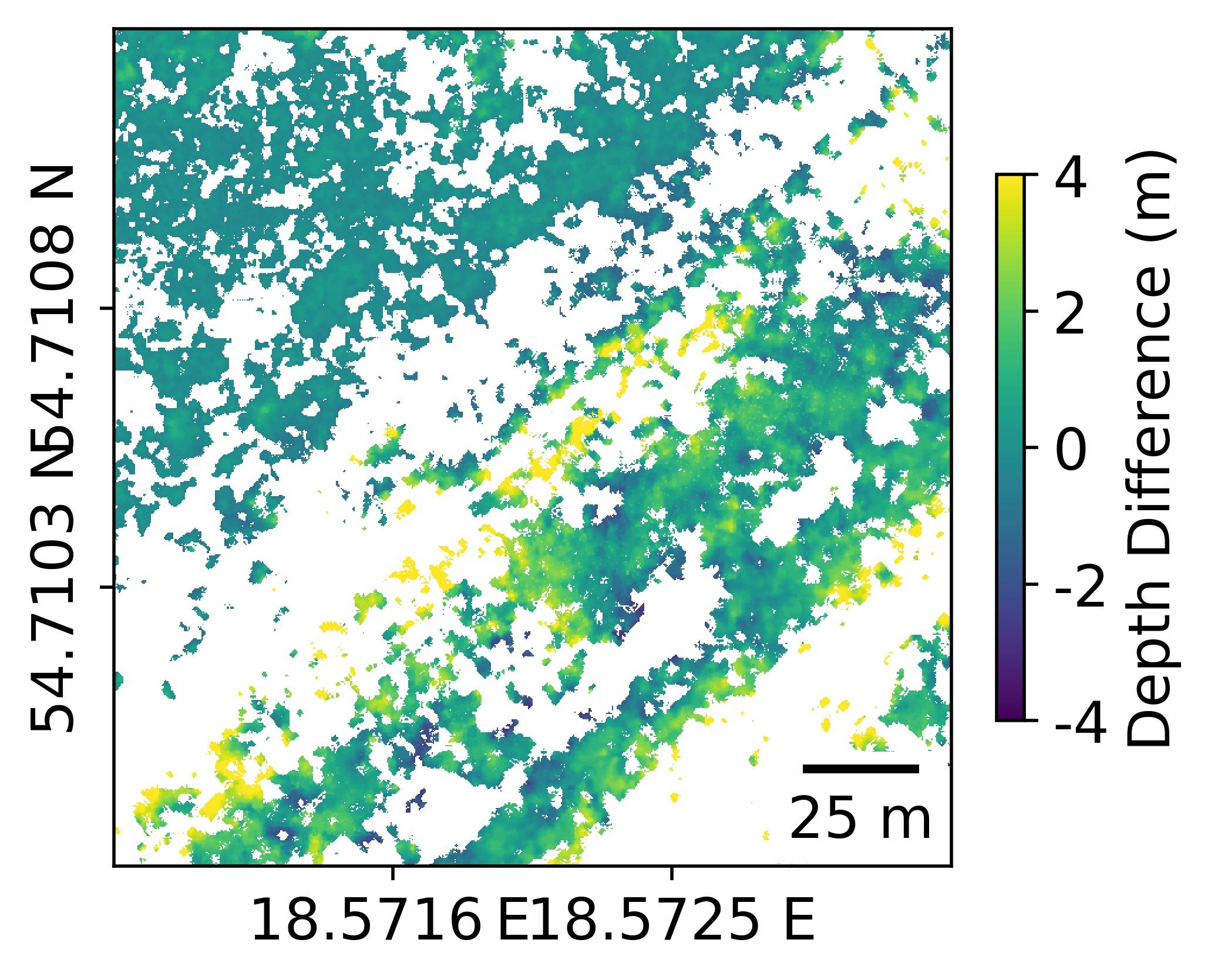}
    \end{minipage}  \vspace{1pt}\\

       \textbf{(d)} &\begin{minipage}[c]{0.44\columnwidth}
        \centering
        \includegraphics[width=\linewidth]{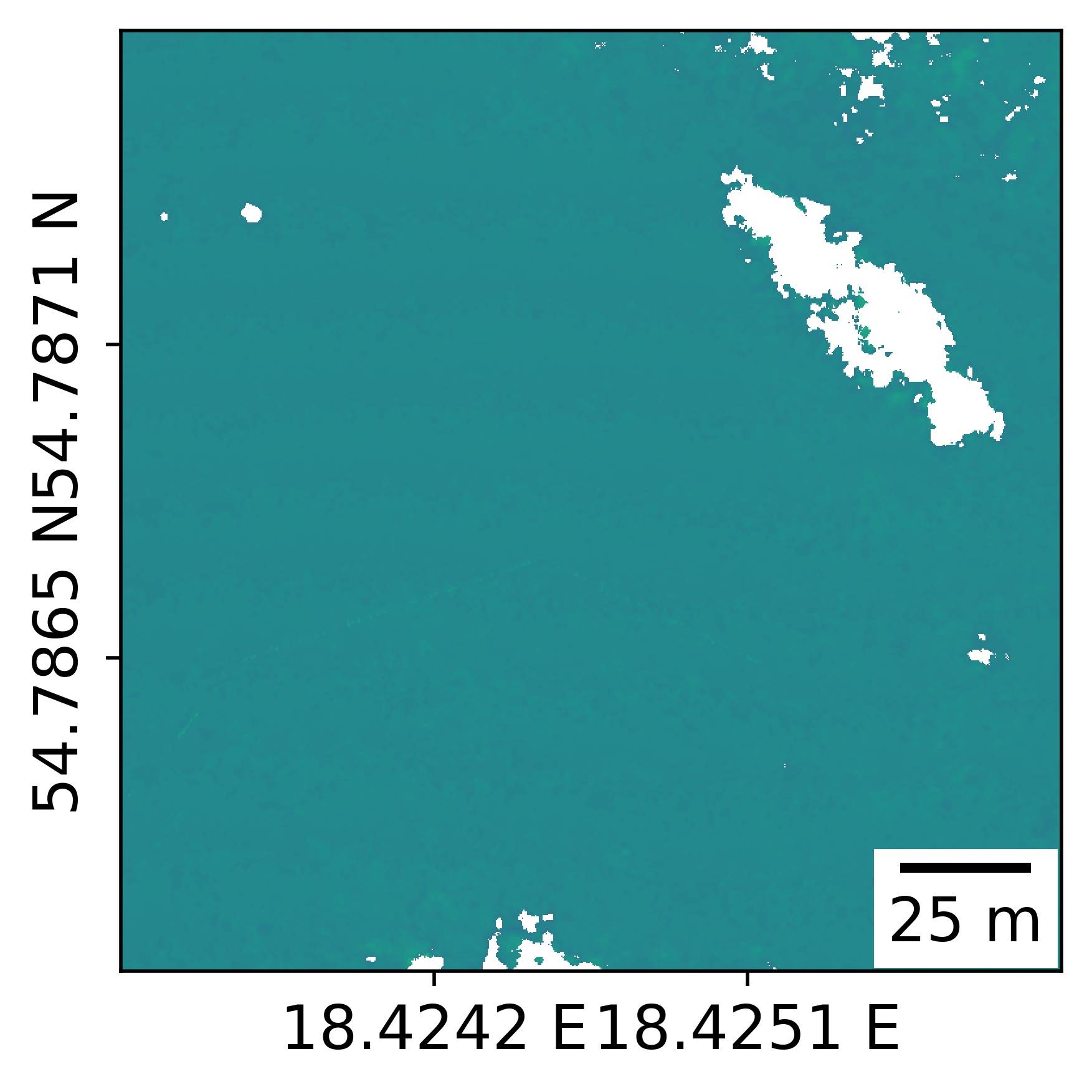}
    \end{minipage}& 
    \hfill

    \begin{minipage}[c]{0.44\columnwidth}
        \centering
        \includegraphics[width=\linewidth]{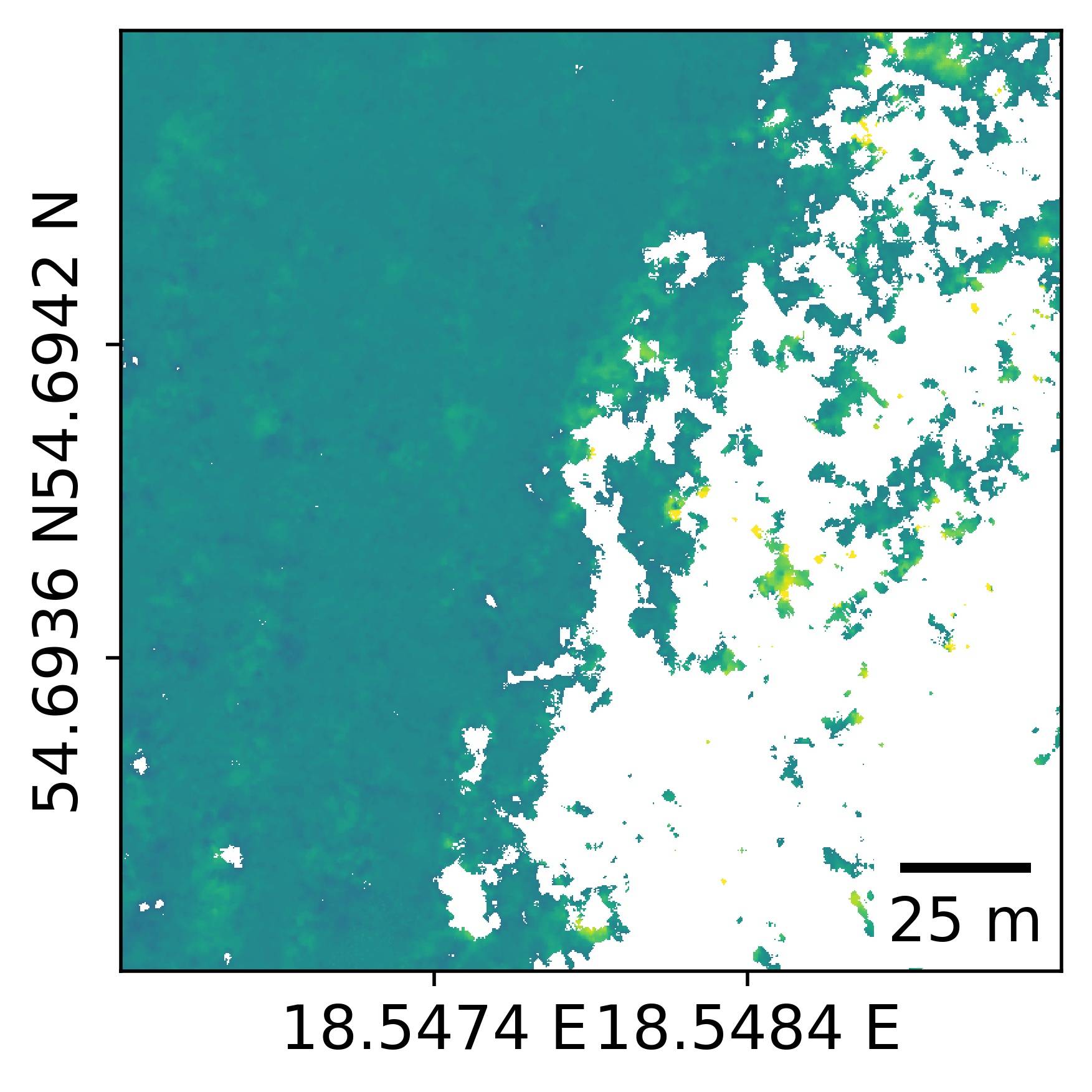}
    \end{minipage}&    
    \hfill

    \begin{minipage}[c]{0.55\columnwidth}
        \centering
        \includegraphics[width=\linewidth]{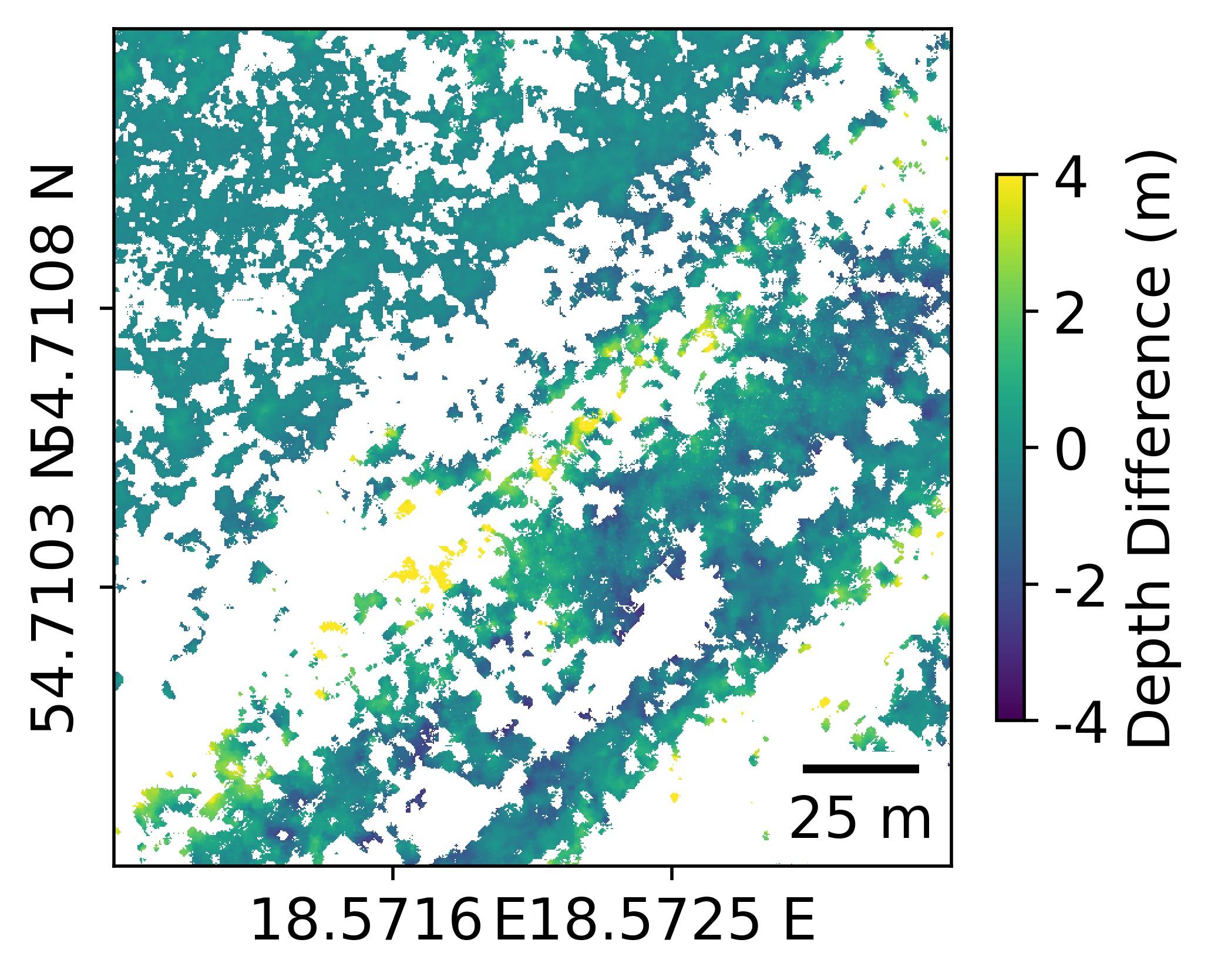}
    \end{minipage}  \vspace{1pt}\\
\end{tabular}
\vspace{-0.07in}
\caption{(a) SfM-MVS bathymetry patches of Agia Napa area with refraction, (b) after refraction correction, (c) SfM-MVS bathymetry patches of Puck Lagoon area with refraction, and (d) after refraction correction. White areas are the data gaps.}

\label{fig:fig8}
\vspace{-0.07in}
\end{figure*}

This subsection provides valuable insights into the accuracy of the bathymetric data used for training in this work, derived from the incomplete refraction-corrected SfM-MVS DSM. Figure \ref{fig:fig8} illustrates the vertical discrepancies between the reference data, acquired via LiDAR/MBES systems, and the bathymetry produced by the SfM-MVS method, before and after refraction correction, across three representative patches in the Agia Napa and Puck Lagoon regions.

\begin{figure*}[bt!]
\vspace{-0.05in}
  \setlength{\tabcolsep}{1.5pt}
  \renewcommand{\arraystretch}{1}
  \footnotesize
  \centering
  \begin{tabular}{cc}
    \begin{minipage}[c]{1\columnwidth}
        \centering
        \includegraphics[width=1\linewidth]{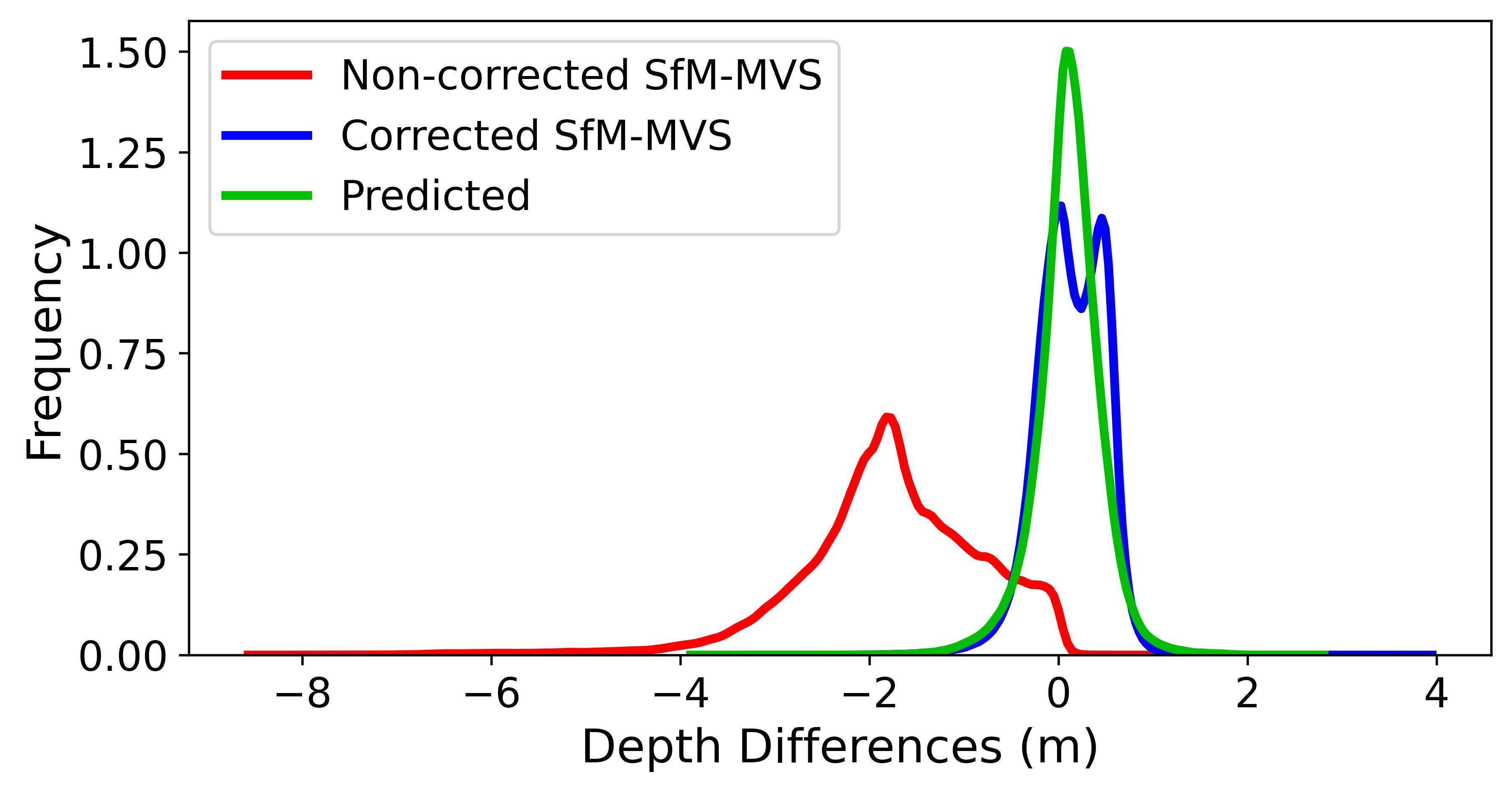}\\
        \textbf{(a)}
    \end{minipage} &
    \hfill
    \begin{minipage}[c]{1\columnwidth}
        \centering
        \includegraphics[width=\linewidth]{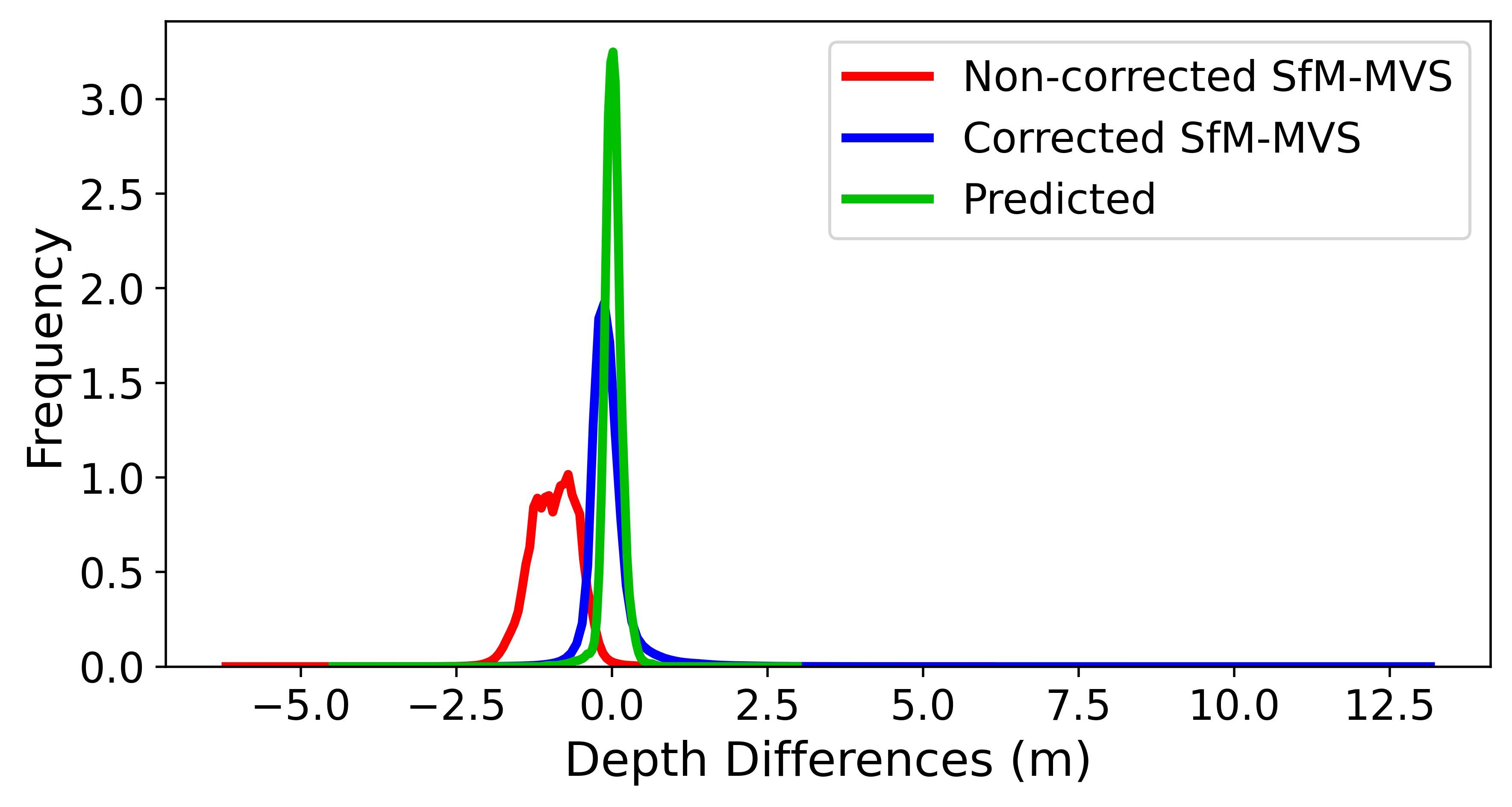}\\
        \textbf{(b)}
    \end{minipage}
  \end{tabular}
  
  \caption{Normalized histograms of depth differences to the reference data for (a) Agia Napa and (b) Puck Lagoon areas.}
  \label{fig:11}
  \vspace{-0.05in}
  \vfill
\end{figure*}

For Agia Napa area, Figure \ref{fig:fig8}a shows that before refraction correction, the discrepancy of SfM-MVS depths and reference data increases with depth (for actual depths see Figure \ref{fig:fig1}). As depth increases, the depth errors become more pronounced, shifting from green to blue on the colormap, indicating larger deviations, reaching up to 4m. However, Figure \ref{fig:fig8}b shows that the refraction corrected depths align more closely with the reference data, and the colormap predominantly shows values around 0m. The remaining green and blue areas, mostly observed in regions with abrupt depth changes, represent noise in the DSM. This noise is caused by erroneous matching in deeper, homogeneous parts of the seabed or by inconsistencies between the reference data and the SfM-MVS data. As shown in Table \ref{table:table1}, before refraction correction, the RMSE is 1.96m, the MAE is 1.86m, and the Std. is 0.59m, highlighting significant inaccuracies in the SfM-MVS depths at this stage. After applying the refraction correction method, these metrics are significantly reduced, with the RMSE, MAE, and Std. decreasing by 80.7\%, 83.3\%, and 54.1\%, respectively. The smaller reduction in Std. compared to RMSE and MAE indicates that outlier depths still contribute to a greater spread of values. This suggests that the correction method effectively reduces systematic errors (captured by RMSE and MAE) but is less effective at addressing random errors and inconsistencies in the depth data. This residual noise from the SfM-MVS process, caused by erroneous matches, will be corrected by the Swin-BathyUNet in the next step. The improvement is also evident in Table\ref{table:table1.2} which provides additional metrics. Before refraction correction, 90.64\% of the DSM pixels exhibit absolute depth differences greater than 1m from the reference depths. After applying refraction correction, this figure decreases significantly to just 1.78\%. Similarly, 89.27\% of the pixels initially show depth differences exceeding 0.50m, but this percentage drops to 19.28\% after refraction correction. The horizontal shift of the difference histogram towards and around 0, following the refraction correction step, is clearly visible in Figure \ref{fig:11}a. There, differences from the non-corrected SfM-MVS depths are shown in red, while differences from the corrected SfM-MVS depths are depicted in blue. The predicted depths are represented in green, and these will be discussed in the following sections.

For Puck Lagoon area, Figure \ref{fig:fig8}c illustrates that prior to refraction correction, depth discrepancies increase with depth, as in Agia Napa (for actual depths see Figure \ref{fig:fig2}). As shown in Table \ref{table:table2}, the average RMSE is 1.01m, the MAE is 0.93m, and the Std. is 0.39m, indicating considerable and systematic inaccuracies in the SfM-MVS depth estimates. After applying the refraction correction method (Figure \ref{fig:fig8}d),  these metrics are significantly reduced by 87.57\%, 88.23\%, and 80.46\%, respectively. This improvement is also evident in Table\ref{table:table2.2}. Being a much shallower area, before refraction correction, 44.25\% of the DSM pixels exhibit absolute depth differences $>$ 1m and 90\% of the pixels initially show depth differences $>$ 0.50m from the reference depths. After applying refraction correction, these figures decrease by 95.68\% and 92.28\% respectively. The horizontal shift of the differences histogram toward and around 0, following the refraction correction step, is clearly visible in Figure \ref{fig:11}b.  However, despite this correction, the middle and right patches in Figure \ref{fig:fig8}d still exhibit noise, highlighted in reddish, greenish, and bluish colors, due to the high variability in the point clouds. This noise, which became worse after refraction correction, arises from the extensive homogeneous seabed, predominantly composed of sand, and the characteristics of the water column. Nevertheless, this noise will be corrected by the Swin-BathyUNet in subsequent steps.

\begin{figure*}[h!]
  \setlength{\tabcolsep}{1.5pt}
  \renewcommand{\arraystretch}{1}
  \footnotesize
\centering
  \begin{tabular}{m{0.5cm}ccc}
   \textbf{(a)} &\begin{minipage}[c]{0.44\columnwidth}
        \centering
        \includegraphics[width=\linewidth]{img/depth_410.jpg}
    \end{minipage}& 
    \hfill

    \begin{minipage}[c]{0.44\columnwidth}
        \centering
        \includegraphics[width=\linewidth]{img/depth_379.jpg}
    \end{minipage}&    
    \hfill

    \begin{minipage}[c]{0.55\columnwidth}
        \centering
        \includegraphics[width=\linewidth]{img/depth_409.jpg}
    \end{minipage}  \vspace{1pt}\\

    \textbf{(b)} & \begin{minipage}[c]{0.44\columnwidth}
        \centering
        \includegraphics[width=\linewidth]{img/sfm_410.jpg}
    \end{minipage}&

    \begin{minipage}[c]{0.44\columnwidth}
        \centering
        \includegraphics[width=\linewidth]{img/sfm_379.jpg}
    \end{minipage}&

    \begin{minipage}[c]{0.55\columnwidth}
        \centering 
        \includegraphics[width=\linewidth]{img/sfm_409.jpg}
    \end{minipage}\\

       \textbf{(c)} &\begin{minipage}[c]{0.44\columnwidth}
        \centering
        \includegraphics[width=\linewidth]{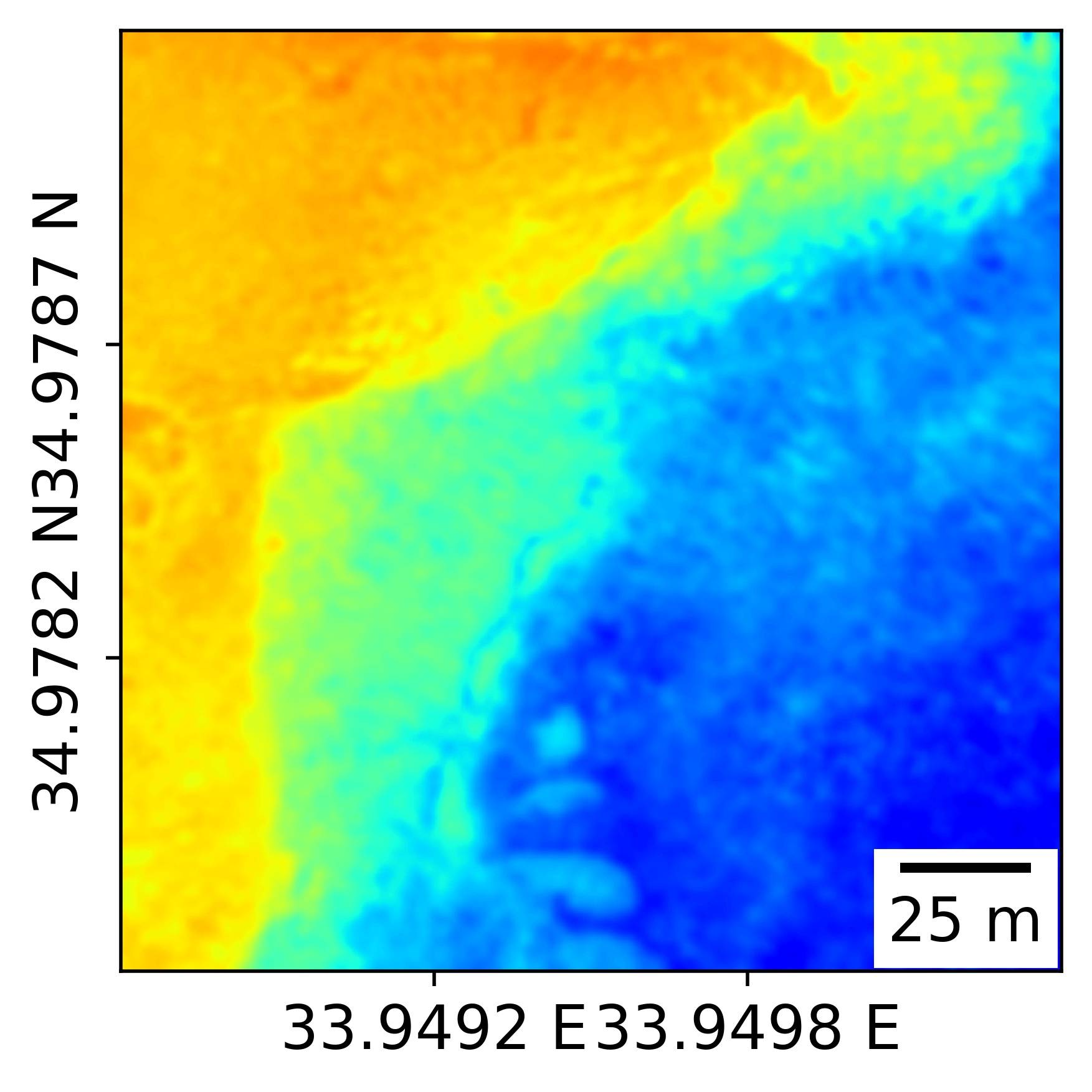}
    \end{minipage}& 
    \hfill

    \begin{minipage}[c]{0.44\columnwidth}
        \centering
        \includegraphics[width=\linewidth]{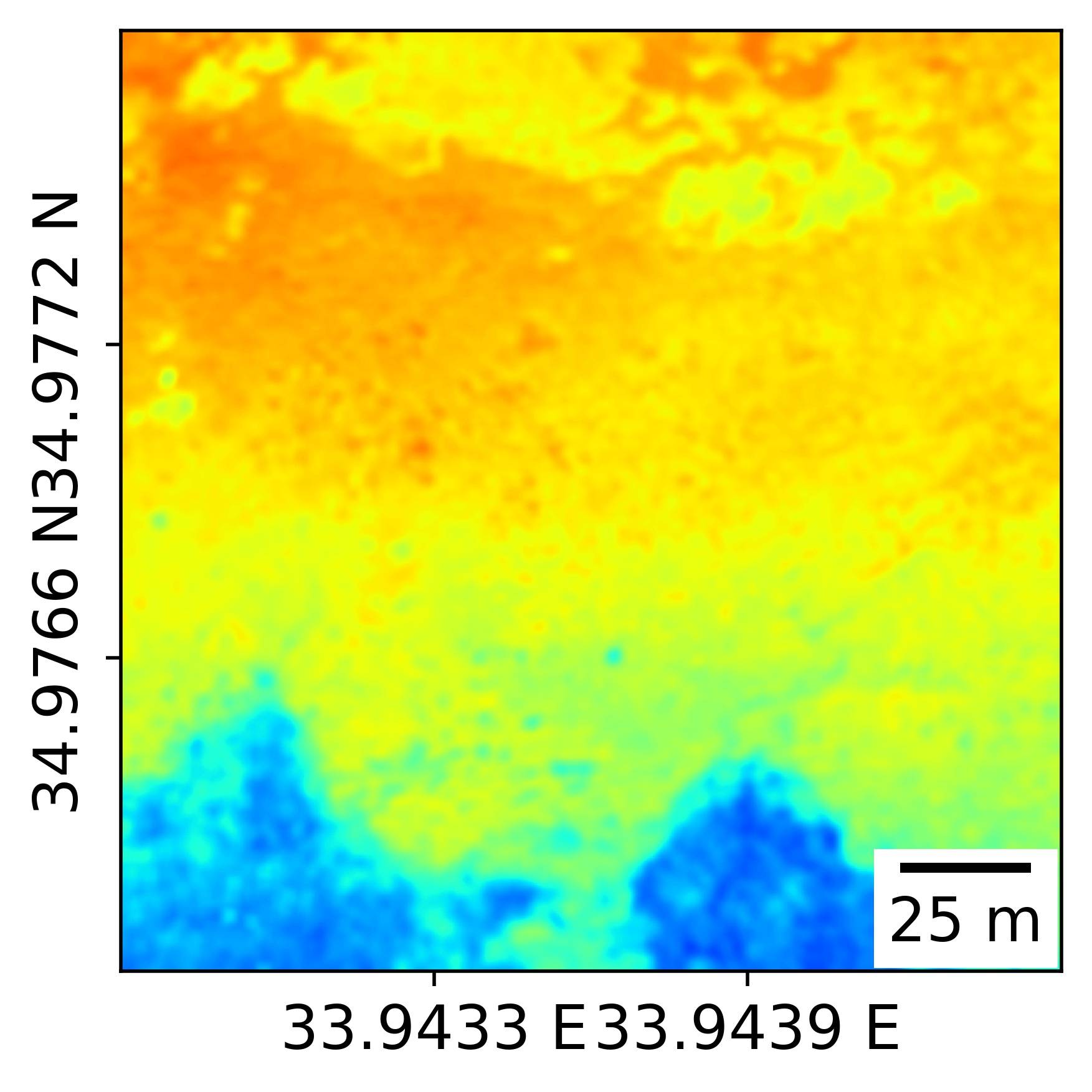}
    \end{minipage}&    
    \hfill

    \begin{minipage}[c]{0.55\columnwidth}
        \centering
        \includegraphics[width=\linewidth]{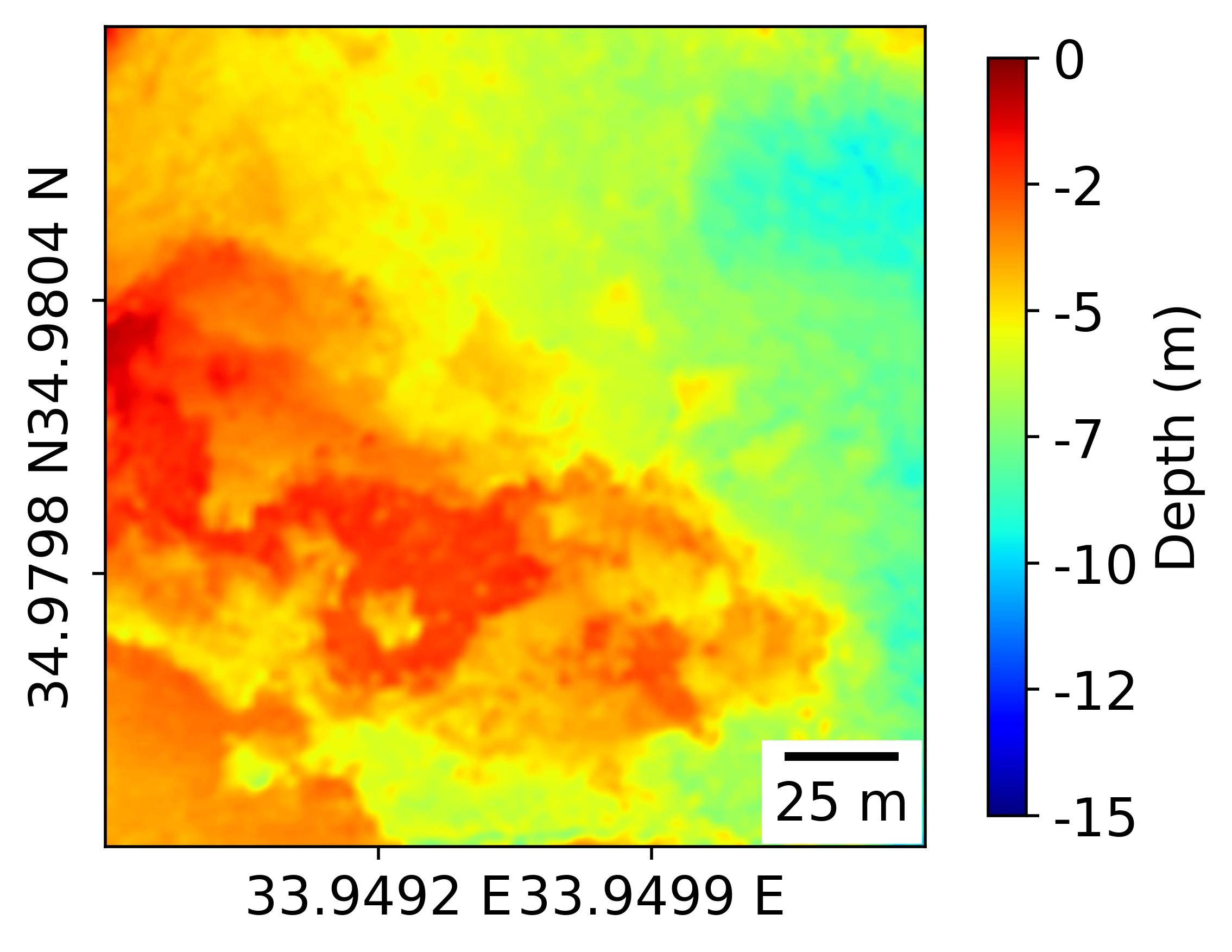}
    \end{minipage}  \vspace{1pt}\\
\end{tabular}
\vspace{-0.07in}
\caption{Example patches of Agia Napa depicting (a) reference bathymetry, (b) refraction-corrected SfM-MVS bathymetry used for training, and (c) Swin-BathyUNet predictions, relative to WGS ’84. White areas indicate data gaps.}
\label{fig:fig9}
\vspace{-0.07in}
\end{figure*}

\begin{figure*}[h!]
  \setlength{\tabcolsep}{1.5pt}
  \renewcommand{\arraystretch}{1}
  \footnotesize
  
\centering

  \begin{tabular}{m{0.5cm}ccc}
   \textbf{(a)} &\begin{minipage}[c]{0.44\columnwidth}
        \centering
        \includegraphics[width=\linewidth]{img/depth_1.jpg}
    \end{minipage}& 
    \hfill

    \begin{minipage}[c]{0.44\columnwidth}
        \centering
        \includegraphics[width=\linewidth]{img/depth_2873.jpg}
    \end{minipage}&    
    \hfill

    \begin{minipage}[c]{0.54\columnwidth}
        \centering
        \includegraphics[width=\linewidth]{img/depth_3164.jpg}
    \end{minipage}  \vspace{1pt}\\

      \textbf{(b)} &   \begin{minipage}[c]{0.44\columnwidth}
        \centering
        \includegraphics[width=\linewidth]{img/sfm_1.jpg}
    \end{minipage}& 
    \hfill

    \begin{minipage}[c]{0.44\columnwidth}
        \centering
        \includegraphics[width=\linewidth]{img/sfm_2873.jpg}
    \end{minipage}&    
    \hfill

    \begin{minipage}[c]{0.54\columnwidth}
        \centering
        \includegraphics[width=\linewidth]{img/sfm_3164.jpg}
    \end{minipage} \vspace{1pt}\\

       \textbf{(c)} &\begin{minipage}[c]{0.44\columnwidth}
        \centering
        \includegraphics[width=\linewidth]{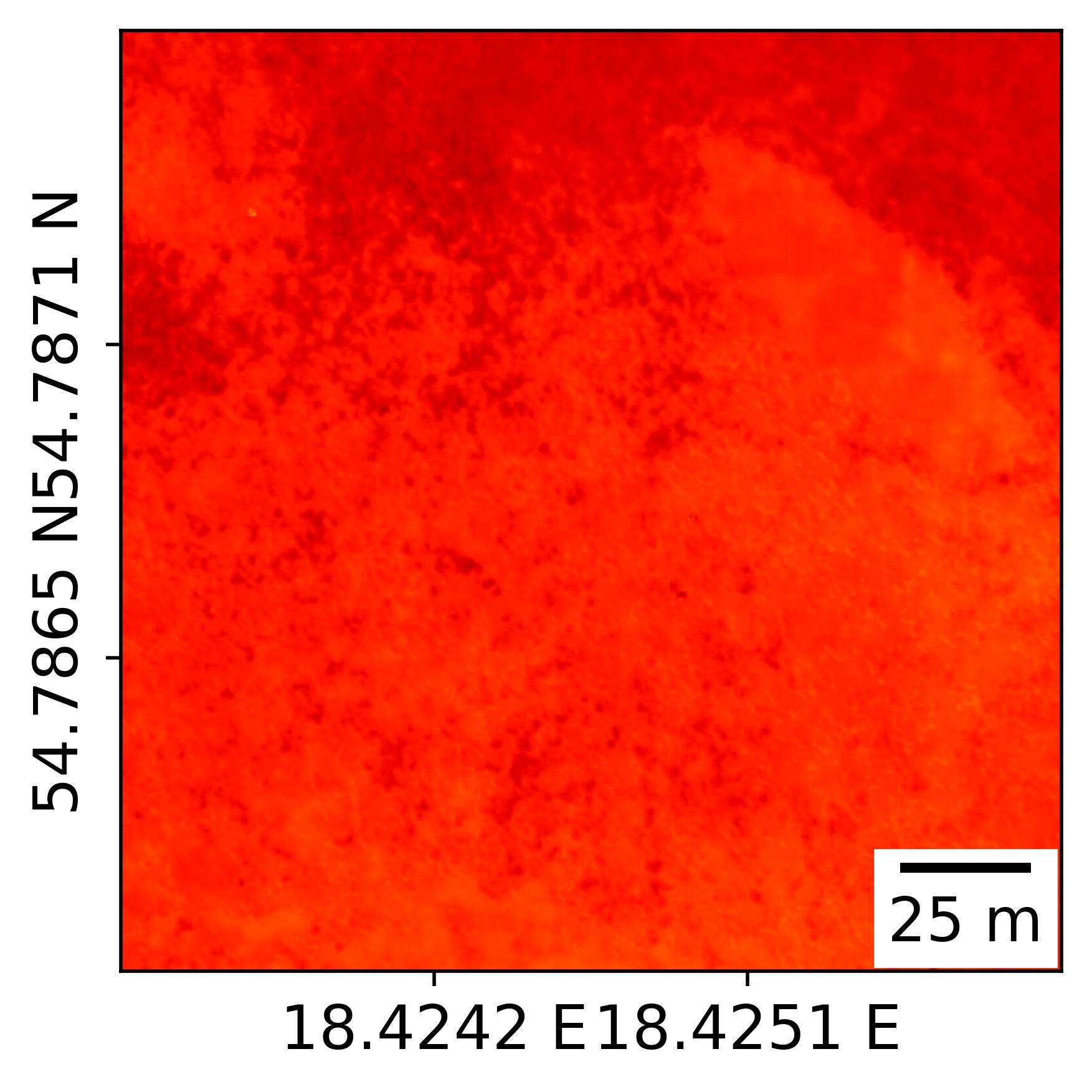}
    \end{minipage}& 
    \hfill

    \begin{minipage}[c]{0.44\columnwidth}
        \centering
        \includegraphics[width=\linewidth]{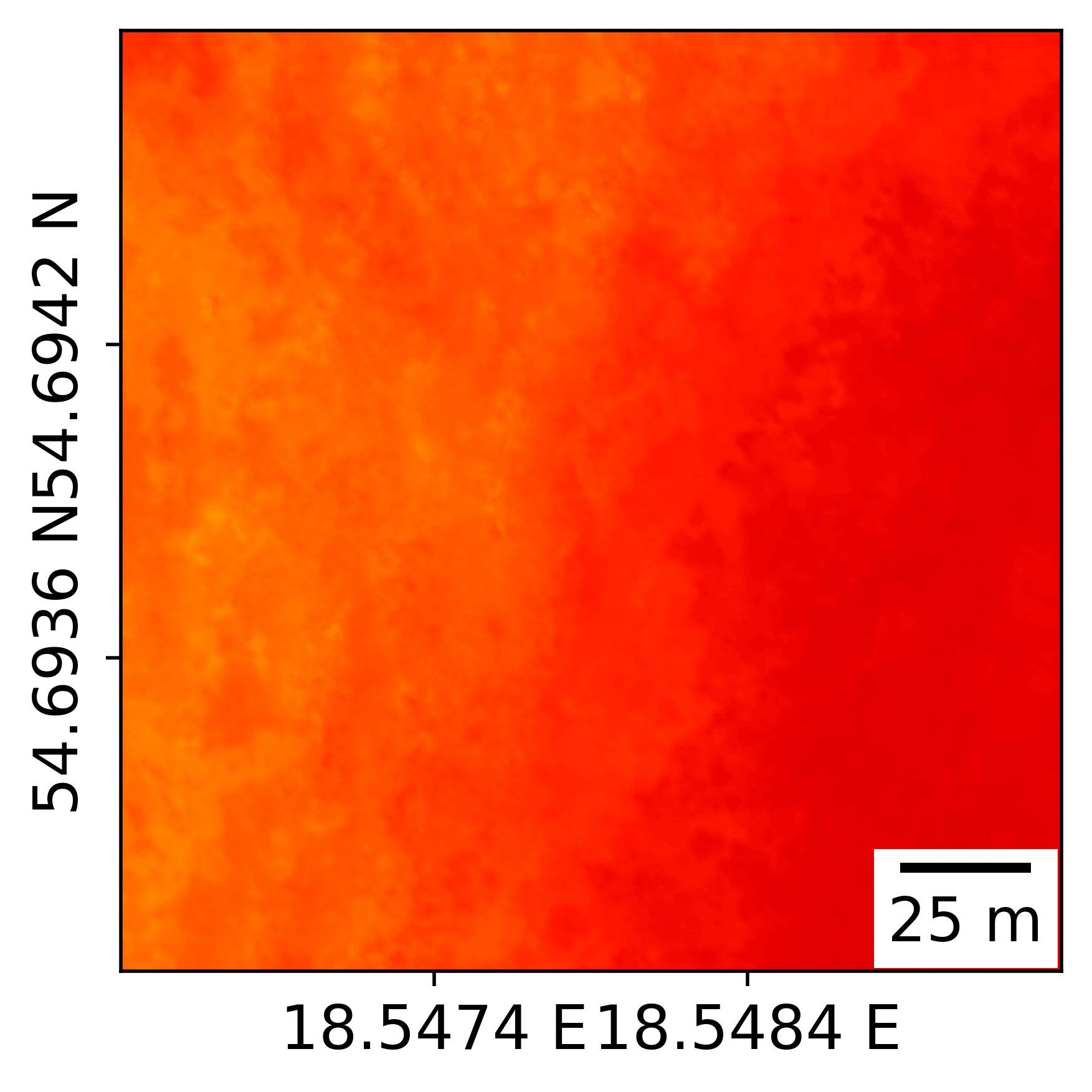}
    \end{minipage}&    
    \hfill

    \begin{minipage}[c]{0.54\columnwidth}
        \centering
        \includegraphics[width=\linewidth]{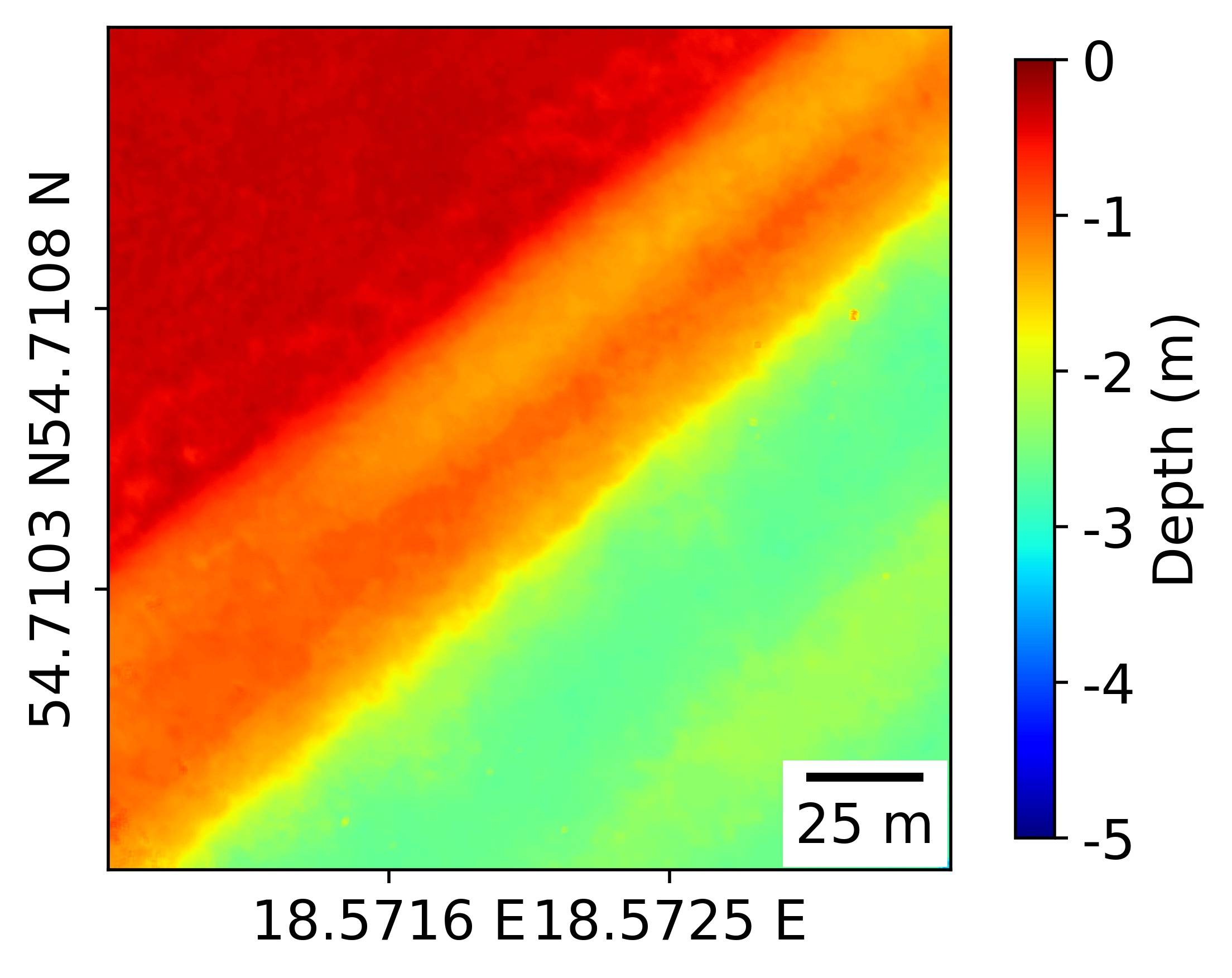}
    \end{minipage}  \vspace{1pt}
\end{tabular}
\vspace{-0.07in}
\caption{Example patches of Puck Lagoon depicting (a) reference bathymetry, (b) refraction-corrected SfM-MVS bathymetry used for training, and (c) Swin-BathyUNet predictions, relative to WGS ’84. White areas indicate data gaps.}
\label{fig:fig10}
\vspace{-0.07in}
\end{figure*}

\begin{table*}[h!]
\caption{Average$_{3}$ testing performance obtained on Agia Napa area using the proposed Swin-BathyUNet and baseline model U-Net \citep{unet}. Entities in bold indicate the best score.}
\centering
\begin{tabularx}{\textwidth}{l>{\centering\arraybackslash}X>{\centering\arraybackslash}X>{\centering\arraybackslash}X>{\centering\arraybackslash}X>{\centering\arraybackslash}X>{\centering\arraybackslash}X>{\centering\arraybackslash}X>{\centering\arraybackslash}X>{\centering\arraybackslash}X}
\toprule
\multirow{2}{*}{Depth Data} & \multicolumn{3}{c}{Swin-BathyUNet + BSW loss} & \multicolumn{3}{c}{Swin-BathyUNet + RMSE loss} & \multicolumn{3}{c}{U-Net + RMSE loss} \\
\cmidrule{2-10}
                   & RMSE(m) & MAE(m) & Std.(m) & RMSE(m) & MAE(m) & Std.(m) & RMSE(m) & MAE(m) & Std.(m)\\
\midrule
SfM-MVS  & 1.96    & 1.86   & 0.59   & 1.96    & 1.86   & 0.59 & 1.96    & 1.86   & 0.59   \\
Corrected SfM-MVS  & 0.38    & 0.31   & 0.27   & 0.38    & 0.31   & 0.27 & 0.38    & 0.31   & 0.27 \\
\midrule
Predicted "non-gaps"  & \textbf{0.38} &  \textbf{0.29} & \textbf{0.32}  & 0.43    & 0.34   & 0.34 & 0.57    & 0.47   & 0.40   \\
Predicted "gaps"  & \textbf{0.58}    & \textbf{0.43}    & \textbf{0.49}  & 0.61   & 0.47   & 0.52 & 0.76    & 0.61   & 0.58   \\
Combined Prediction   &  \textbf{0.52}    & \textbf{0.38}    & \textbf{0.44} & 0.53    & 0.40   & 0.46 & 0.60    & 0.45   & 0.51\\
Whole prediction      & \textbf{0.49}     & \textbf{0.37}   & \textbf{0.44}   & 0.53  & 0.40   & 0.46 & 0.67    & 0.53   & 0.52  \\
\bottomrule
\end{tabularx}
\label{table:table1}
\end{table*}

\begin{table*}[h!]
\caption{Additional metrics for Agia Napa. For SfM-MVS and Corrected SfM-MVS are computed over "non-gaps" only. Entities in bold indicate the best score.}
\centering
\begin{tabularx}{\textwidth}{l>{\centering\arraybackslash}m{15mm}>{\centering\arraybackslash}X>{\centering\arraybackslash}X>{\centering\arraybackslash}X>{\centering\arraybackslash}X}
\toprule
Metric & SfM-MVS & Corrected SfM-MVS & Swin-BathyUNet + BSW loss & Swin-BathyUNet + RMSE loss & U-Net + RMSE loss\\
\midrule
$\lvert Depth Error\lvert >$ 1m  & 90.64\%   &1.78\%   & \textbf{1.62\%}  &  2.75\%   &  7.36\%  \\
$\lvert Depth Error\lvert >$ 0.50m   &  89.27\%   & 19.28\%   & \textbf{9.29\%}   &  13.52\%   & 18.02\%   \\
Coverage  & 56.49\% & 56.49\% & 100\% & 100\% & 100\%   \\
\bottomrule
\end{tabularx}
\label{table:table1.2}
\end{table*}

\begin{table*}[t]
\caption{Average$_{3}$ testing performance obtained on Puck Lagoon area using the proposed Swin-BathyUNet and baseline model U-Net \citep{unet}. Entities in bold indicate the best score.}
\centering
\begin{tabularx}{\textwidth}{l>{\centering\arraybackslash}X>{\centering\arraybackslash}X>{\centering\arraybackslash}X>{\centering\arraybackslash}X>{\centering\arraybackslash}X>{\centering\arraybackslash}X>{\centering\arraybackslash}X>{\centering\arraybackslash}X>{\centering\arraybackslash}X}
\toprule
\multirow{2}{*}{Depth Data} & \multicolumn{3}{c}{Swin-BathyUNet + BSW loss} & \multicolumn{3}{c}{Swin-BathyUNet + RMSE loss} & \multicolumn{3}{c}{U-Net + RMSE loss} \\
\cmidrule{2-10}
           & RMSE(m) & MAE(m) & Std.(m) & RMSE(m) & MAE(m) & Std.(m) & RMSE(m) & MAE(m) & Std.(m)\\
\midrule
SfM-MVS    & 1.01    & 0.93   & 0.39    & 1.01    & 0.93   & 0.39  & 1.01    & 0.93   & 0.39    \\
Corrected SfM-MVS    &  0.13   & 0.11   &  0.08    &  0.13   & 0.11   &  0.087 &  0.13   & 0.11   &  0.08   \\
\midrule
Predicted "non-gaps"            & \textbf{0.15}    & \textbf{0.10}     & \textbf{0.15}    & 0.23  & 0.16  & 0.22  & 0.22 & 0.15 & 0.22   \\
Predicted "gaps"                & \textbf{0.20}    & \textbf{0.13}     &  \textbf{0.20}   & 0.30  & 0.19  & 0.29  & 0.31 & 0.19  & 0.31 \\
Combined Prediction                  & \textbf{0.28}    &  \textbf{0.08}    & \textbf{0.28}    & 0.29  & 0.08 &  0.29  & 0.29 & 0.08  & 0.29\\
Whole prediction                     & \textbf{0.16}    &  \textbf{0.11}    & \textbf{0.16}    & 0.24  & 0.17  & 0.23  & 0.22 & 0.15 & 0.22 \\
\bottomrule

\end{tabularx}
\label{table:table2}
\end{table*}

\begin{table*}[h!]
\caption{Additional metrics for Puck Lagoon. For SfM-MVS and Corrected SfM-MVS are computed over "non-gaps" only. Entities in bold indicate the best score.}
\centering
\begin{tabularx}{\textwidth}{l>{\centering\arraybackslash}m{15mm}>{\centering\arraybackslash}X>{\centering\arraybackslash}X>{\centering\arraybackslash}X>{\centering\arraybackslash}X}
\toprule
Metric & SfM-MVS & Corrected SfM-MVS & Swin-BathyUNet + BSW loss & Swin-BathyUNet + RMSE loss & U-Net + RMSE loss\\
\midrule
$\lvert Depth Error\lvert >$ 1m  & 44.25\%   & 1.91\%   & \textbf{0.09\%}  &  0.38\%   &  0.41\%  \\
$\lvert Depth Error\lvert >$ 0.50m   &  90.00\%   & 6.95\%   & \textbf{1.62\%}   &  5.06\%   & 3.90\%   \\
Coverage  & 87.34\% & 87.34\% & 100\% & 100\% & 100\%   \\
\bottomrule
\end{tabularx}
\label{table:table2.2}
\end{table*}

The refraction correction results at both sites demonstrate substantial improvement, indicating the enhanced accuracy and reliability of the training SfM-MVS depths after the refraction correction step. These improvements are critical to ensure the precision of the SDB models trained with these data. For the remainder of the article, "non-gap" areas are those where bathymetry is already provided by the SfM-MVS, while the "gap" areas, already mentioned, refer to the regions with missing SfM-MVS depths. "Whole prediction" refers to the direct output of Swin-BathyUNet model, while "combined prediction" describes a DSM patch where the "non-gaps" areas remain unchanged, and the data gap areas are filled with the model's predictions.

\subsection{Swin-BathyUNet's performance}
\label{proposedvsunet}
\subsubsection{Comparison with baseline}
The learning-based bathymetry obtained from Swin-BathyUNet is compared with reference data that was not used during the training process. To evaluate the performance of this method, the U-Net architecture was used as a baseline, given its widespread application in learning-based SDB \citep{mandlburger2021bathynet, martinsen2023predicting, al2023satellite, magicbathynet} and other ocean remote sensing tasks \citep{deepblue}. The results of Swin-BathyUNet are presented using both the BSW RMSE loss function and the standard RMSE loss function. This comparison highlights the effectiveness of incorporating boundary information to improve depth estimation accuracy. U-Net is trained only using the standard RMSE loss function; however, the differences to Swin-BathyUNet are so significant that they would not be compensated from just using the proposed BSW RMSE loss. The results are shown in Tables \ref{table:table1} - \ref{table:table2.2}. In "non-gaps" regions, Swin-BathyUNet using the BSW RMSE loss function and the standard RMSE loss function, demonstrates superior performance compared to U-Net, with significantly lower RMSE, MAE, and Std. The lower errors indicate that Swin-BathyUNet predicts the depths more accurately in both test sites. This suggests that the proposed model better captures fine details and structures in regions without "gaps". In "gaps" regions, where the depth predictions are more challenging, the lower RMSE, MAE, and Std. values resulted from Swin-BathyUNet indicate that it still outperforms U-Net, handling these difficult areas more effectively. When considering the whole prediction (both "non-gaps" and "gaps"), Swin-BathyUNet continues to show clear advantages. The lower RMSE, MAE, and Std. values reflect the model's more consistent performance across different terrain types. In contrast, U-Net exhibits higher errors. Overall, Swin-BathyUNet shows a more balanced and accurate prediction capability across both simple and complex areas. In Agia Napa, Swin-BathyUNet with the BSW RMSE loss function achieves a 34.03\% improvement in RMSE for "non-gaps" areas, a 23.83\% improvement in RMSE for "gaps" regions, and a 24.32\% improvement in RMSE for the whole prediction compared to U-Net. In Puck Lagoon, Swin-BathyUNet with the BSW RMSE loss function achieves a 31.05\% improvement in RMSE for "non-gaps" areas, a 35.2\% improvement in RMSE for "gaps" regions, and a 30.8\% improvement in RMSE for the whole prediction compared to U-Net. By comparing the amount of predicted depths having absolute depth error $>$ 1m and $>$ 0.50m in Tables \ref{table:table1.2} and \ref{table:table2.2}, it is obvious that Swin-BathyUNet consistently shows the highest improvement in both regions. For the Agia Napa area, it reduces predictions with absolute depth error $>$ 1m by 78\% compared to U-Net, and those with an error $>$ 0.50m by 48\%. Similarly, in the Puck Lagoon area, Swin-BathyUNet achieves a 78\% and 58\% reduction respectively. These figures demonstrate that Swin-BathyUNet provides substantial improvements in depth prediction performance over the baseline method.

We argue that the main reason behind these improvements over the baseline is that in learning-based bathymetry prediction, precise localization and differentiation of seabed structures are essential, and by nature, Swin-BathyUNet efficiently captures multi-scale intricate contextual dependencies. The self-attention mechanism enables the network to recognize interactions between distant pixels in an image, enabling it to extract global context while preserving local details. Additionally, the cross-attention module allows interaction between low and high-level multi-scale encoder features from various stages, capturing the rich global context which reduces the semantic gap between the encoder and decoder features.

\subsubsection{BSW and RMSE loss function performance}
The BSW loss function consistently outperforms the standard RMSE loss function in all prediction scenarios (see Tables \ref{table:table1} and \ref{table:table2}). In the Agia Napa area, the BSW loss improves the RMSE by 11.48\%, demonstrating more precise depth predictions in "non-gaps" areas compared to the reference depths. In "gaps" regions, the BSW loss achieves a 5.87\% improvement in RMSE, better handling the complex terrain compared to the standard loss function. Finally, the BSW loss provides a 5.08\% improvement in RMSE for the whole prediction, offering increased accuracy for both "gaps" and "non-gaps" regions. Similarly, in the Puck Lagoon area, the BSW loss function consistently outperforms the standard RMSE loss function, delivering improvements across all the compared data, with a minimum improvement of 28.3\% in the RMSE of the whole prediction.

The analysis presented above is further supported by the additional metrics shown in Tables \ref{table:table1.2} and \ref{table:table2.2} which indicate that the use of the BSW loss function leads to a remarkable reduction in the number of predictions with absolute depth errors $>$ 1m and $>$ 0.50m, ranging from 31\% to 76\%. This substantial improvement reinforces the significant advantages of Swin-BathyUNet utilizing the BSW loss function over employing the RMSE loss function across both regions.

\subsubsection{Performance on "non-gaps", "gaps", and whole area}
\begin{figure*}[h!]
  \setlength{\tabcolsep}{1.5pt}
  \renewcommand{\arraystretch}{1}
  \footnotesize
  
\centering

  \begin{tabular}{m{0.5cm}ccc}
    
   \textbf{(a)} &\begin{minipage}[c]{0.44\columnwidth}
        \centering
        \includegraphics[width=\linewidth]{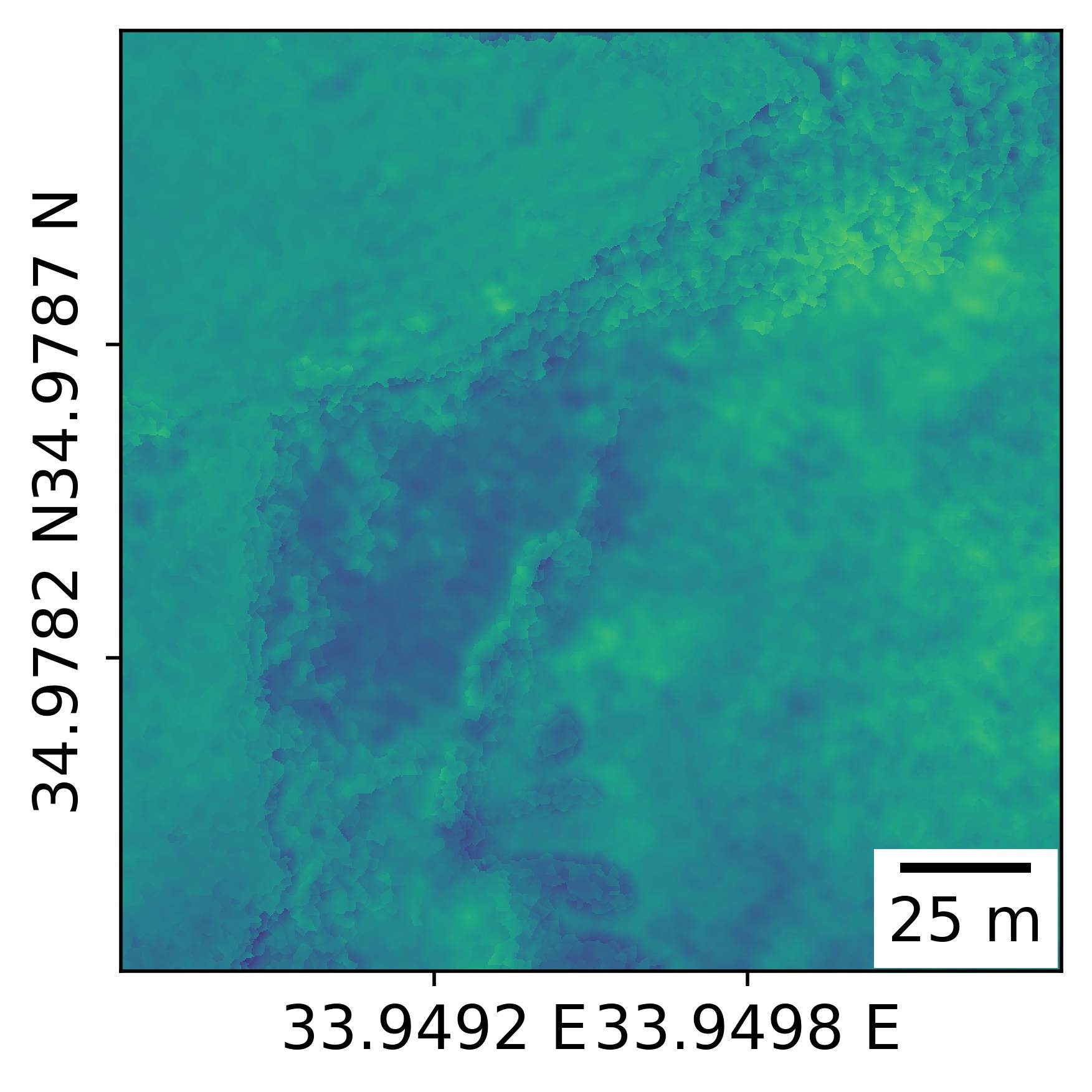}
    \end{minipage}& 
    \hfill

    \begin{minipage}[c]{0.44\columnwidth}
        \centering
        \includegraphics[width=\linewidth]{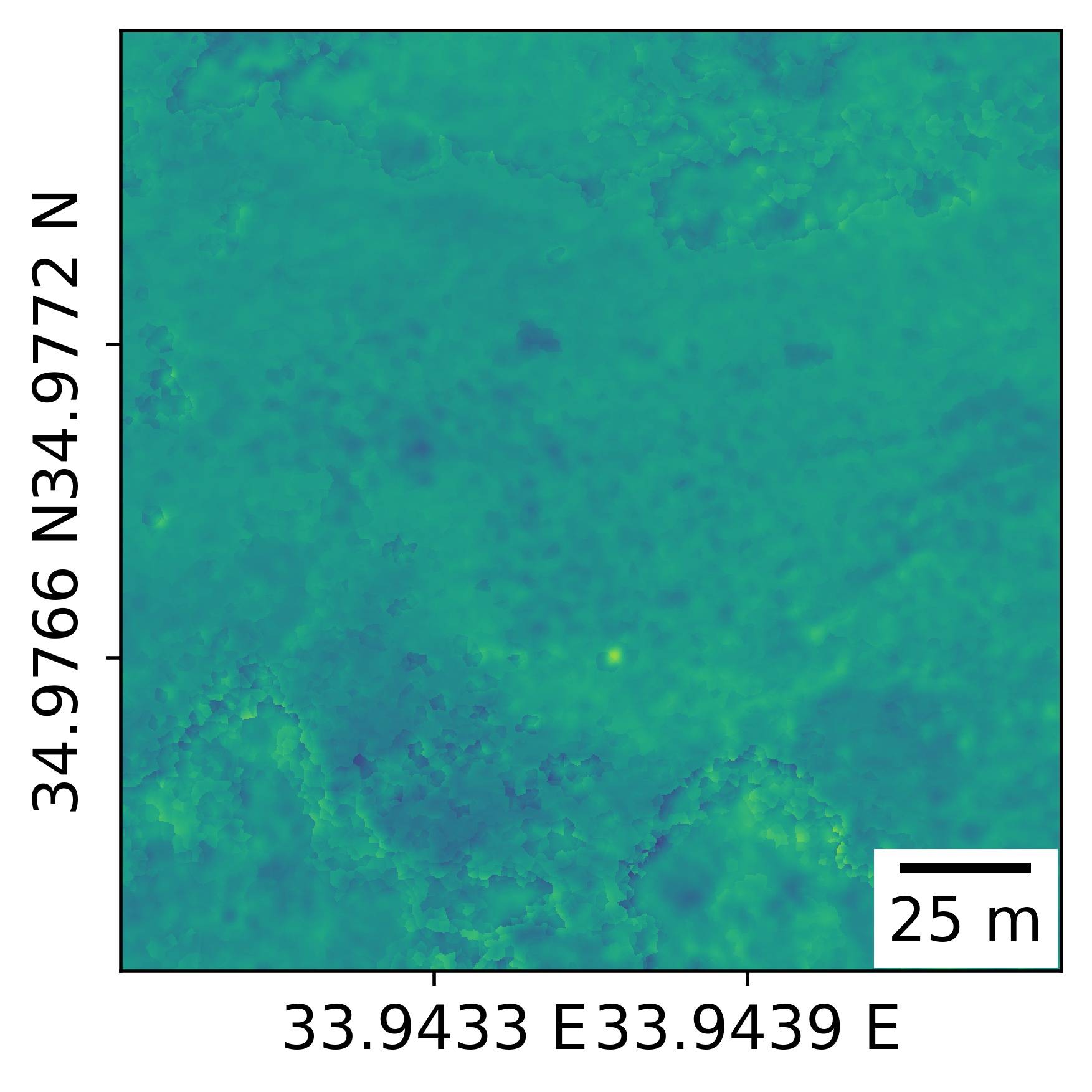}
    \end{minipage}&    
    \hfill

    \begin{minipage}[c]{0.55\columnwidth}
        \centering
        \includegraphics[width=\linewidth]{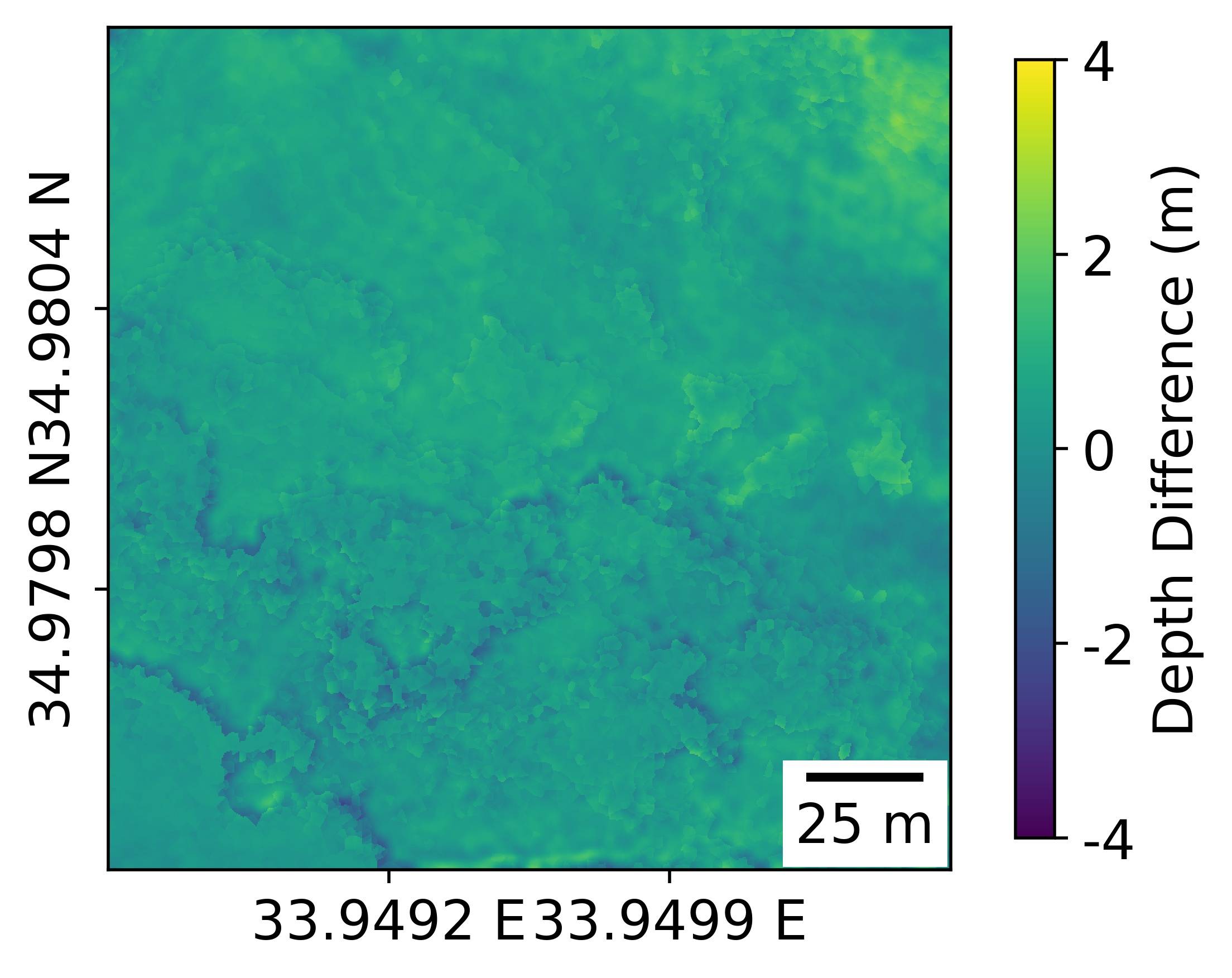}
    \end{minipage}  \vspace{1pt}\\

       \textbf{(b)} &\begin{minipage}[c]{0.44\columnwidth}
        \centering
        \includegraphics[width=\linewidth]{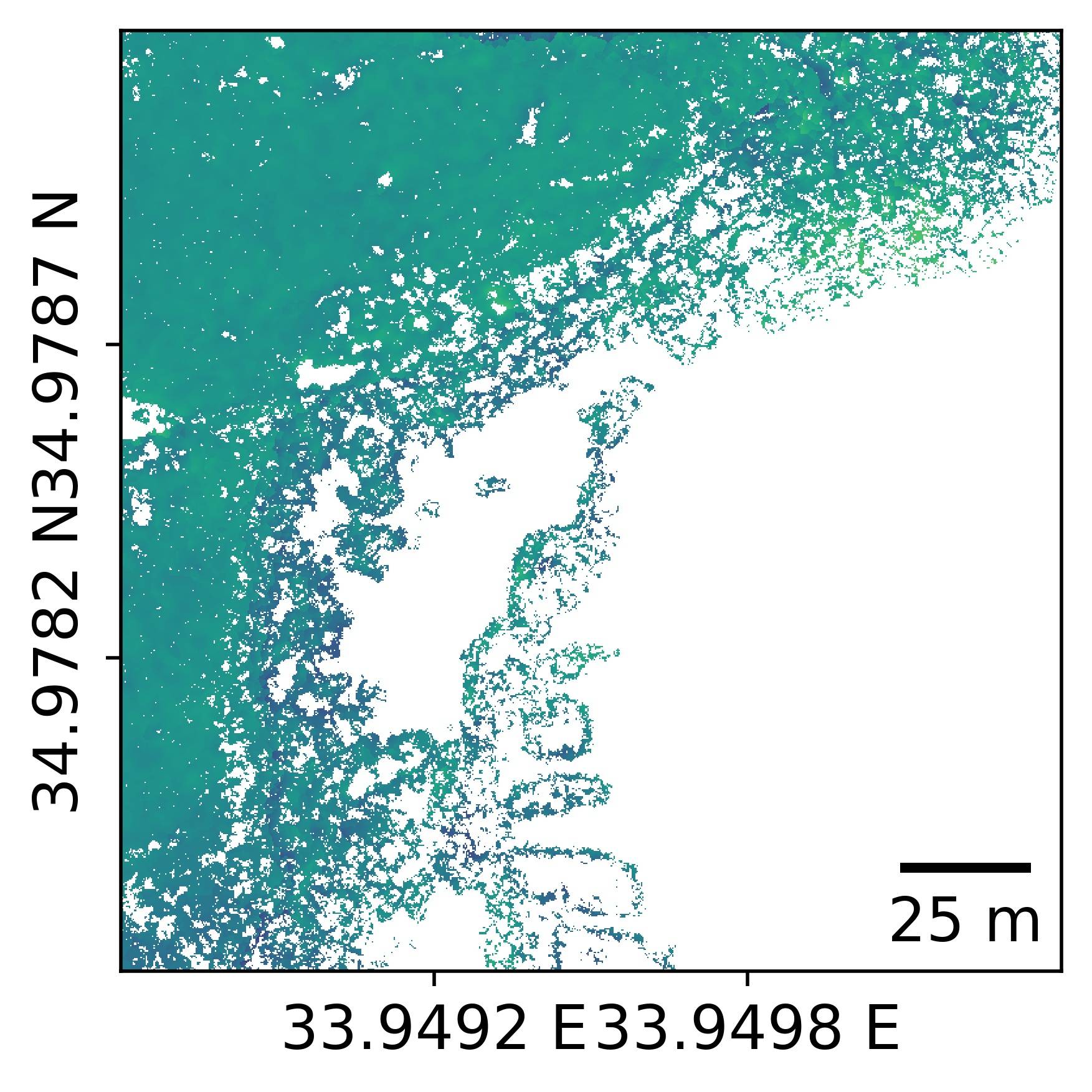}
    \end{minipage}& 
    \hfill

    \begin{minipage}[c]{0.44\columnwidth}
        \centering
        \includegraphics[width=\linewidth]{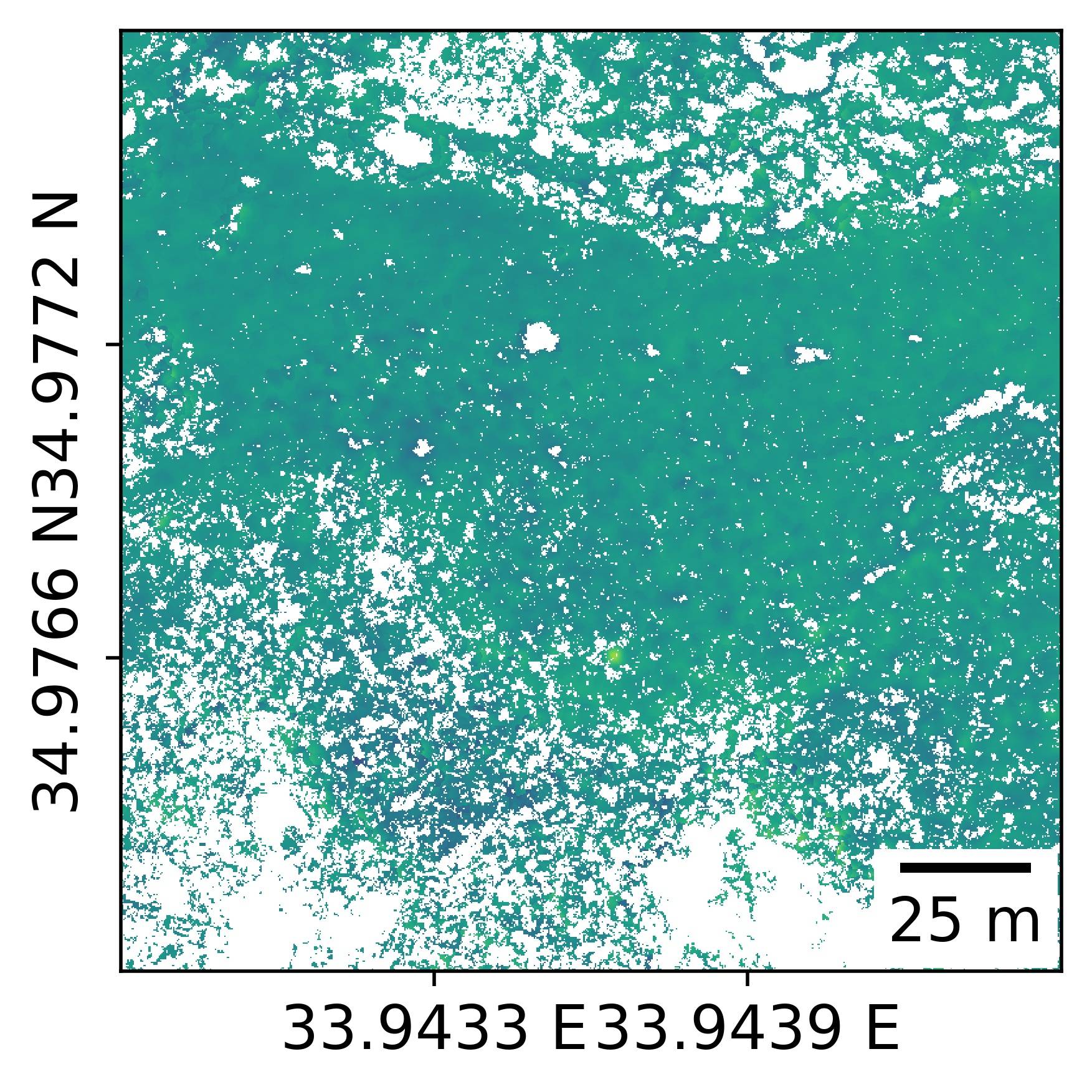}
    \end{minipage}&    
    \hfill

    \begin{minipage}[c]{0.55\columnwidth}
        \centering
        \includegraphics[width=\linewidth]{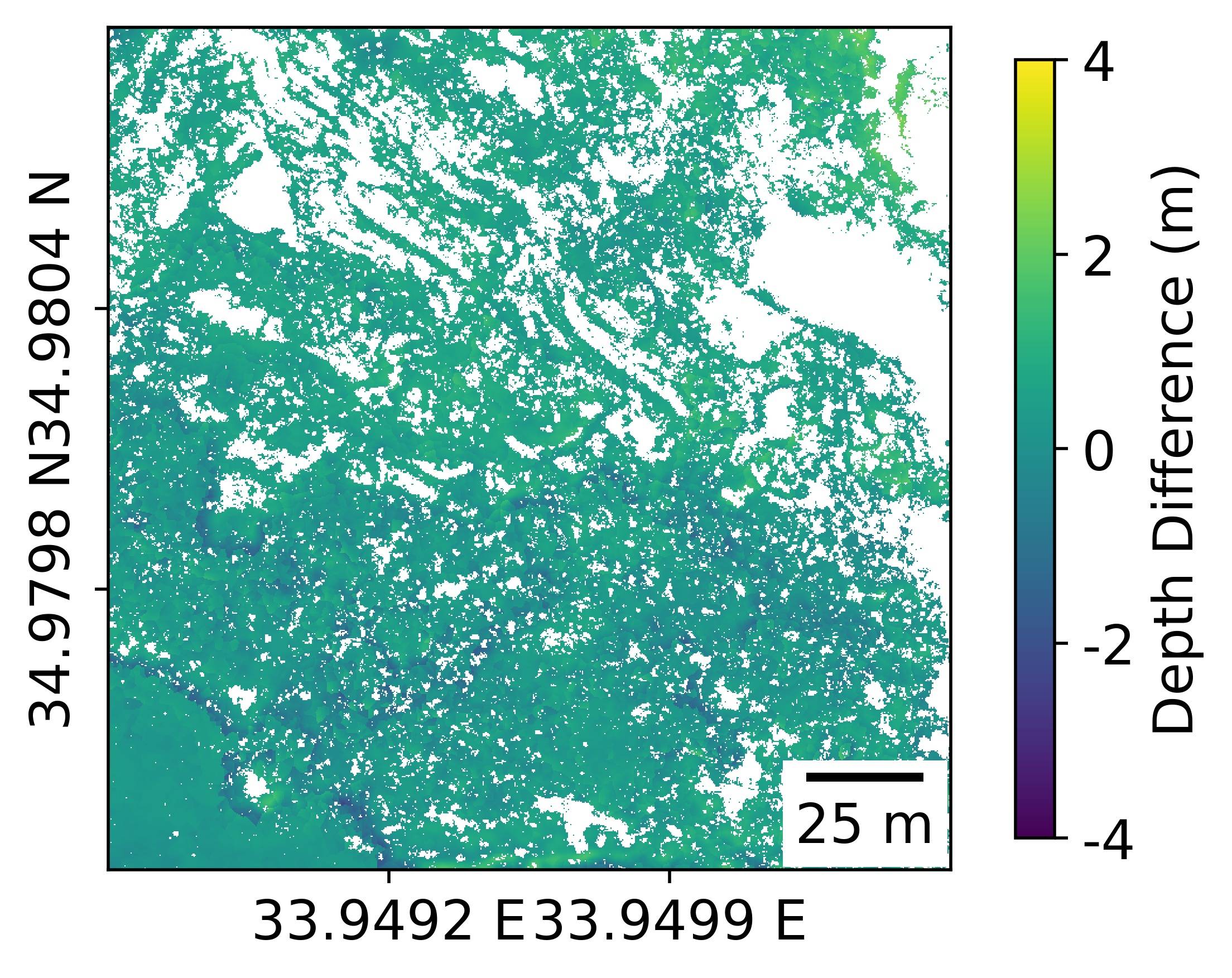}
    \end{minipage}  \vspace{1pt}\\

    \textbf{(c)} &\begin{minipage}[c]{0.44\columnwidth}
        \centering
        \includegraphics[width=\linewidth]{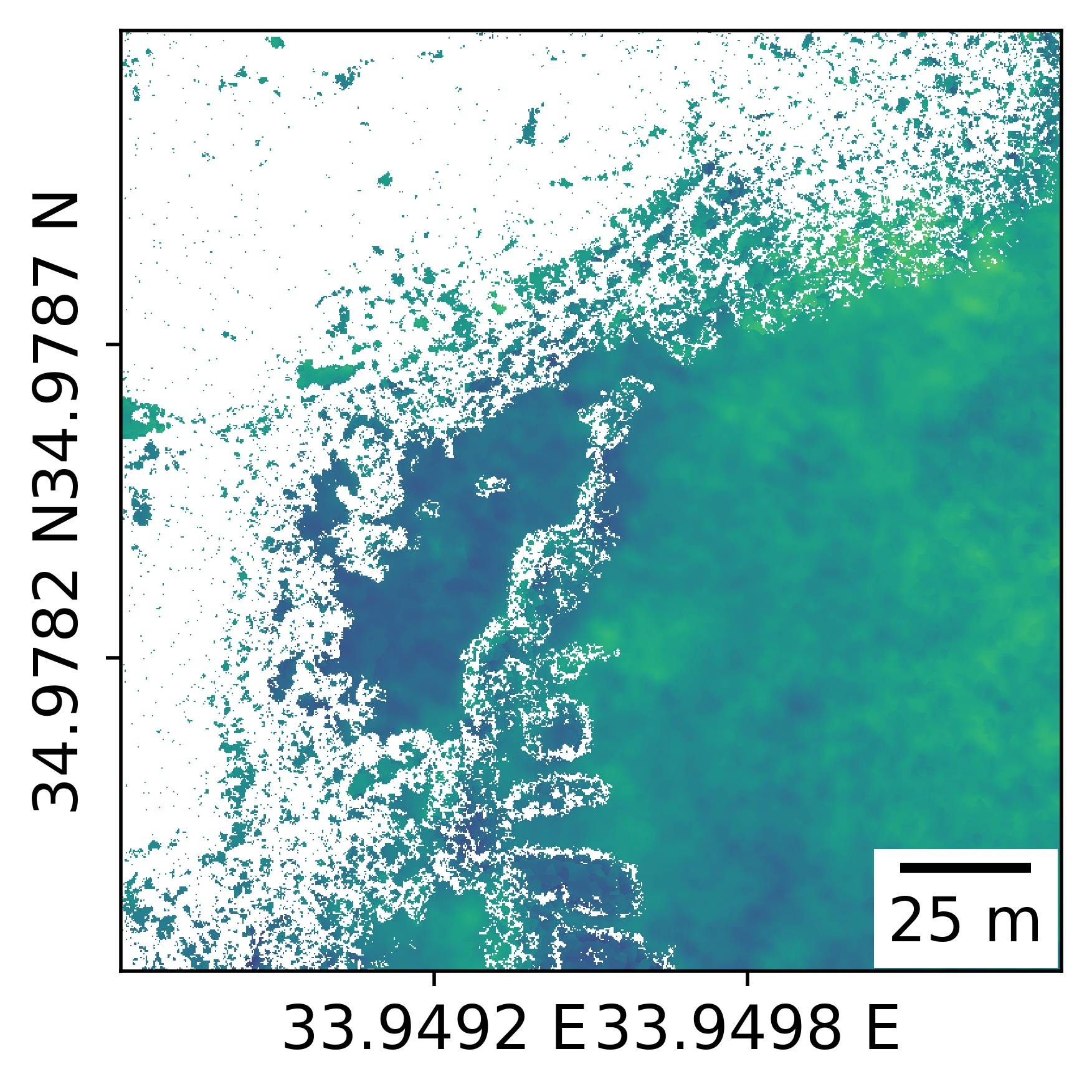}
    \end{minipage}& 
    \hfill

    \begin{minipage}[c]{0.44\columnwidth}
        \centering
        \includegraphics[width=\linewidth]{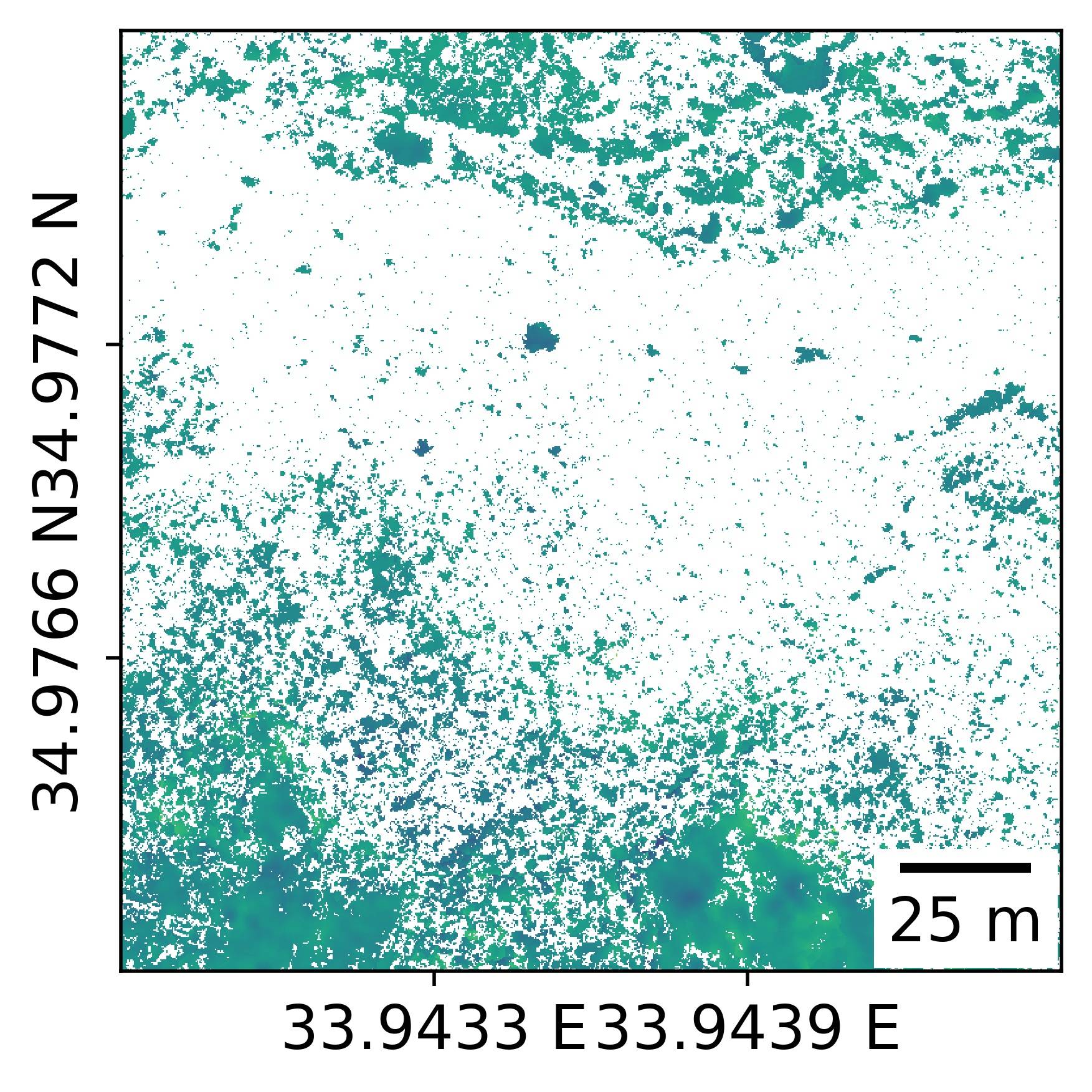}
    \end{minipage}&    
    \hfill

    \begin{minipage}[c]{0.55\columnwidth}
        \centering
        \includegraphics[width=\linewidth]{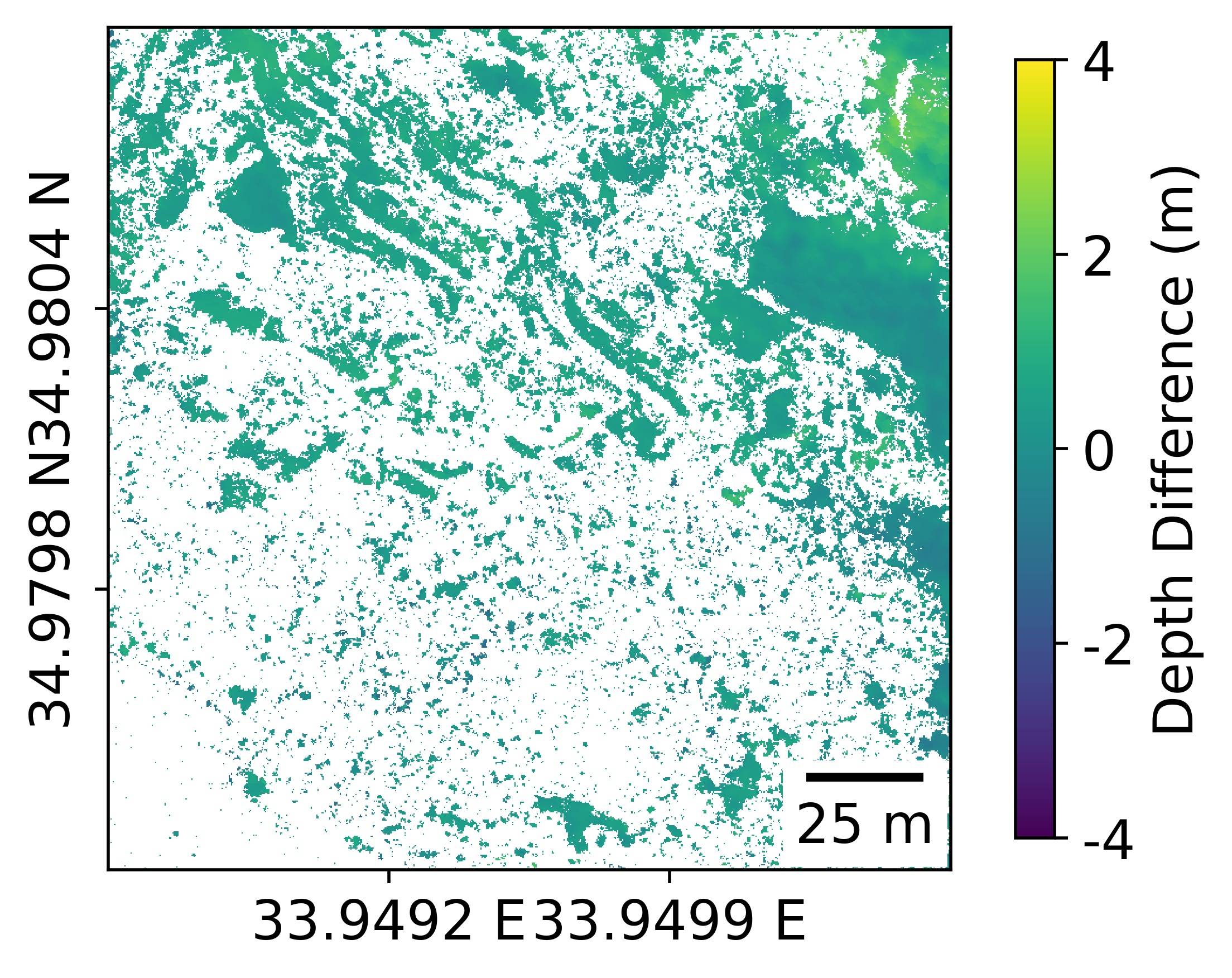}
    \end{minipage}  \vspace{1pt}\\

\end{tabular}
\vspace{-0.07in}
\caption{Depth differences between the reference and the predicted depths from the Swin-BathyUNet (a) for the whole prediction, (b) for the "non-gaps", and (c) for the "gaps" in the Agia Napa area. White areas are the data gaps.}
\vspace{-0.07in}
\label{fig:fig9a}
\end{figure*}

\begin{figure*}[h!]
  \setlength{\tabcolsep}{1.5pt}
  \renewcommand{\arraystretch}{1}
  \footnotesize
  
\centering

  \begin{tabular}{m{0.5cm}ccc}

   \textbf{(a)} &\begin{minipage}[c]{0.44\columnwidth}
        \centering
        \includegraphics[width=\linewidth]{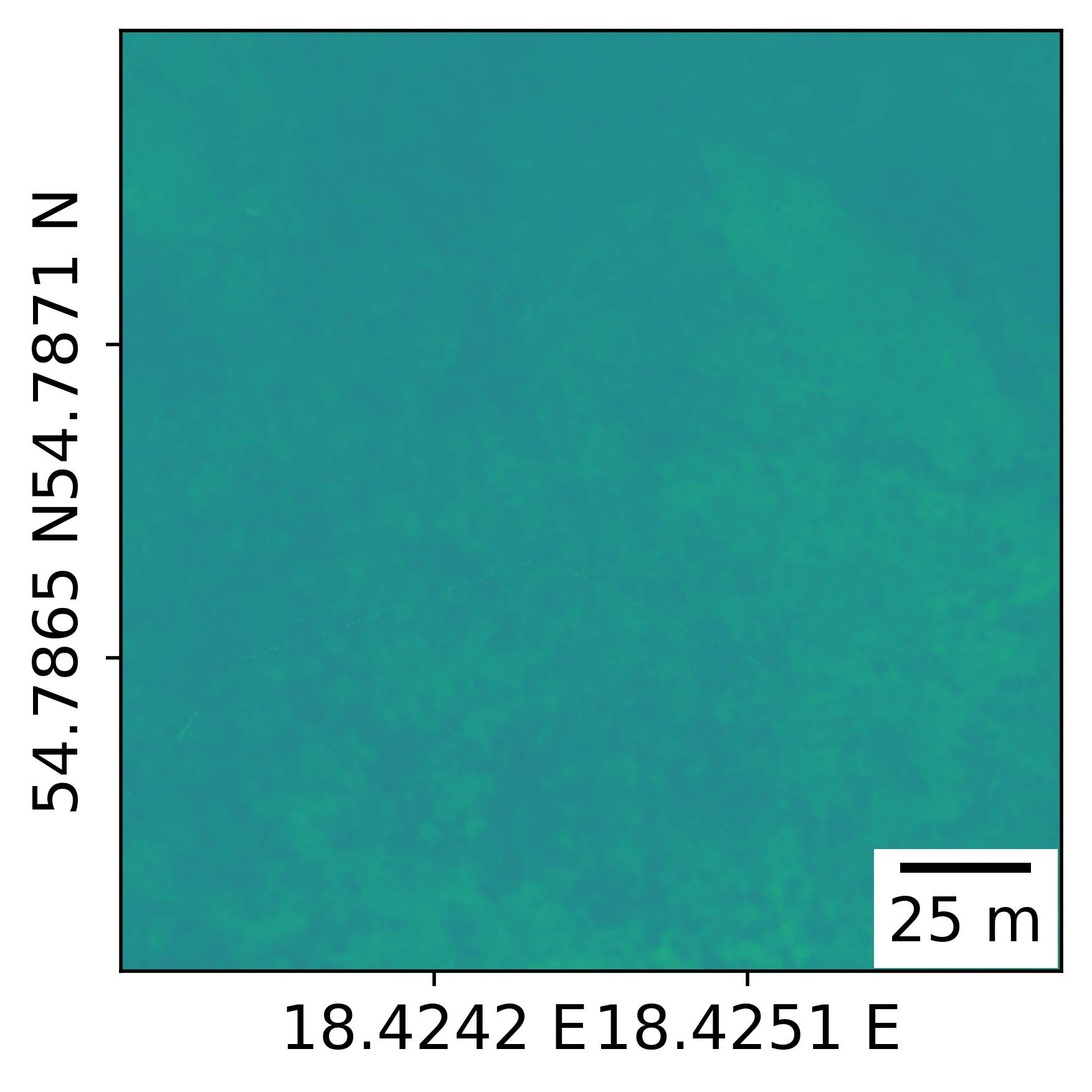}
    \end{minipage}& 
    \hfill

    \begin{minipage}[c]{0.44\columnwidth}
        \centering
        \includegraphics[width=\linewidth]{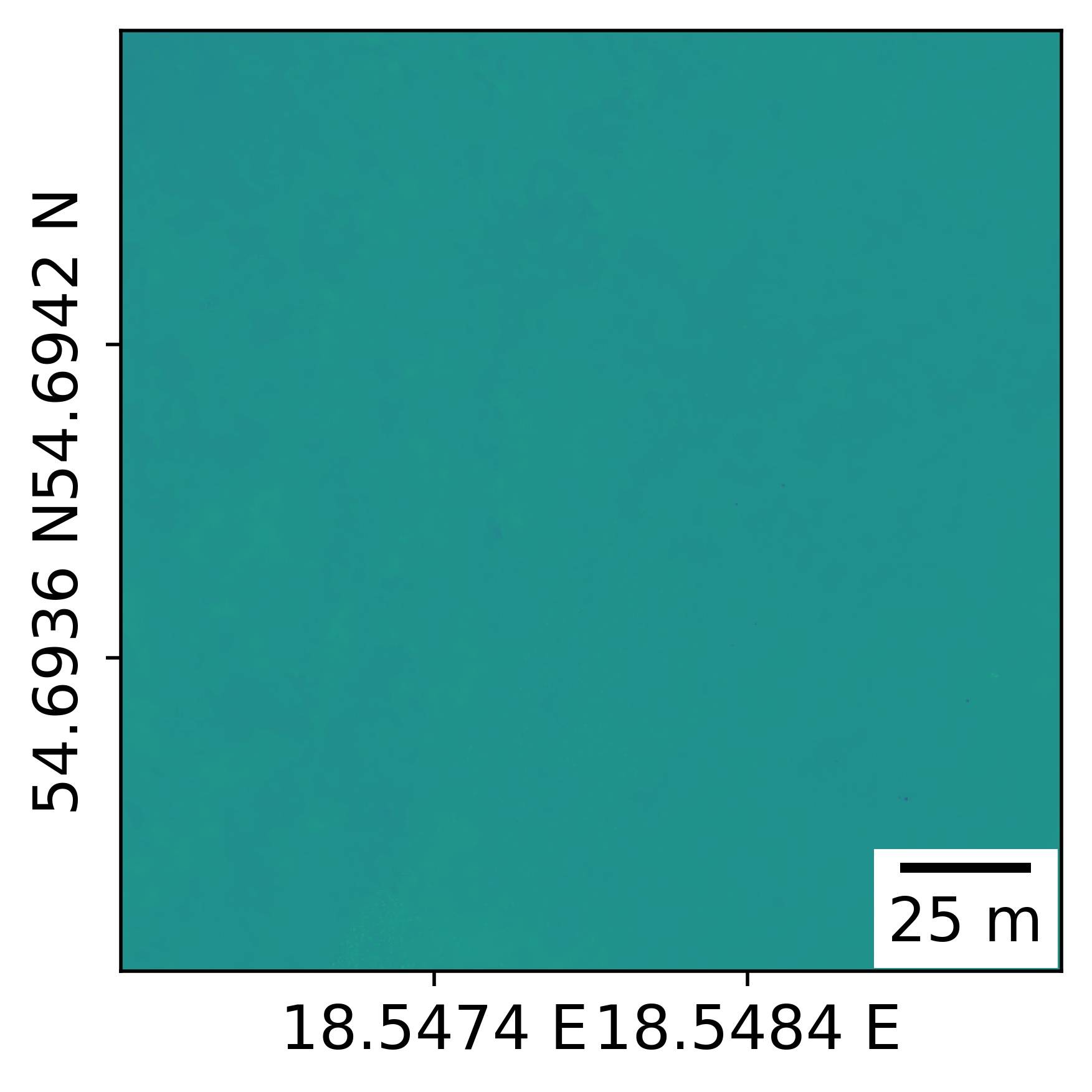}
    \end{minipage}&    
    \hfill

    \begin{minipage}[c]{0.55\columnwidth}
        \centering
        \includegraphics[width=\linewidth]{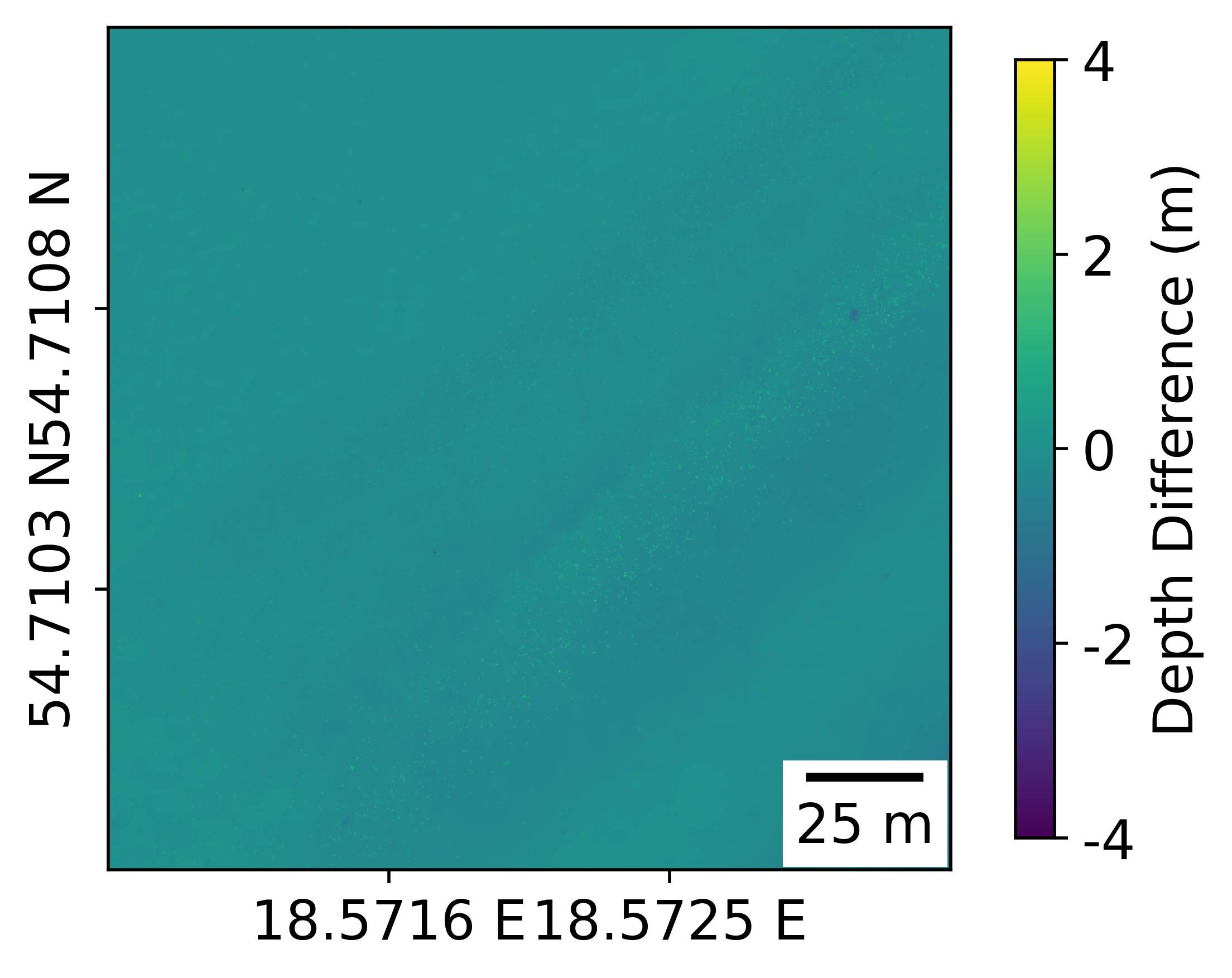}
    \end{minipage}  \vspace{1pt}\\

       \textbf{(b)} &\begin{minipage}[c]{0.44\columnwidth}
        \centering
        \includegraphics[width=\linewidth]{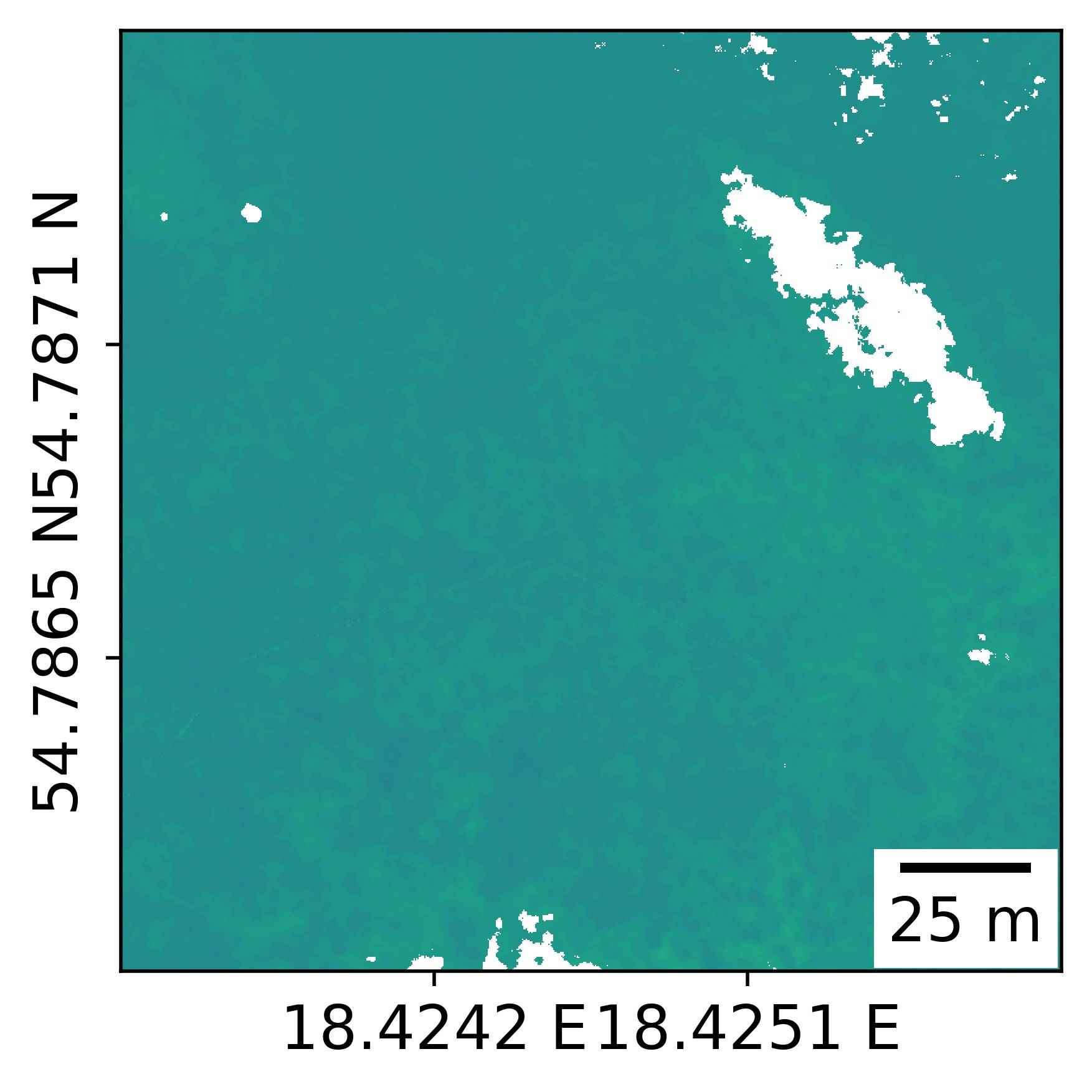}
    \end{minipage}& 
    \hfill

    \begin{minipage}[c]{0.44\columnwidth}
        \centering
        \includegraphics[width=\linewidth]{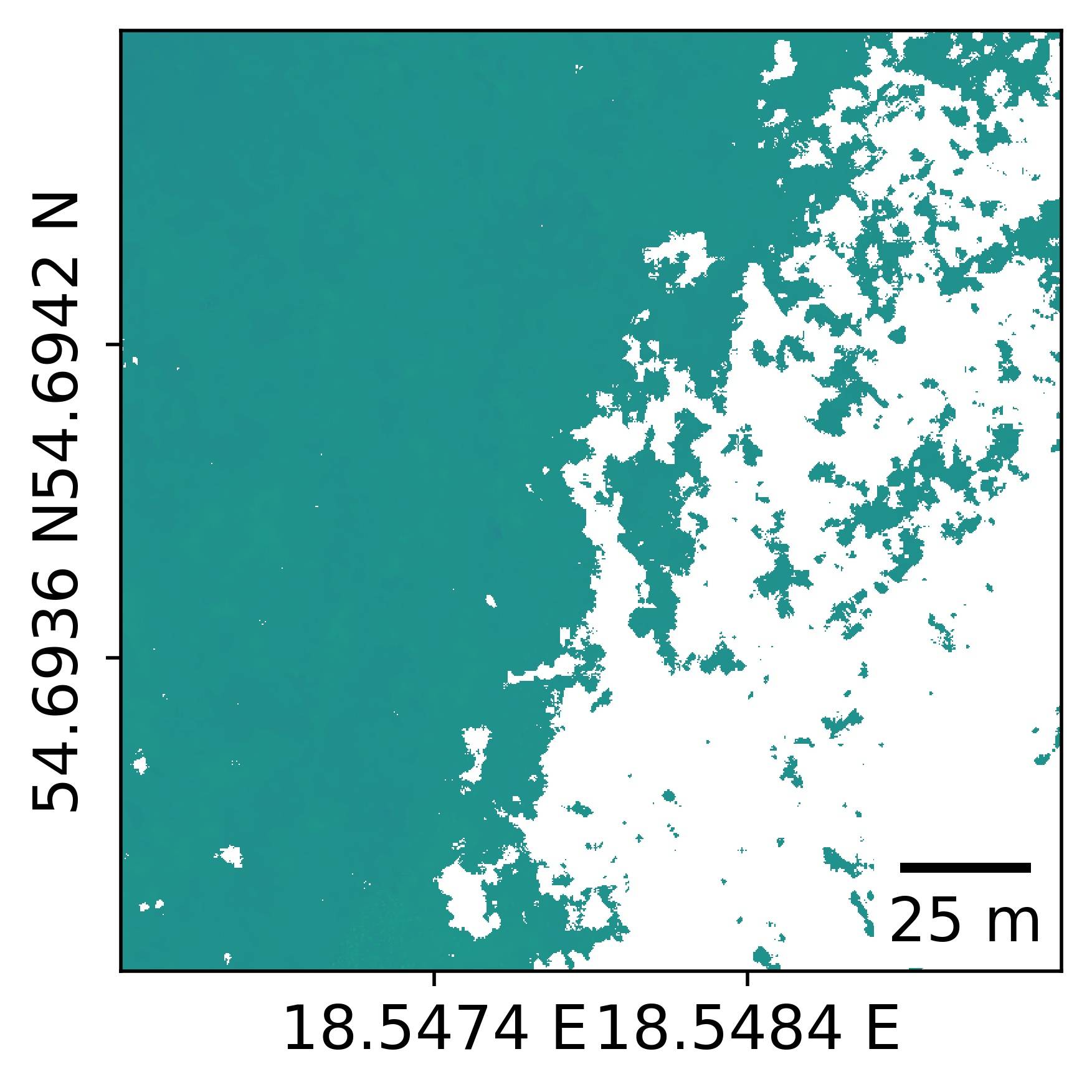}
    \end{minipage}&    
    \hfill

    \begin{minipage}[c]{0.55\columnwidth}
        \centering
        \includegraphics[width=\linewidth]{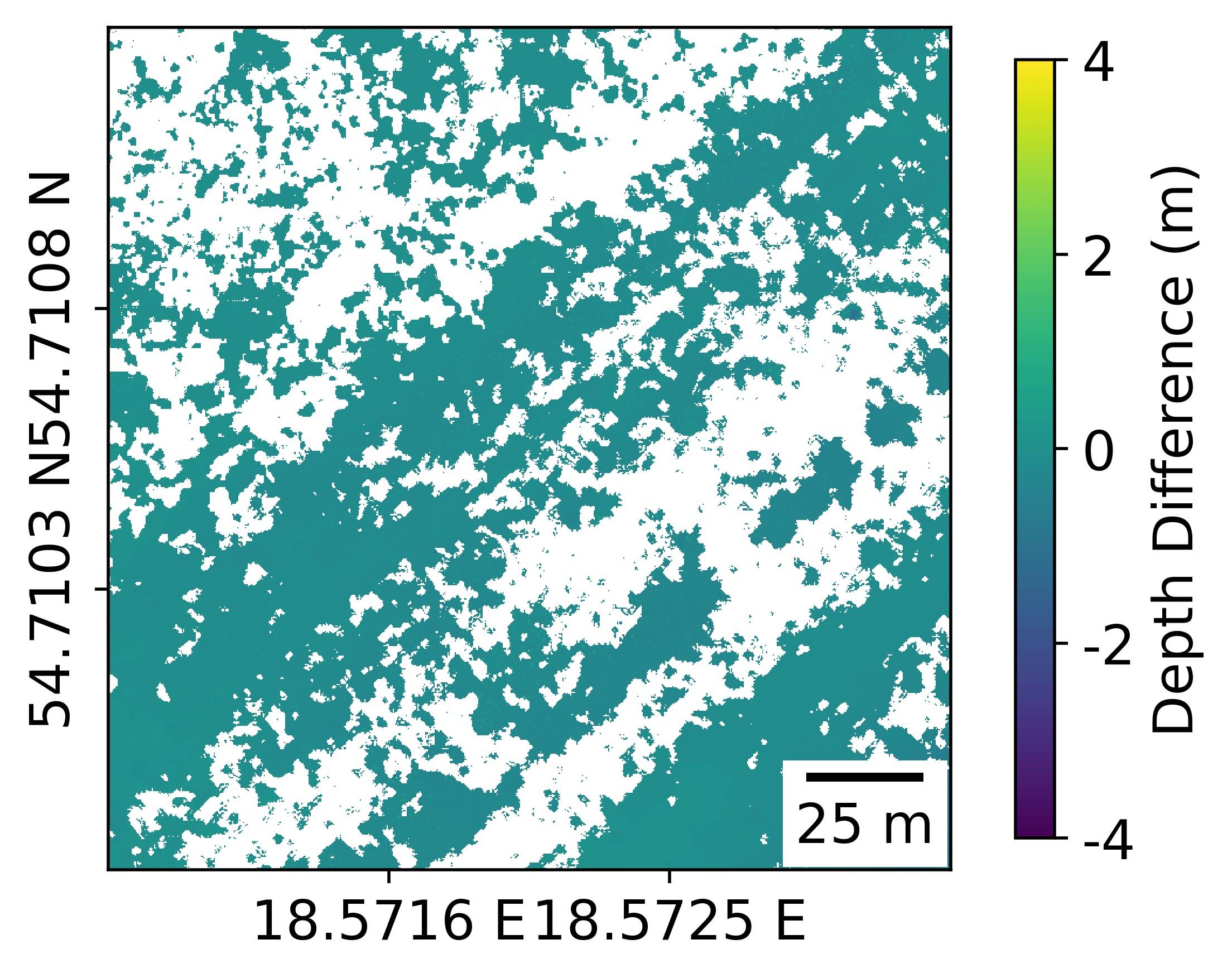}
    \end{minipage}  \vspace{1pt}\\

    \textbf{(c)} &\begin{minipage}[c]{0.44\columnwidth}
        \centering
        \includegraphics[width=\linewidth]{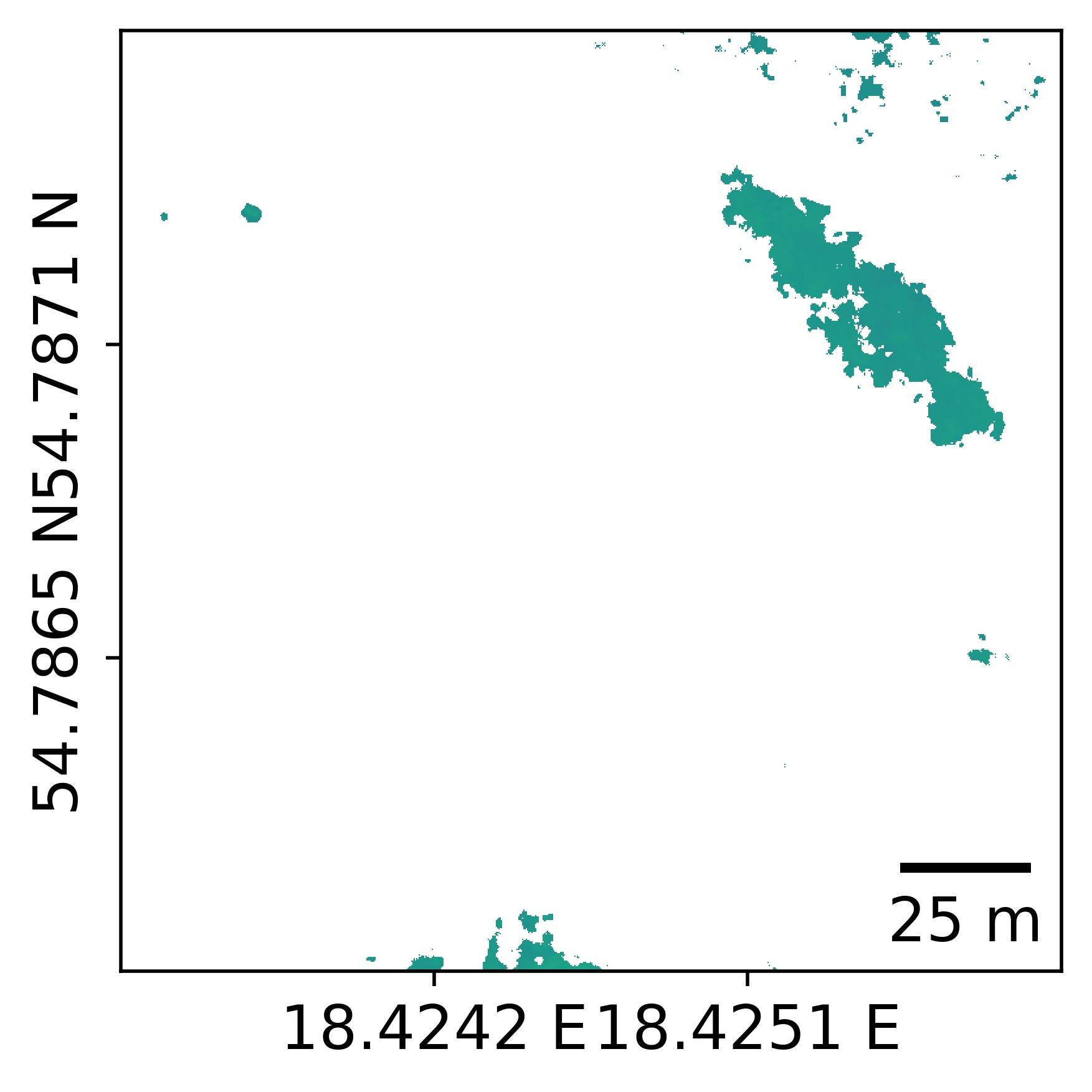}
    \end{minipage}& 
    \hfill

    \begin{minipage}[c]{0.44\columnwidth}
        \centering
        \includegraphics[width=\linewidth]{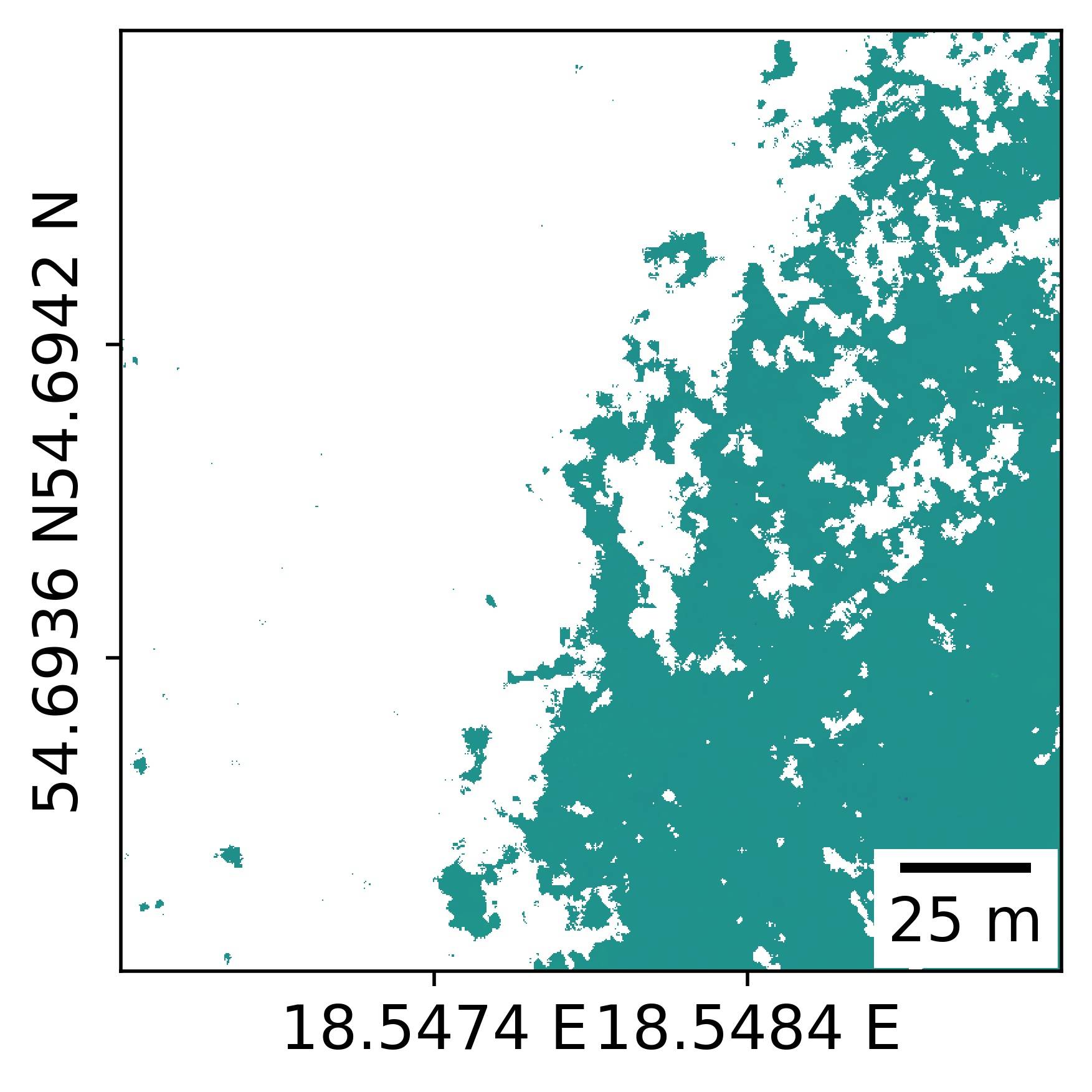}
    \end{minipage}&    
    \hfill

    \begin{minipage}[c]{0.55\columnwidth}
        \centering
        \includegraphics[width=\linewidth]{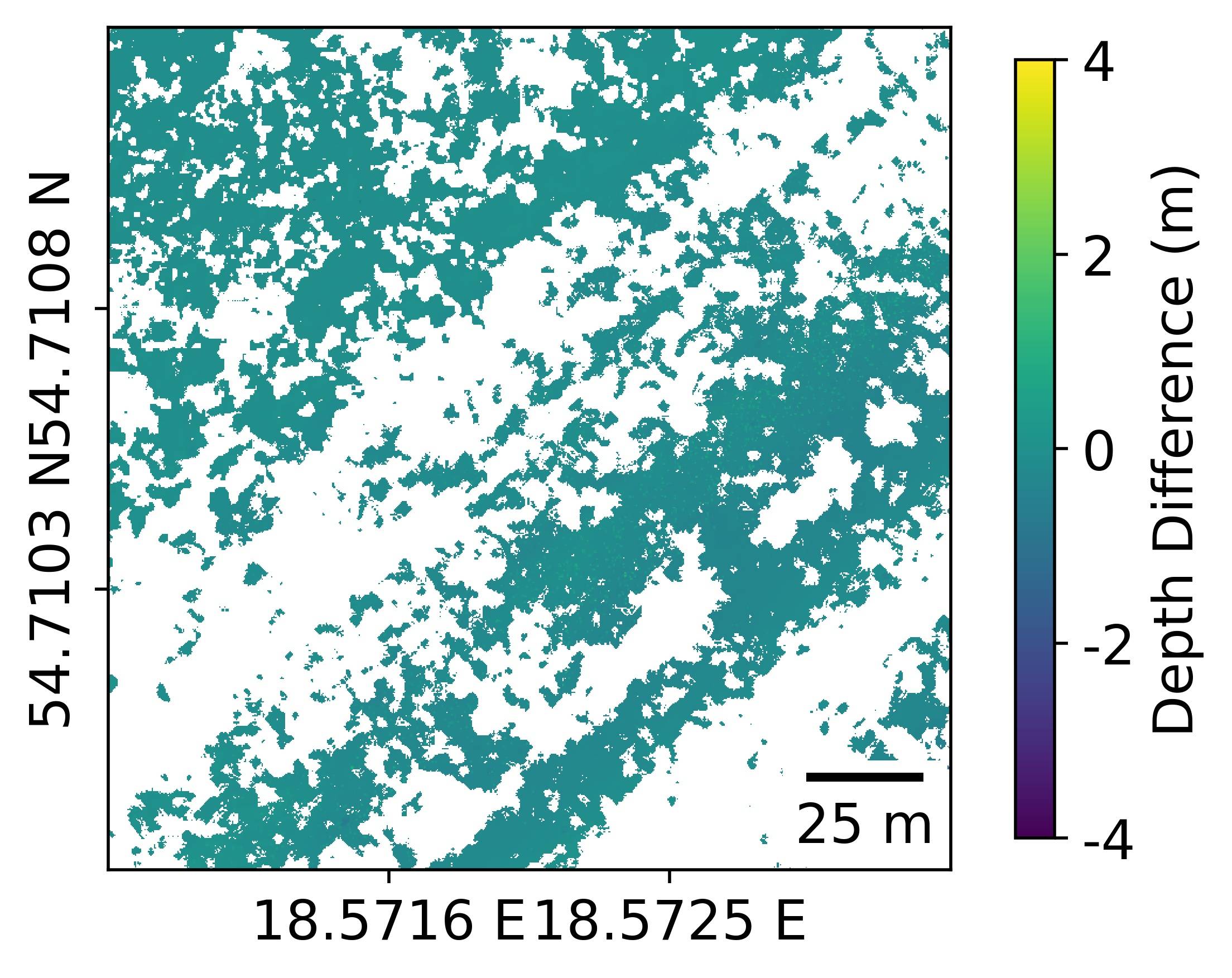}
    \end{minipage}  \vspace{1pt}\\
\end{tabular}
\vspace{-0.07in}
\caption{Depth differences between the reference and the predicted depths from the Swin-BathyUNet (a) for the whole prediction, (b) for the "non-gaps", and (c) for the "gaps" in the Puck Lagoon area. White areas are the data gaps.}
\label{fig:fig10a}
\vspace{-0.07in}
\end{figure*}

The evaluation of depth predictions in Agia Napa (Table \ref{table:table1}) reveals differences between predictions for "non-gaps" areas and "gaps" areas when compared to reference data (Figure \ref{fig:fig9a}). This was however expected since no depth labels are available for training in the "gaps" areas. For "non-gaps" areas, the RMSE is 0.38m, the MAE is 0.29m, and the Std. is 0.32m. These results are closely aligned with resulted metrics between the corrected by the refraction SfM-MVS depth data and the reference data, before being fed into the deep learning model. The predictions for "gaps" areas show slightly increased error metrics, with an RMSE of 0.58m, MAE of 0.43m, and Std. of 0.49m. This indicates that the model estimates depths in the "gaps" regions with slightly less accuracy, primarily due to the absence of depth labels. When considering the whole predictions against the reference, the predictions show an RMSE of 0.49m, MAE of 0.37m, and Std. of 0.44m. These errors are comparable with the ones achieved in \cite{magicbathynet} using sparse LiDAR data as ground truth data for training in a subset of the same area. To provide an example, for a randomly selected pixel belonging in the "gaps" regions, the reference depth is equal to -13.59m while it's predicted depth by Swin-BathyUNet here is -13.26m. This equals with an error of 2.4\% of the real depth. Unlike Agia Napa, the evaluation of depth predictions in Puck Lagoon (Table \ref{table:table2}) reveals similar differences between predictions for "non-gaps" areas and "gaps" areas when compared to reference data (Figure \ref{fig:fig10a}). This is explained by the shallower depths as well as the smaller percentages of the "gaps" in this area (see Tables \ref{table:table1.2} and \ref{table:table2.2}). For "non-gaps" areas, the RMSE is 0.15m, the MAE is 0.10m, and the Std. is 0.15m, indicating a slight increase compared to the corrected SfM-MVS metrics. In contrast, the "gaps" areas exhibit approximately 33\% higher error metrics. The predictions for the whole area display similar metrics to those of the "non-gaps" areas.

The results indicate a relatively high accuracy in predicting depths in "gaps" regions as well as for the whole prediction, suggesting Swin-BathyUNet's effectiveness in these areas. 

\subsubsection{CATZOC- and Survey-based evaluation of the predicted depths}
In this subsection, we evaluate the predicted bathymetry using CATZOC (S-57/S-101) and Survey (S-44) criteria \citep{IHO}. The Category Zone of Confidence (CATZOC) framework, established by the International Hydrographic Organization (IHO), provides a standardized approach for assessing the quality and reliability of bathymetric data. S-44 refers to the IHO S-44 standard, which defines accuracy and quality requirements for bathymetric surveys. By assigning CATZOC and S-44 ratings, we assess the suitability of depth predictions' for various applications, including navigation, habitat mapping, and coastal monitoring.

According to the S-57/S-101 (ZOC) specifications, the bathymetric data produced in this study meets the horizontal uncertainty tolerance for CATZOC A1 standard. Bathymetry is derived from orthoimages generated after bundle adjustment with achieved RMSEs of  0.05m (X), 0.05m (Y), and 0.07m (Z) for Agia Napa, with an average reprojection error of 1.11 pixels (pixel size: 0.06m). In Puck Lagoon, the results are even more precise, with RMSEs of 0.01m (X), 0.01m (Y), and 0.01m (Z), with an average reprojection error of 1.19 pixels (pixel size: 0.07m). Based on CATZOC vertical uncertainty thresholds, 72\% of the predicted depths in Agia Napa fall within CATZOC A1, while the remaining 28\% are classified as A2. In Puck Lagoon, 94.4\% of the data meets A1 standards, 4.6\% falls under A2, and only 1\% is classified as C. The completeness of bathymetric data is another critical aspect of CATZOC evaluation. ZOC A1 requires 100\% seafloor coverage and feature detection of 2m, criteria that are fully met.  

In addition to the above, the uncertainty of the predicted bathymetric data was assessed according to the IHO S-44 classification system, which divides the data into four zones - Exclusive, Special, 1a, 1b and 2 - each representing different levels of uncertainty. For the horizontal uncertainty, the achieved results comply with the Exclusive Order of S-44, which specifies a survey tolerance of 1m. For the vertical uncertainty, in Agia Napa, 66.67\% of the predicted depths meet the Exclusive standard, indicating very high accuracy, while 28.57\% falls under the Special category, reflecting slightly lower one. A smaller portion, 4.76\%, is classified as 1a, and another 4.76\% is categorized as 1b, signifying less reliable measurements. In Puck Lagoon, the data shows a high level of accuracy, with 94.64\% meeting the Exclusive standard. 4.10\% is classified as Special, and 1.26\% falls into both 1a and 1b, reflecting lower confidence in these areas. Exclusive standard is also met for seafloor coverage and feature detection. 

Overall, these results demonstrate that the vast majority of the predicted bathymetry for both Agia Napa and Puck Lagoon meet the highest vertical accuracy standards of both S-57/S-101 and S-44, with only a small fraction of data falling into lower confidence categories, demonstrating Swin-BathyUNet's suitability for various scientific and operational applications.

\subsection{Accuracy analysis by depth intervals and seabed habitats}
\label{Accuracy analysis by depth intervals and seabed habitats}
\begin{figure}[h!]
  \setlength{\tabcolsep}{1.5pt}
  \renewcommand{\arraystretch}{1}
  \footnotesize
  \centering
  \begin{tabular}{cc}
    \begin{minipage}[c]{0.44\columnwidth}
        \centering
        \includegraphics[width=1\linewidth]{img/img_410.jpg}\\
        \textbf{(a)}
        \vspace{0.5em}
    \end{minipage} &
    \hfill
    \begin{minipage}[c]{0.44\columnwidth}
        \centering
        \includegraphics[width=1\linewidth]{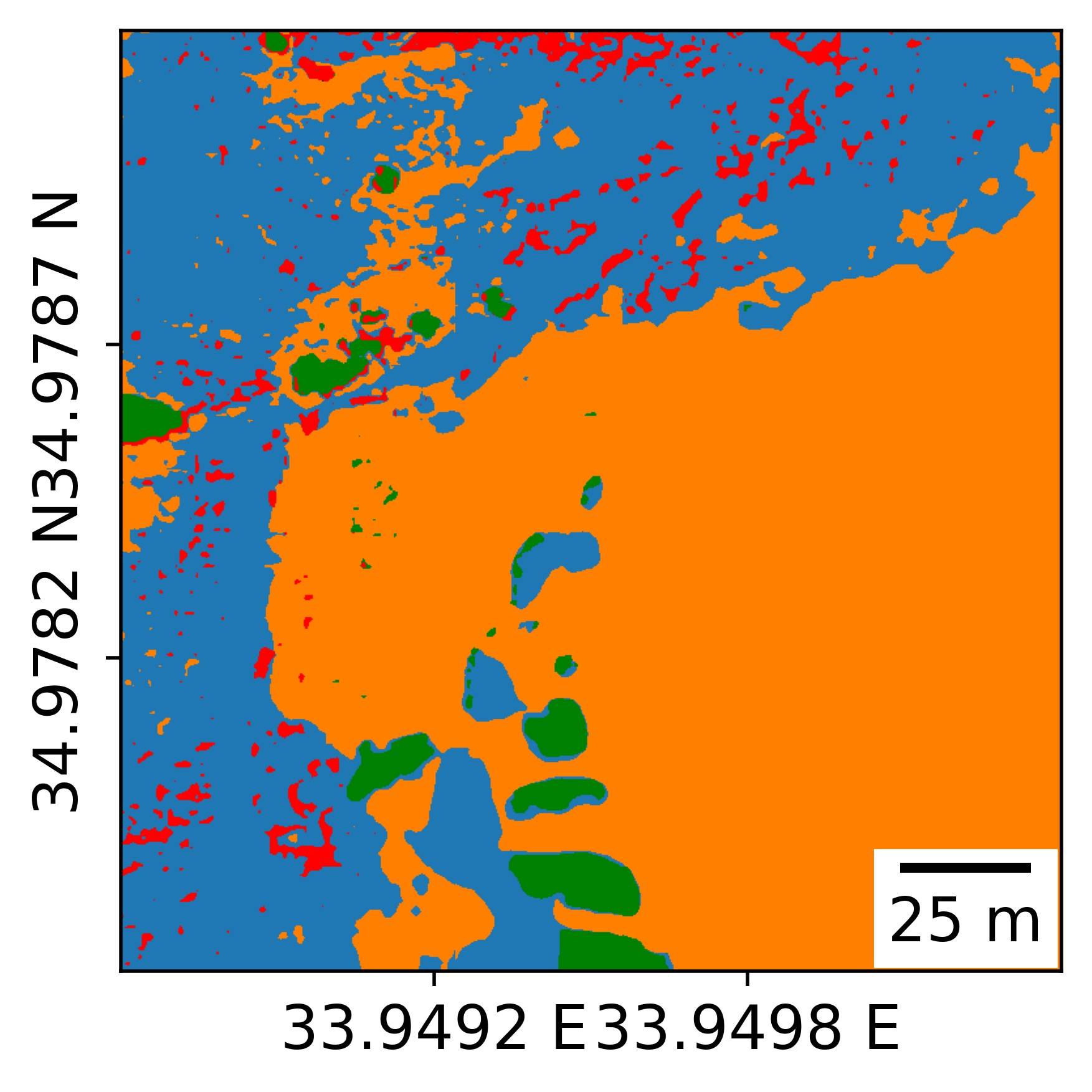}\\
        \textbf{(b)}
        \vspace{0.5em}
    \end{minipage}\\
        \begin{minipage}[c]{0.44\columnwidth}
        \centering
        \includegraphics[width=1\linewidth]{img/img_1.jpg}\\
        \textbf{(c)}
    \end{minipage} &
    \hfill
    \begin{minipage}[c]{0.44\columnwidth}
        \centering
        \includegraphics[width=1\linewidth]{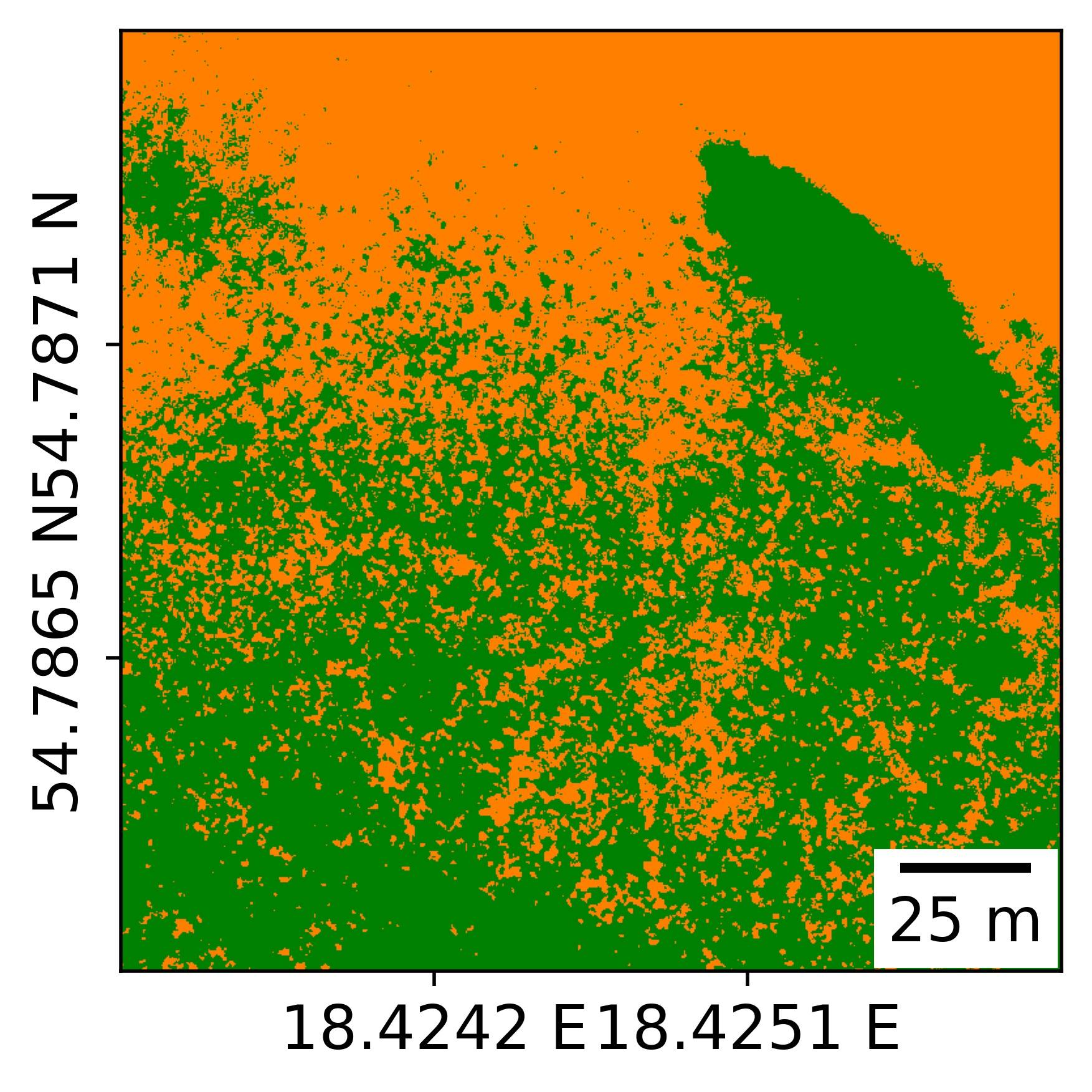}\\
        \textbf{(d)}
    \end{minipage}\\
     \\
    \multicolumn{2}{c}{
    \begin{minipage}[c]{1\columnwidth}
        \centering
        \includegraphics[width=\linewidth]{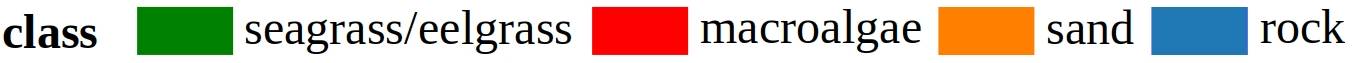}
    \end{minipage}}\\
  \end{tabular}
  
  \caption{Example RGB orthoimage patches acquired over Agia Napa (a) and Puck Lagoon (c) and the respective seabed classes (b) and (d) obtained by the pretrained SegFormer models from MagicBathyNet \citep{magicbathynet}.}
  \label{fig:labels}
\end{figure}

In this subsection, we provide a more granular analysis of the bathymetric accuracy by examining the performance of Swin-BathyUNet across different depth intervals and diverse seabed habitats. To that direction, a detailed breakdown by depth ranges (0-5m, 5-10m, and 10-20m) and habitat types (e.g., seagrass, sand etc.) offers a deeper insight into the stability and reliability of the proposed approach in varying environmental conditions. This stratified analysis aims to highlight the strengths and potential weaknesses of the proposed method, ensuring a more comprehensive understanding of its performance in complex and heterogeneous underwater environments. For the pixel-based classification in this analysis, we utilized the pretrained models from MagicBathyNet \citep{magicbathynet} to generate predictions on the aerial orthoimage patches used for bathymetry estimation. Specifically, we used the models trained with SegFormer \citep{segformer}, a state-of-the-art transformer-based architecture for semantic segmentation. As reported in \citep{magicbathynet}, these models achieved high pixel-based classification accuracy of 94.67\% in Agia Napa and of 97.33\% in Puck Lagoon, demonstrating their appropriateness for this task. Examples are shown in Figure \ref{fig:labels}.

\begin{figure*}[h!]
  \setlength{\tabcolsep}{1.5pt}
  \renewcommand{\arraystretch}{1}
  \footnotesize
  \centering
  \begin{tabular}{c}
    \begin{minipage}[c]{1.65\columnwidth}
        \centering
        \includegraphics[width=1\linewidth]{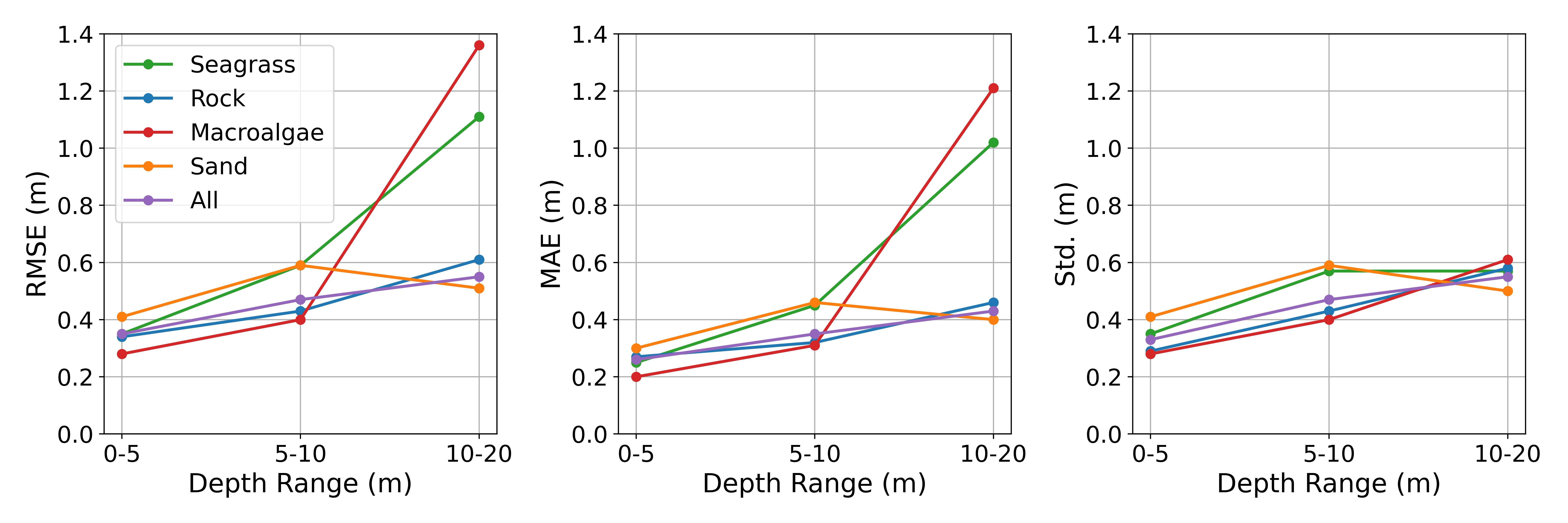}\\
        \textbf{(a)}
    \end{minipage} 
    \\

    \begin{minipage}[c]{1.65\columnwidth}
        \centering
        \includegraphics[width=1\linewidth]{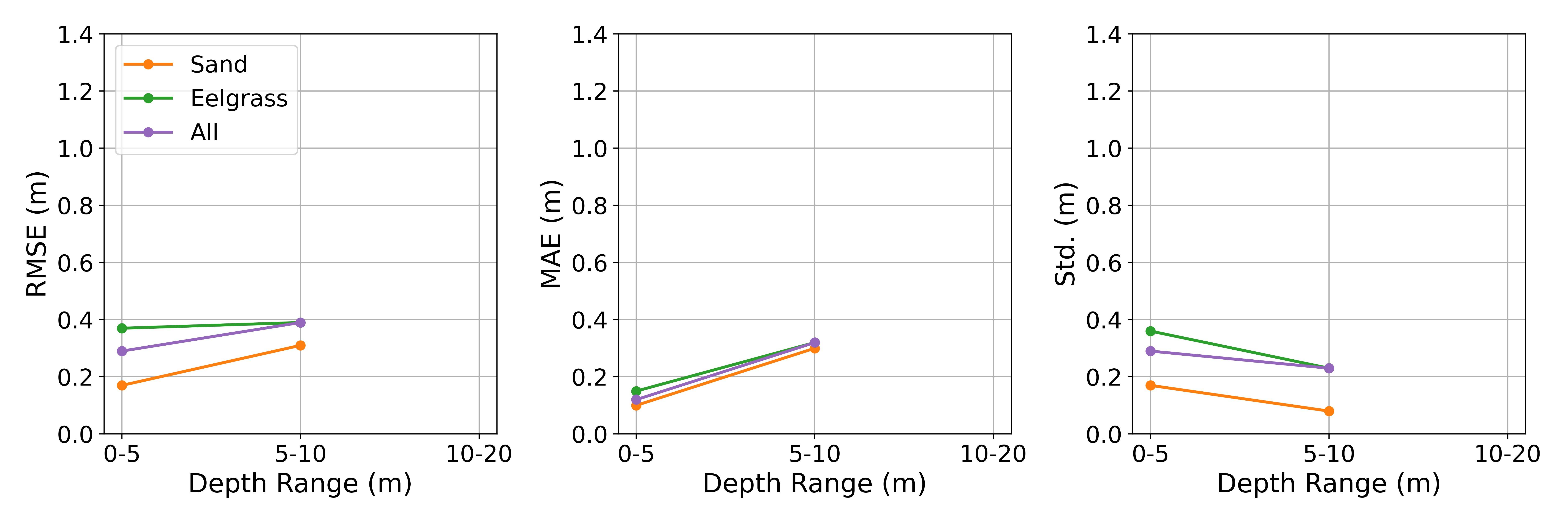}\\
        \textbf{(b)}
    \end{minipage}
  \end{tabular}
  
  \caption{Metrics by depth range and seabed class in (a) Agia Napa and in (b) Puck Lagoon. No depth data are available in Puck Lagoon for the 10-20m depth range.}
  \label{fig:labels_metrics}
  \vspace{-0.1in}
\end{figure*}

The error analysis across depth ranges and seabed classes reveals patterns in both the Agia Napa and Puck Lagoon areas (Figure \ref{fig:labels_metrics}). In both sites, all three metrics (RMSE, MAE, and Std.) tend to increase as depth increases, with the highest values observed in the deppest range. This shows that the model's performance slightly degrades with depth, likely due to factors such as increased light attenuation, scattering, and reduced texture information in deeper waters affecting the training data (Section \ref{III.B}). In Agia Napa, seabed classes with lower representation, such as seagrass (1.74\%) and macroalgae (14.29\%), exhibit the highest RMSE values in every depth range. Notably, their RMSEs increase significantly in the deepest regions. This is attributed to the limited exposure of the model to these classes. It can also be attributed to the fact that seagrass-predominate areas are particularly prone to SfM failures and matching difficulties, as discussed in Section \ref{III.B}, especially in the deeper parts. However, these two classes account for just 16.03\% of the total pixels. For the remaining 83.97\% pixels, the error increase in this deepest zone is limited to a few centimeters. Rock that covers 57.48\% of the area shows relatively lower RMSE values, indicating more reliable and stable predictions. It is also noted that a significant portion of the sand, which covers 26.49\% of the area, lacks SfM-MVS depth annotations, unlike rock, which is more comprehensively annotated. However, the results remain comparable between these two classes. Interestingly, in the deepest zone, the RMSE for sand is even lower than in the preceding shallower depth range. In Puck Lagoon, where eelgrass (51\%) and sand (49\%) dominate, sand exhibits the smallest RMSE (0.31m at 5-10m and 0.17m at 0-5m). The higher RMSE for eelgrass, particularly in shallow areas, suggests that vegetation complexity may introduce additional challenges in depth estimation. 

Overall, error trends suggest that the variation in performance is more intense depending on the seabed class rather than just the depth. Rare classes, like macroalgae and seagrass, exhibit a steeper increase in error contributing to higher uncertainty, particularly at greater depths, while others remain relatively stable. This indicates that class-specific characteristics play a more significant role in error propagation than depth alone.

\subsection{Ablation study}
To explore the influence of different settings on the model performance, we performed ablation analyses on the number of ViT attention blocks, the window size, the attention heads and the presence of the cross-attention module. Results are presented in Table \ref{table:ablation}. The impact of the combined Swin attention mechanisms used on the base architecture of U-Net is discussed in Section \ref{proposedvsunet}.

Initially, we explored their effect on our model by changing the number ($i$) of presence of the ViT attention blocks from $i \in \{1, 2, 3\}$. Results in Agia Napa area demonstrate that the bathymetry performance improves when the number of the attention modules increases from 1 to 2, and consequently form 2 to 3. Specifically, from 1 to 2 blocks, there is a substantial improvement in both RMSE and MAE, with a 2.73\% and 10.68\% reduction respectively, indicating that adding a second block significantly enhances the model's accuracy. However, this comes with a 4.58\% increase in standard deviation. When increasing the blocks from 2 to 3, the performance continues to improve, though at a slower rate. RMSE decreases by 1.94\%, and MAE by 2.03\%, while the Std. decreases slightly by 0.9\%. This suggests that increasing the number of blocks enhances the model's predictive accuracy and reduces the error. Overall, the trade-off seems beneficial, as the improvements in RMSE and MAE outweigh the minor increase in standard deviation. 

\begin{table}[!b]
\caption{Ablation study demonstrating the Average$_{3}$ testing performance on the Agia Napa area (whole prediction) compared to the reference data. 3 ViT blocks is the base configuration.}
\centering
\begin{tabular}{lccc}
\toprule
    Setting & RMSE(m) & MAE(m) & Std.(m) \\
\midrule
    1 ViT block  & 0.52 & 0.42   & \textbf{0.43}  \\
    2 ViT blocks & 0.50  & \textbf{0.37}   & 0.45   \\
    3 ViT blocks & \textbf{0.49}   & \textbf{0.37} & 0.44  \\
\midrule
    Window size 32 & 0.50 & 0.38 & 0.45  \\
\midrule
    4 attention heads & 0.51 & 0.38 & 0.45  \\
\midrule
    No cross-attention & 0.53 & 0.39 & 0.47  \\
\bottomrule
  \end{tabular}
\label{table:ablation}
\end{table}

By reducing the window size used by the Swin from 64 to 32, a reduction in accuracy is noticed. The RMSE increases by 1.90\%, the MAE by 2.63\%, and the Std. decreases by 2.03\%. By reducing the number of attention heads from 8 to 4, the model exhibits a notable deterioration in all metrics. The RMSE increases by 3.14\%, the MAE by 3.07\%, and the Std. by 2.63\%. Finally, removing the cross attention block results in significant error increases in bathymetry prediction. RMSE increases by 7.31\%, MAE  by 5.78\%, and Std. by 5.40\%. Overall results of the ablation study and \ref{proposedvsunet} support the effectiveness of the combined attention module used in capturing rich representations and increase bathymetry prediction accuracy. 

\subsection{Assessing Swin-BathyUNet for bathymetry prediction using Sentinel-2 imagery and refraction-corrected aerial SfM-MVS DSMs}
To demonstrate the potential use of our proposed approach in predicting depths across different modalities, we used co-registered Sentinel-2 (Level2A) image patches from the MagicBathyNet dataset \citep{magicbathynet} along with the respective reference data for evaluation (Figures \ref{fig:14} and \ref{fig:15}). We are addressing a scenario where there is no external reference bathymetric data, such as LiDAR, MBES, or nautical charts available to train the proposed Swin-BathyUNet model. To obtain the required training depth data, only a small portion of the target area is covered with refraction-corrected aerial SfM-MVS.

Figure \ref{fig:14}a shows the true color composites of example Sentinel-2 Level2A patches acquired over Agia Napa, while Figure \ref{fig:14}b shows the respective co-registered SfM-MVS refraction corrected aerial DSM. Figure \ref{fig:14}c demonstrates the respective LiDAR bathymetry from \cite{magicbathynet} which will be used for the evaluation of the predicted bathymetry. Finally, Figure \ref{fig:14}d presents the predicted bathymetry using the proposed model, trained on pairs of Sentinel-2 Level2A and co-registered SfM-MVS aerial DSM patches corrected by the refraction. Figure \ref{fig:15} follows the same structure, presenting however representative patches of the Puck Lagoon area. The experimental setup described in Section \ref{experimentalsetup} was used for training; however, Sentinel-2 patches were resized to 720x720, to match the spatial resolution of the SfM-MVS DSM.

Table \ref{beyond} shows the metrics derived for the predicted bathymetry, comparing two versions of the proposed Swin-BathyUNet: one trained on complete LiDAR depth data and the other on incomplete refraction-corrected SfM-MVS data. Both sets of results are evaluated against the reference LiDAR data for comparison. For the Agia Napa area, training with aerial refraction-corrected SfM-MVS data yields RMSE, MAE, and Std. values that are around 13\% higher than those obtained when using LiDAR data for training. This was expected because of the missing areas in the refraction-corrected SfM-MVS data. In the Puck Lagoon area, training with Sentinel-2 images and SfM-MVS data corrected for refraction, results in approximately 4\% lower RMSE, 8\% lower MAE, and 1\% lower standard deviation compared to the LiDAR-based training, where the metrics are slightly worse. The obtained results indicate that refraction-corrected SfM-MVS bathymetry can serve as essential reference data for SDB methods, reducing dependence on costly fieldwork and external LiDAR/MBES datasets while improving model applicability across modalities like Sentinel-2 Level-2A. Additionally, it is shown that Swin-BathyUNet performs robustly when trained with complete LiDAR depth data, further demonstrating its potential for standard SDB approaches. Combining refraction-corrected SfM-MVS bathymetry with satellite imagery works best when both are captured within a short time frame. This helps ensure that environmental factors such as lighting, weather, and water clarity remain consistent, minimizing discrepancies between the datasets and improving the accuracy of bathymetry predictions.

\begin{table}[h!]
  \setlength{\tabcolsep}{1.5pt}
  \renewcommand{\arraystretch}{1}
  \caption{Results for bathymetry prediction from Sentinel-2 Level 2A imagery using Swin-BathyUNet (in meters). Entities in bold indicate the best score.}
  \centering
  \label{archive_list}
  \begin{tabular}{m{23mm}>{\centering\arraybackslash}m{9mm}>{\centering\arraybackslash}m{9mm}>{\centering\arraybackslash}m{7mm}>{\centering\arraybackslash}m{9mm}>{\centering\arraybackslash}m{9mm}>{\centering\arraybackslash}m{7mm}}
\toprule 
  Training data                & \multicolumn{3}{c}{Agia Napa} & \multicolumn{3}{c}{Puck Lagoon} \\
    \cmidrule{2-7}
                     
                     & RMSE & MAE& Std. & RMSE & MAE & Std.\\
\midrule
  LiDAR&\textbf{0.60}&\textbf{0.42}&\textbf{0.59}&0.35&0.26&0.34 \\
\midrule
  Aerial SfM-MVS &0.69&0.48&0.69&\textbf{0.34}&\textbf{0.24}&0.34 \\
\bottomrule
  \end{tabular}
\vspace{-0.1in}
\label{beyond}
\end{table}

\begin{figure*}[h!]
\vspace{-0.05in}
  \setlength{\tabcolsep}{1.5pt}
  \renewcommand{\arraystretch}{1}
  \footnotesize
  \centering
  \begin{tabular}{cccc}
    \begin{minipage}[c]{0.425\columnwidth}
 \centering
    \includegraphics[width=\linewidth]{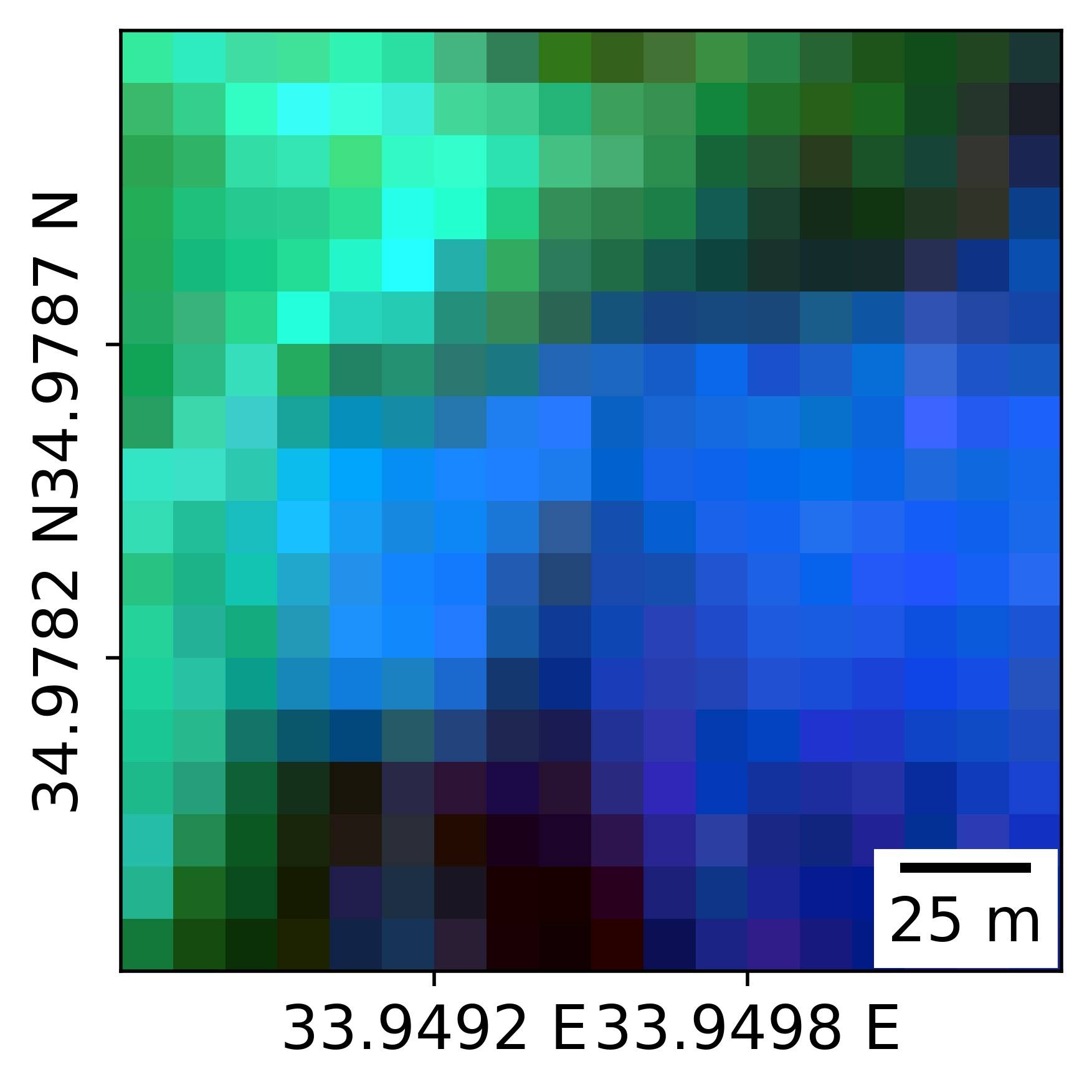}

    \end{minipage} 
    \hfill
    \begin{minipage}[c]{0.425\columnwidth}
        \centering
        \includegraphics[width=\linewidth]{img/sfm_410.jpg}

    \end{minipage}
    \vspace{2pt}
    \hfill
    \begin{minipage}[c]{0.425\columnwidth}
        \centering
        \includegraphics[width=\linewidth]{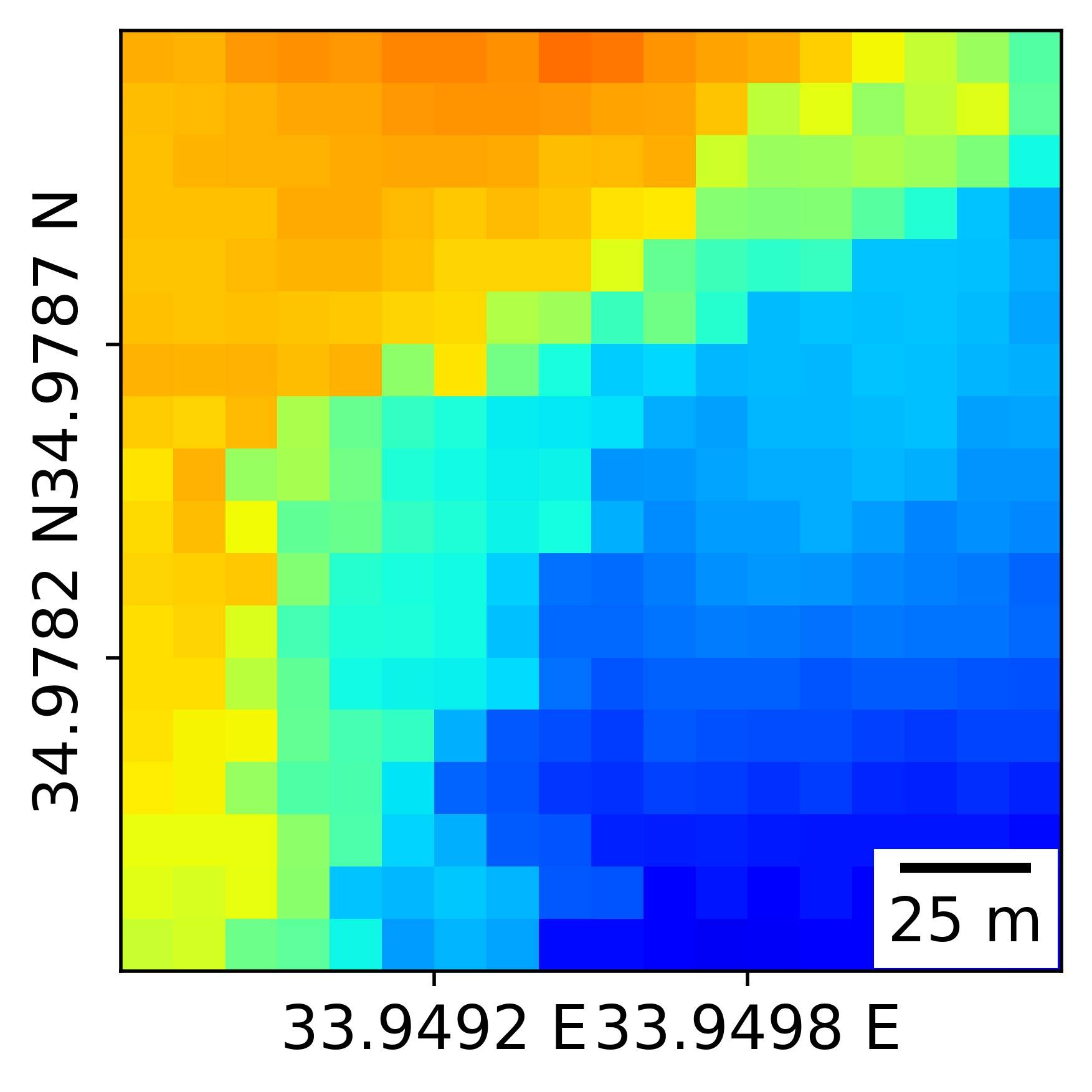}

    \end{minipage}
    \hfill
    \begin{minipage}[c]{0.525\columnwidth}
        \centering
        \includegraphics[width=\linewidth]{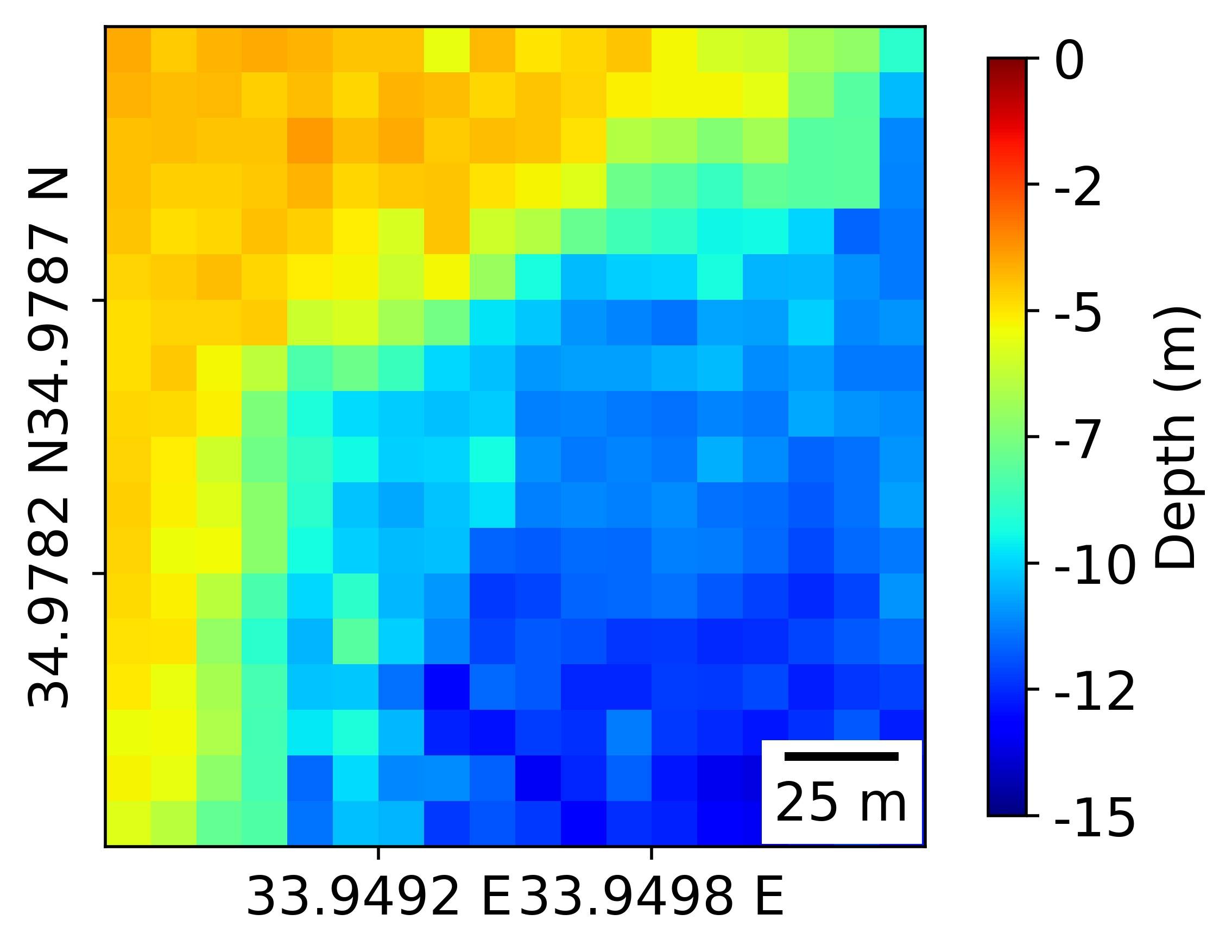}

    \end{minipage}\\
    \begin{minipage}[c]{0.425\columnwidth}
 \centering
    \includegraphics[width=\linewidth]{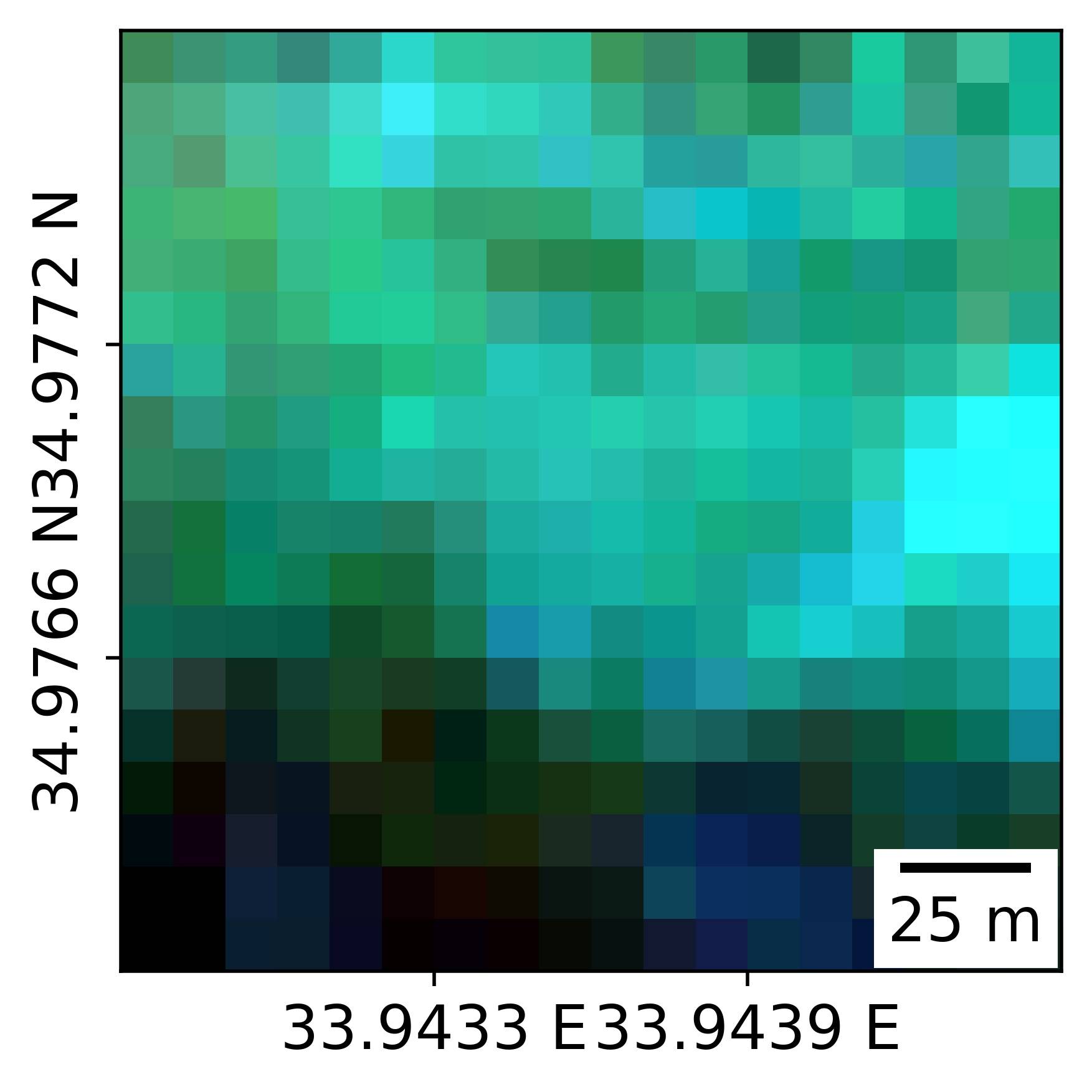}

    \end{minipage} 
    \hfill
    \begin{minipage}[c]{0.425\columnwidth}
        \centering
        \includegraphics[width=\linewidth]{img/sfm_379.jpg}

    \end{minipage}
    \vspace{2pt}
    \hfill
    \begin{minipage}[c]{0.425\columnwidth}
        \centering
        \includegraphics[width=\linewidth]{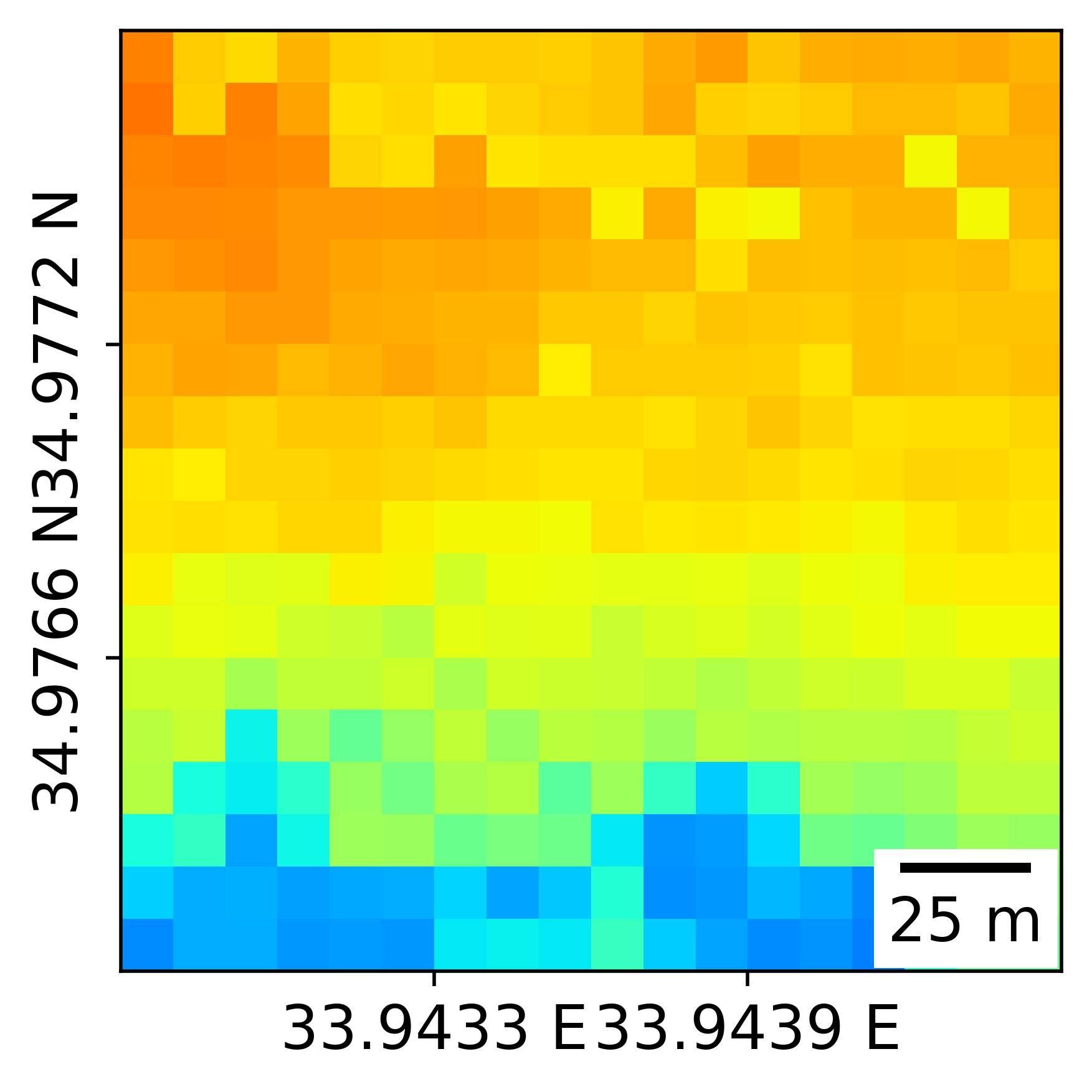}

    \end{minipage}
    \hfill
    \begin{minipage}[c]{0.525\columnwidth}
        \centering
        \includegraphics[width=\linewidth]{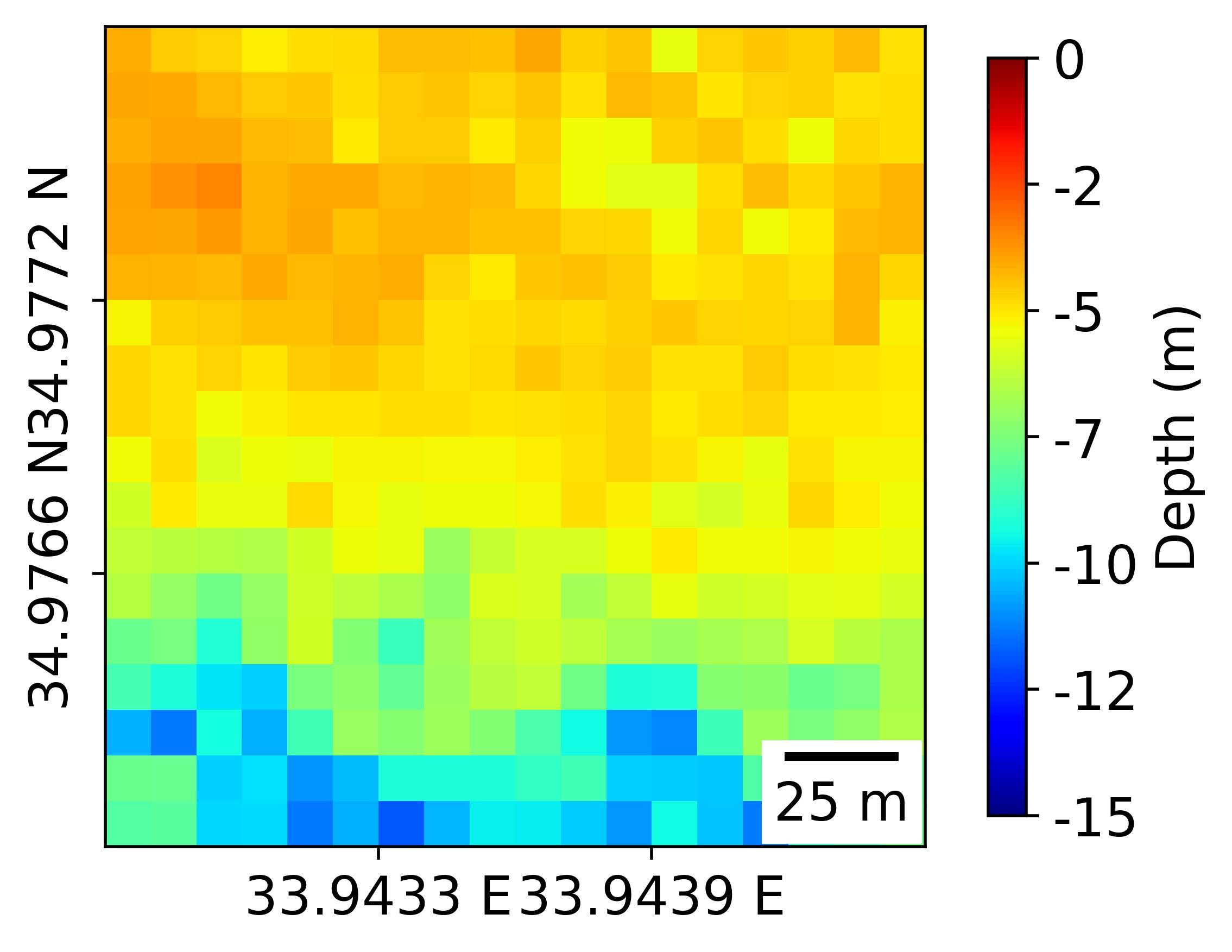}

    \end{minipage}\\
    \begin{minipage}[c]{0.425\columnwidth}
 \centering
    \includegraphics[width=\linewidth]{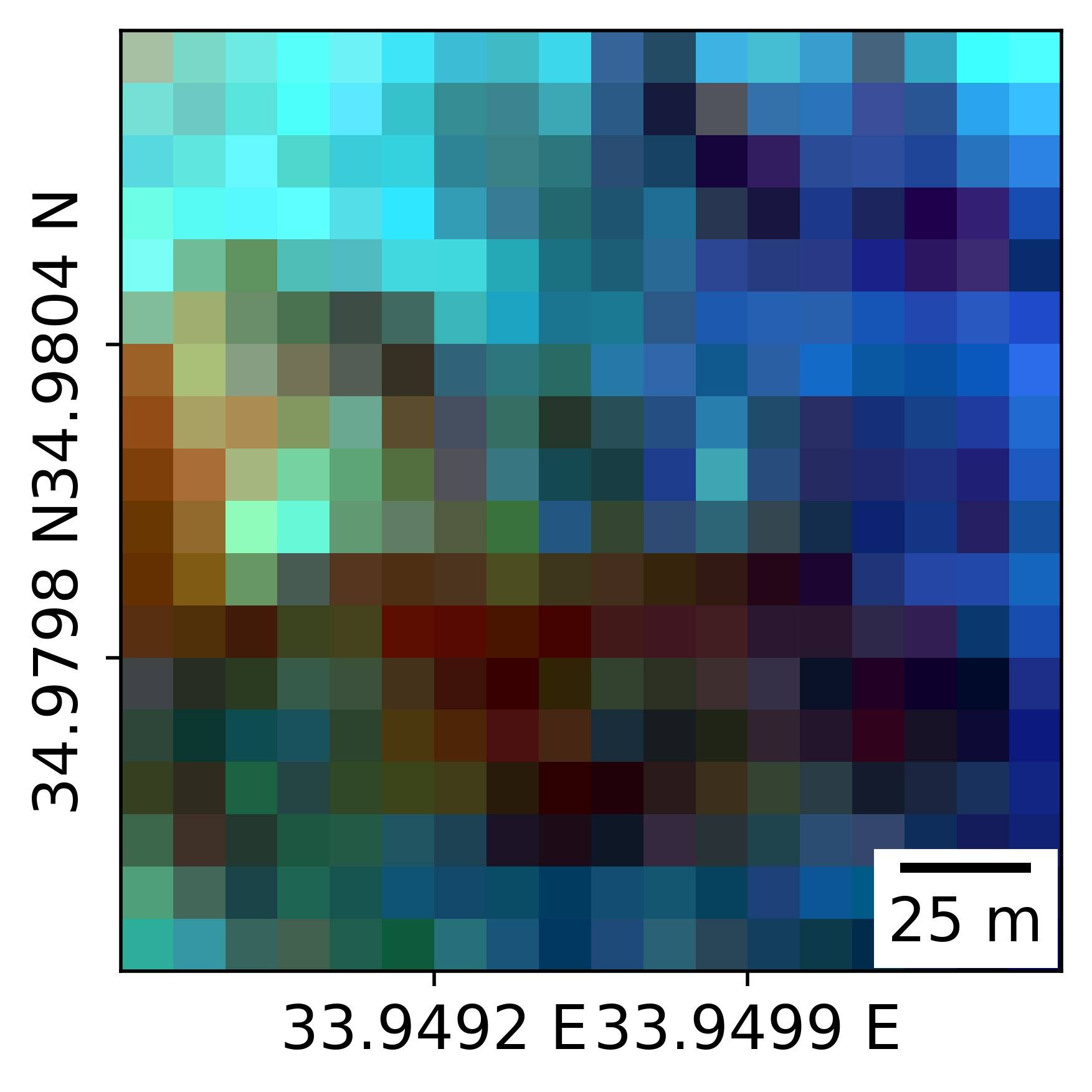}
    
        \textbf{(a)}
    \end{minipage} 
    \hfill
    \begin{minipage}[c]{0.425\columnwidth}
        \centering
        \includegraphics[width=\linewidth]{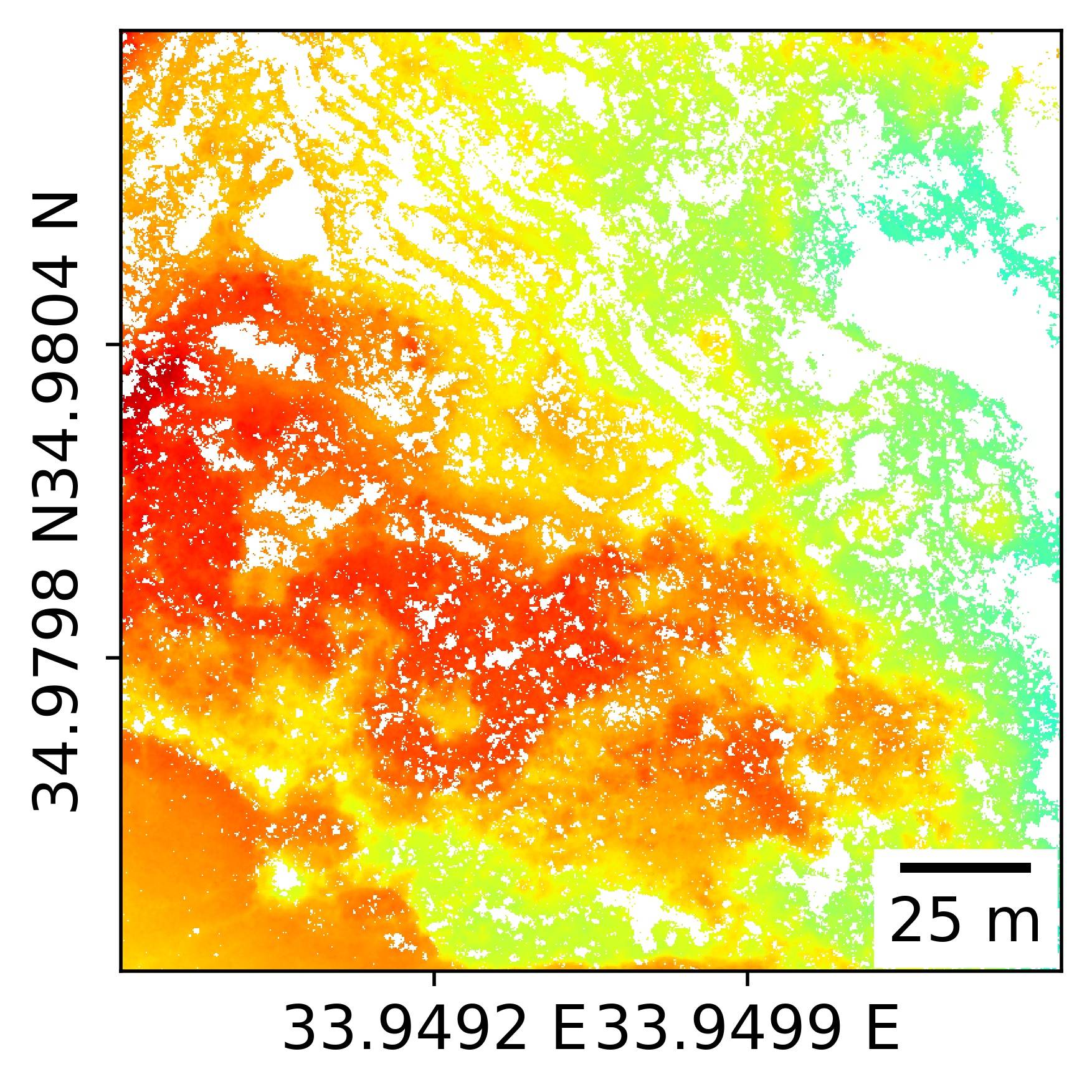}
        
        \textbf{(b)}
    \end{minipage}
    \vspace{2pt}
    \hfill
    \begin{minipage}[c]{0.425\columnwidth}
        \centering
        \includegraphics[width=\linewidth]{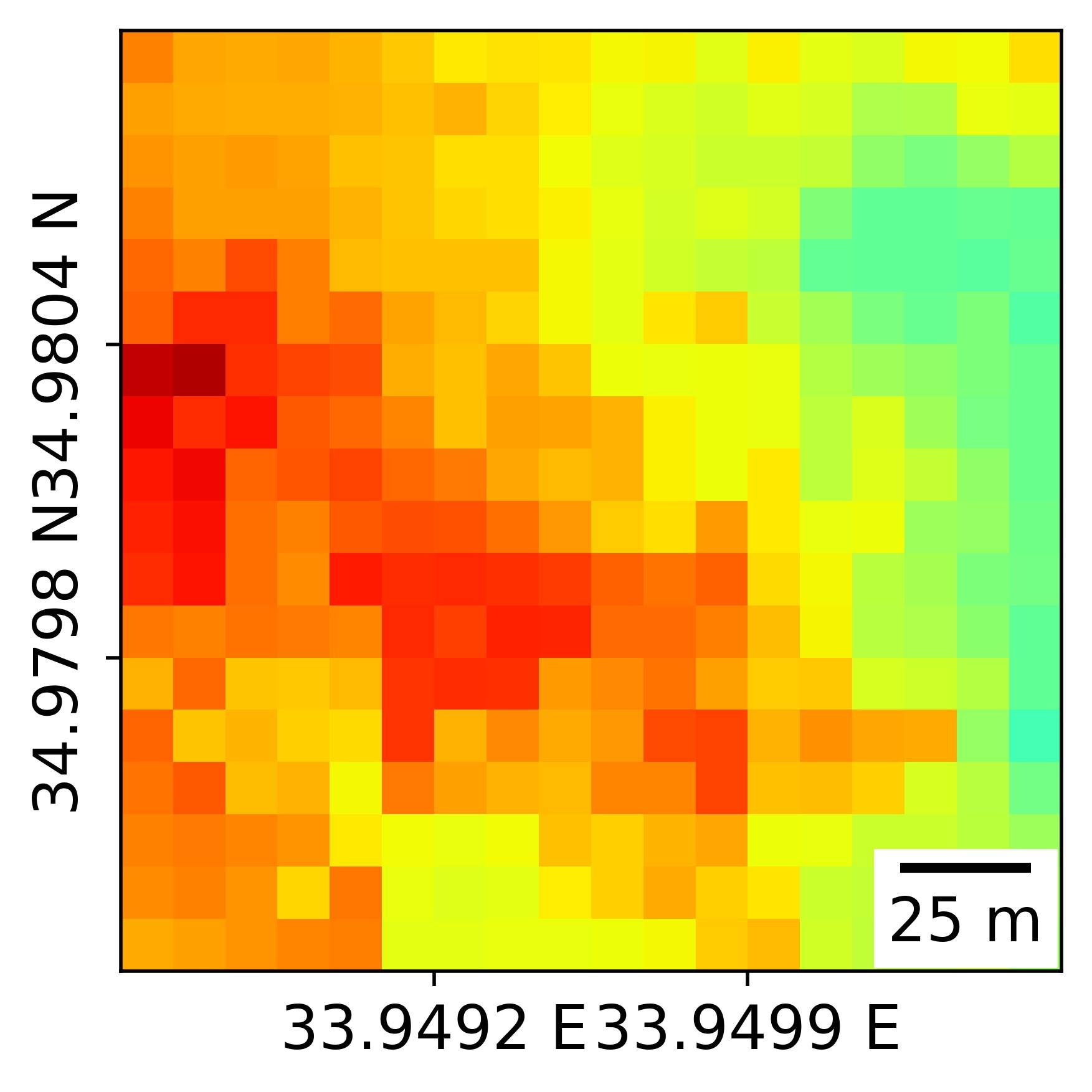}
        
        \textbf{(c)}
    \end{minipage}
    \hfill
    \begin{minipage}[c]{0.525\columnwidth}
        \centering
        \includegraphics[width=\linewidth]{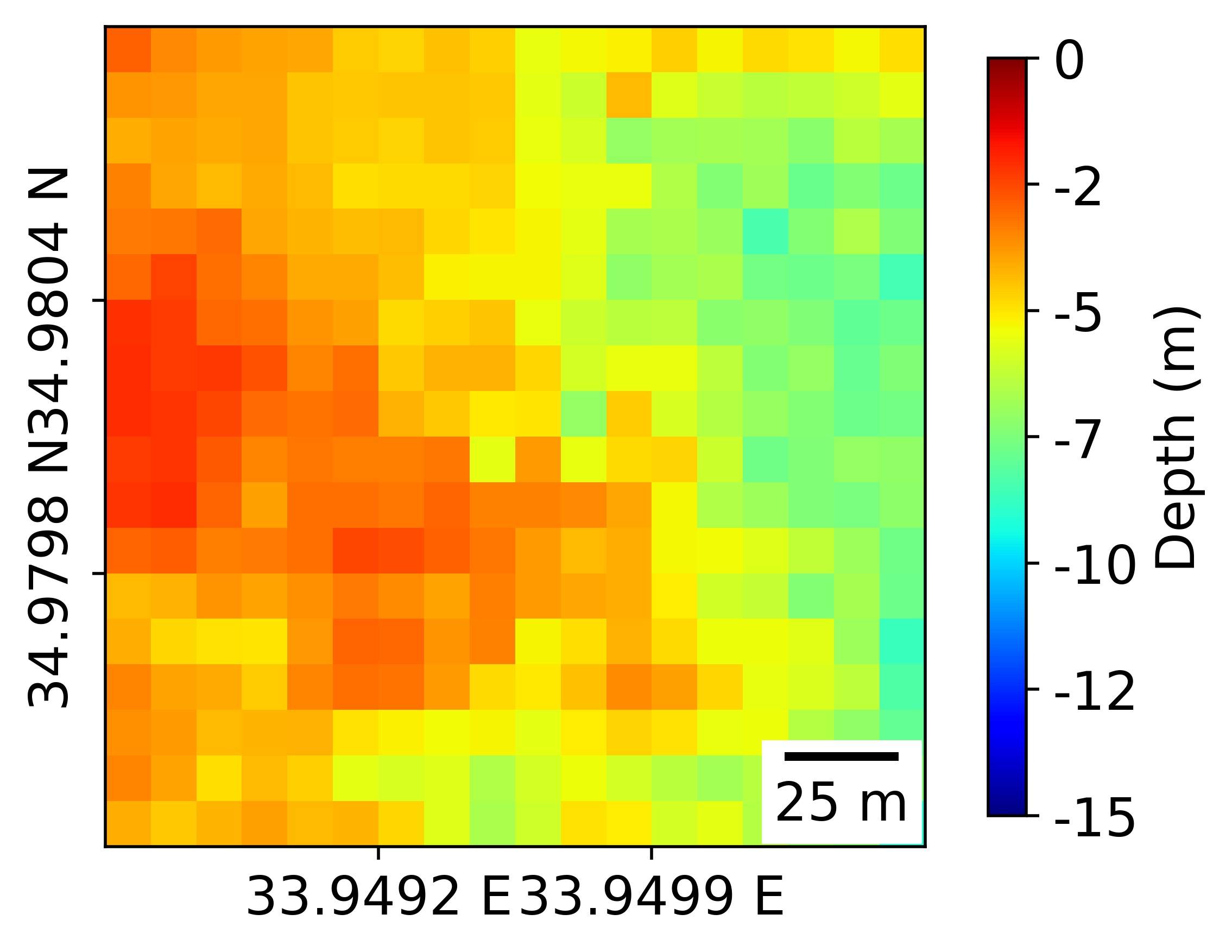}
        
        \textbf{(d)}
    \end{minipage}

  \end{tabular}
    \vspace{-0.09in}
  \caption{(a) True color composite of example patches acquired over Agia Napa from Sentinel-2, (b) the SfM-MVS refraction corrected depths, (c) the reference depths, and (d) the predicted depths using (a) and (b) for training, relative to WGS ’84.}
  \label{fig:14}
\vspace{-0.1in}
\end{figure*}

\begin{figure*}[h!]
\vspace{-0.05in}
  \setlength{\tabcolsep}{1.5pt}
  \renewcommand{\arraystretch}{1}
  \footnotesize
  \centering
  \begin{tabular}{cccc}
  
    \begin{minipage}[c]{0.425\columnwidth}
 \centering
    \includegraphics[width=\linewidth]{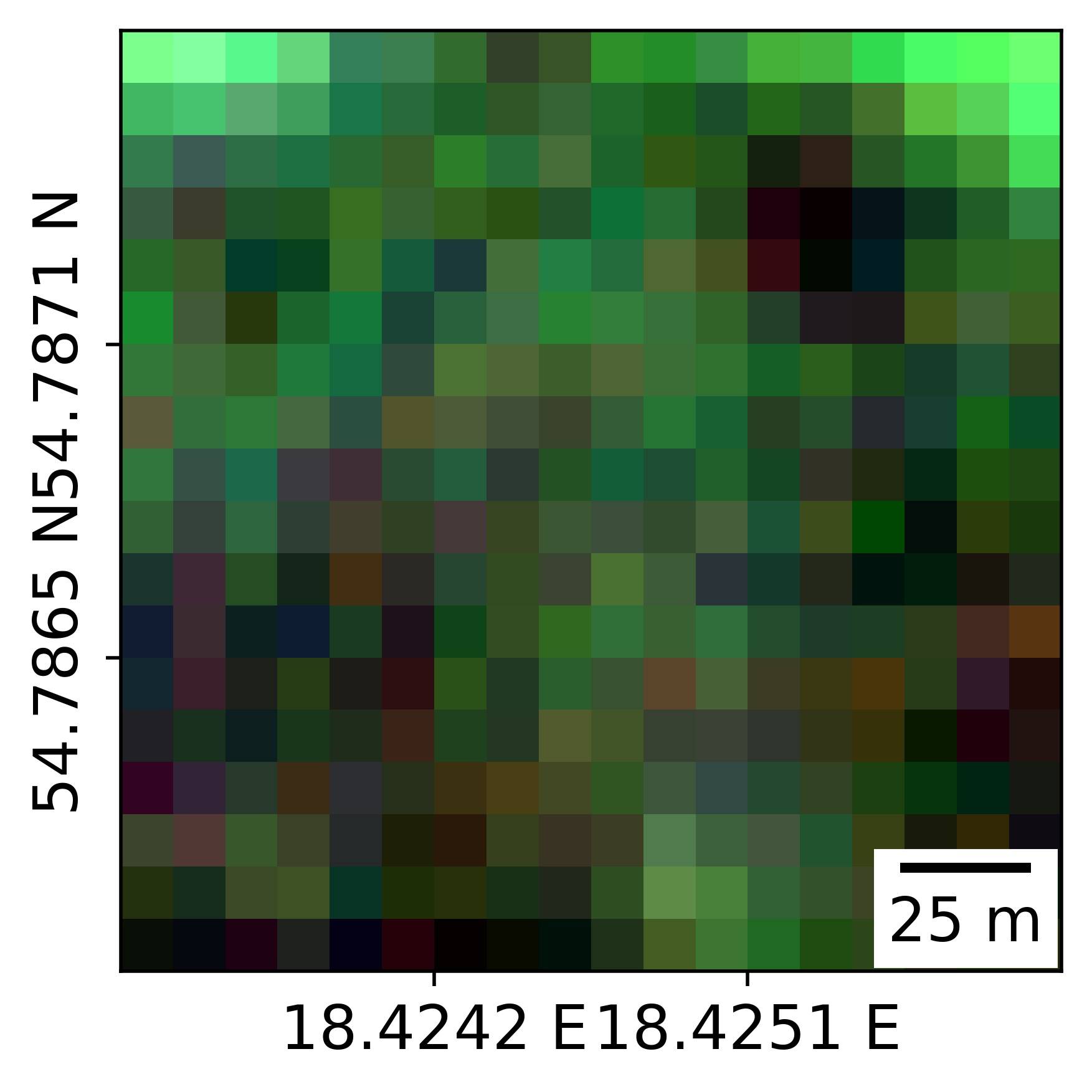}
    \end{minipage} 
    \hfill
    \begin{minipage}[c]{0.425\columnwidth}
        \centering
        \includegraphics[width=\linewidth]{img/sfm_1.jpg}
    \end{minipage}
    \vspace{2pt}
    \hfill
    \begin{minipage}[c]{0.425\columnwidth}
        \centering
        \includegraphics[width=\linewidth]{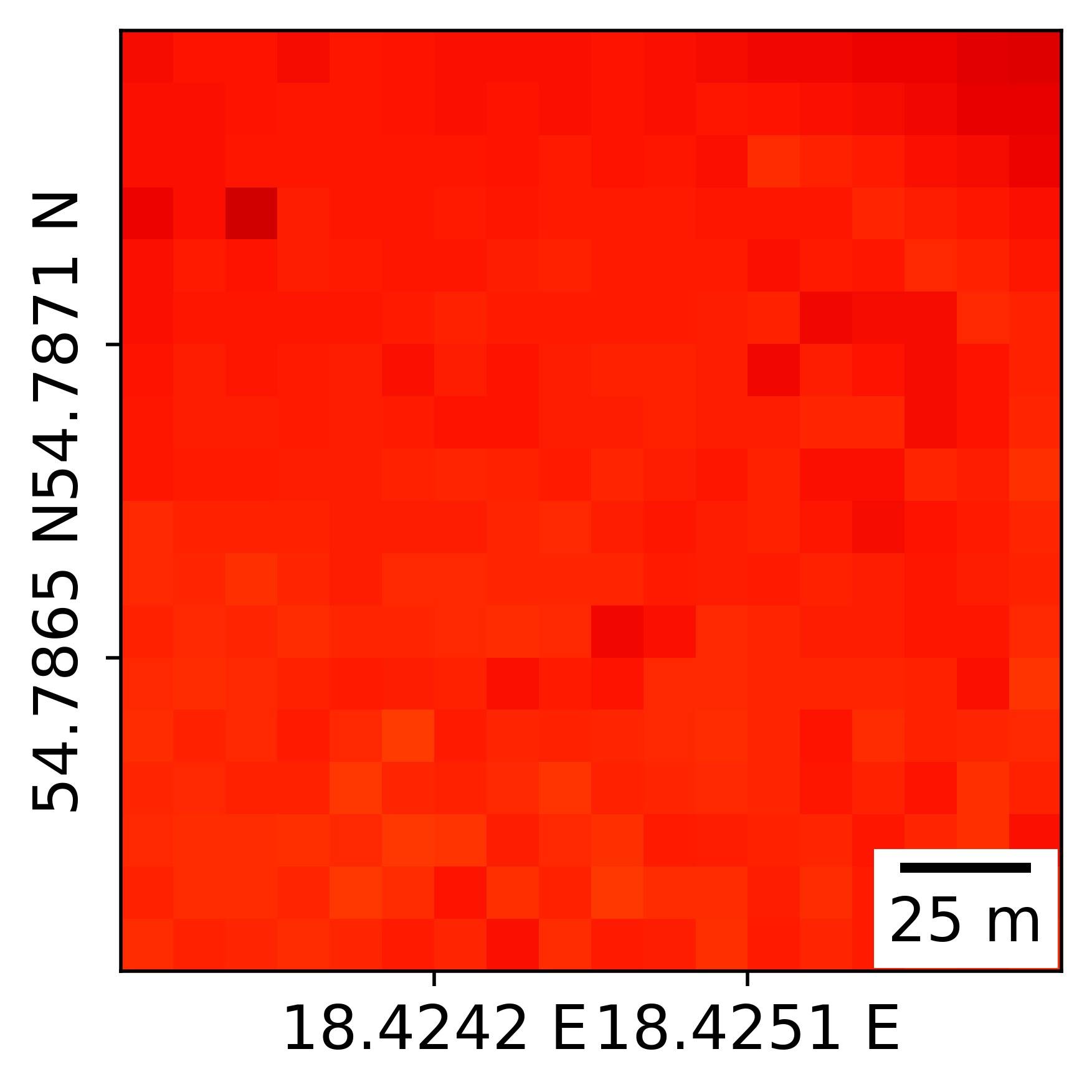}
    \end{minipage}
    \hfill
    \begin{minipage}[c]{0.525\columnwidth}
        \centering
        \includegraphics[width=\linewidth]{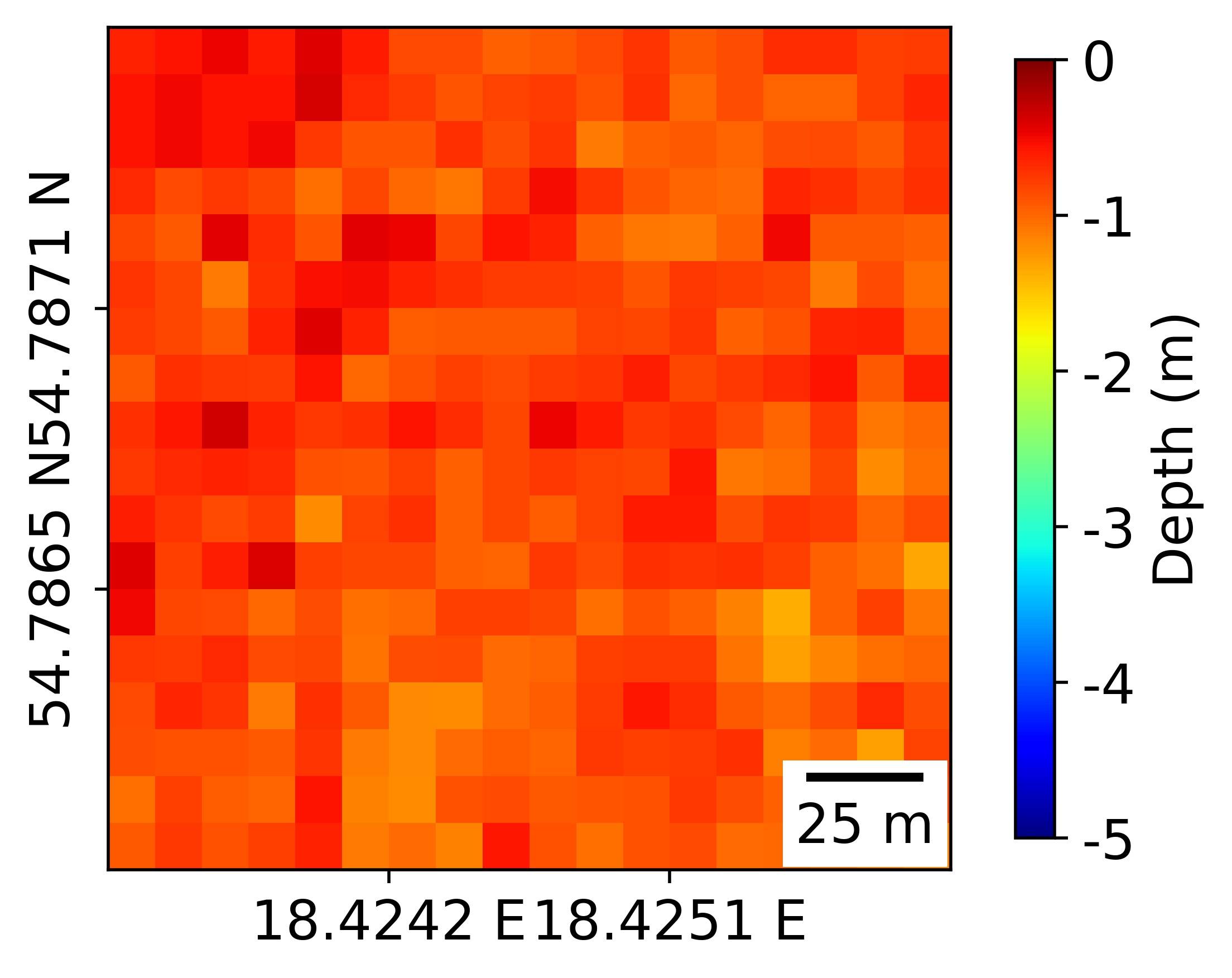}
    \end{minipage}\\
    
     \begin{minipage}[c]{0.425\columnwidth}
 \centering
    \includegraphics[width=\linewidth]{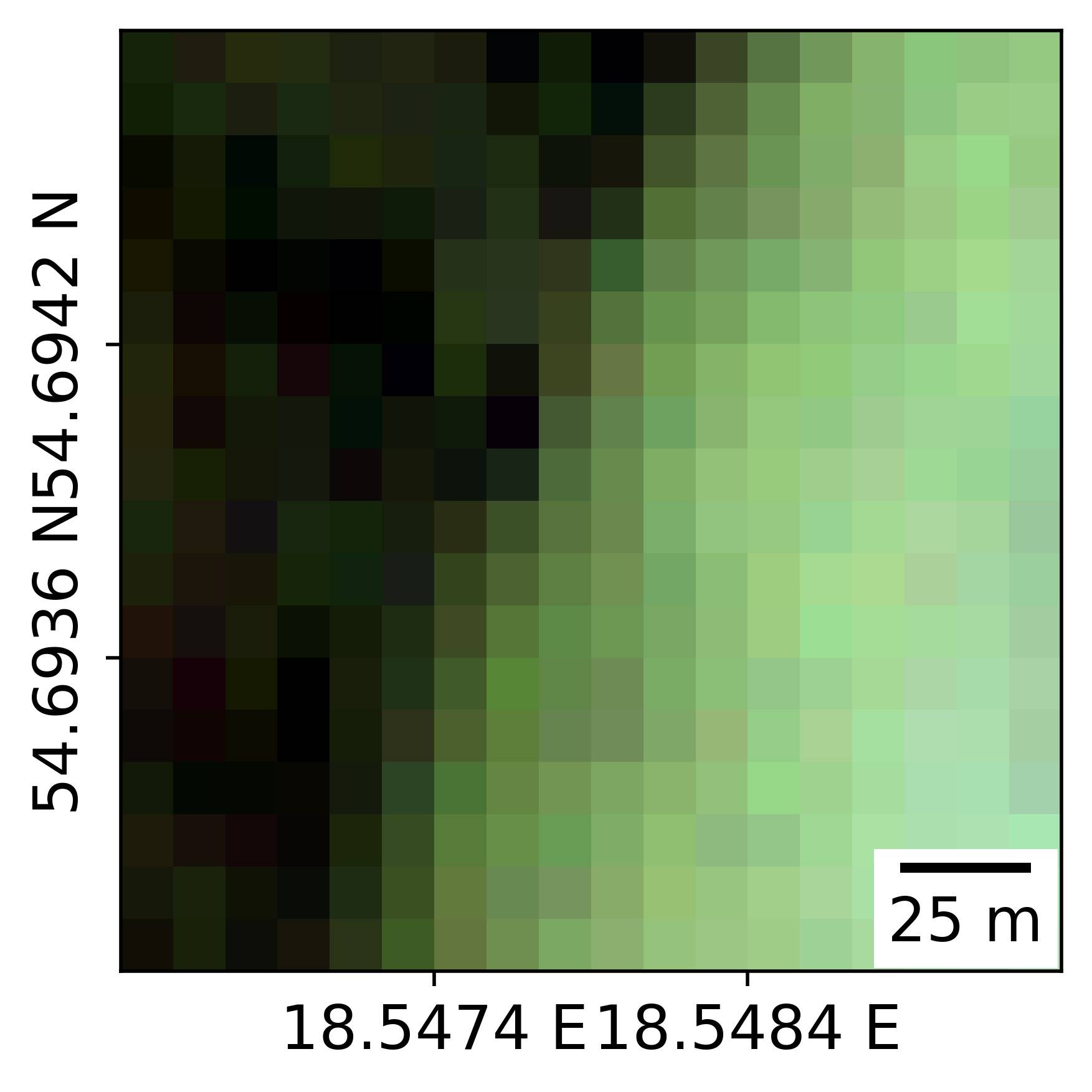}
    \end{minipage} 
    \hfill
    \begin{minipage}[c]{0.425\columnwidth}
        \centering
        \includegraphics[width=\linewidth]{img/sfm_2873.jpg}
    \end{minipage}
    \vspace{2pt}
    \hfill
    \begin{minipage}[c]{0.425\columnwidth}
        \centering
        \includegraphics[width=\linewidth]{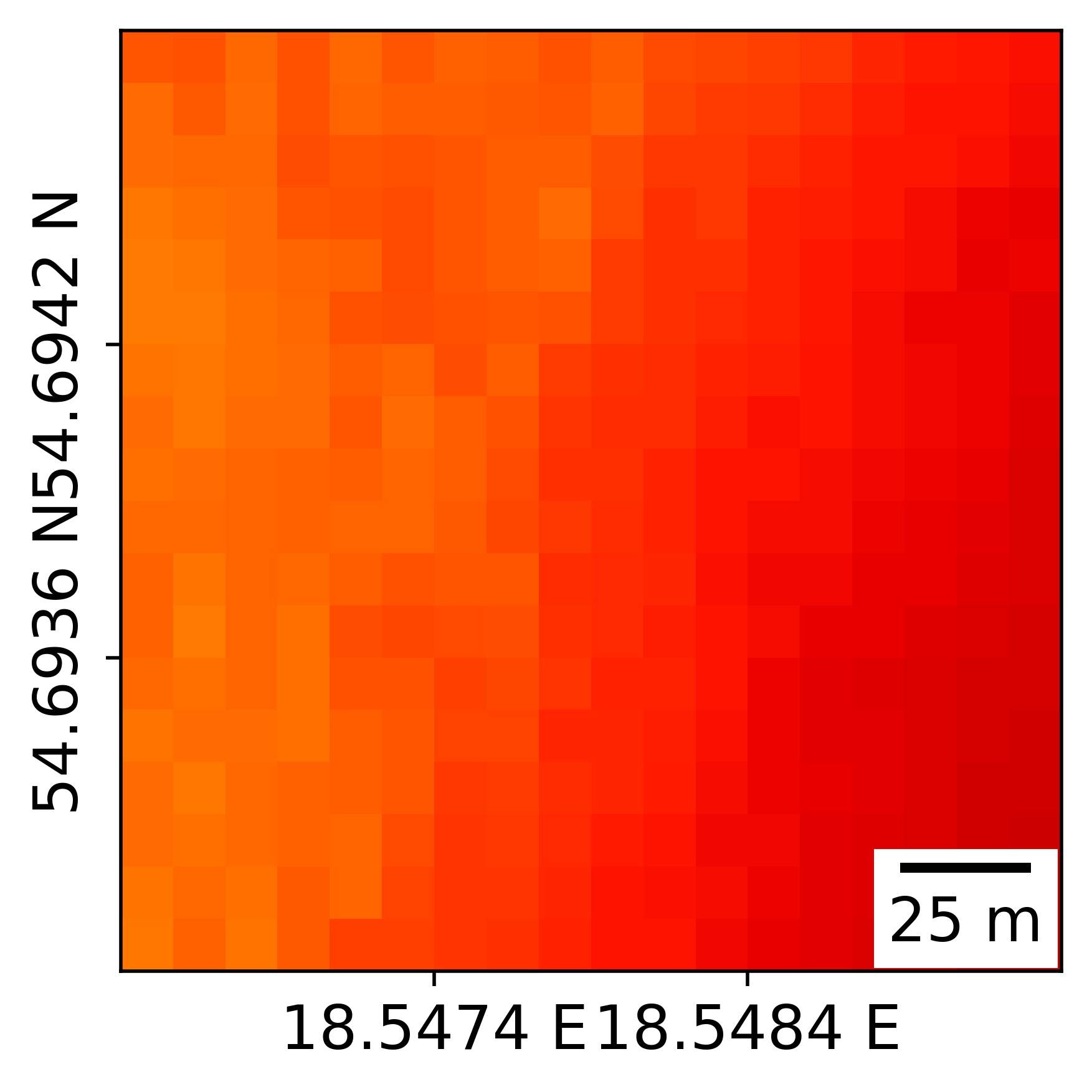}
    \end{minipage}
    \hfill
    \begin{minipage}[c]{0.525\columnwidth}
        \centering
        \includegraphics[width=\linewidth]{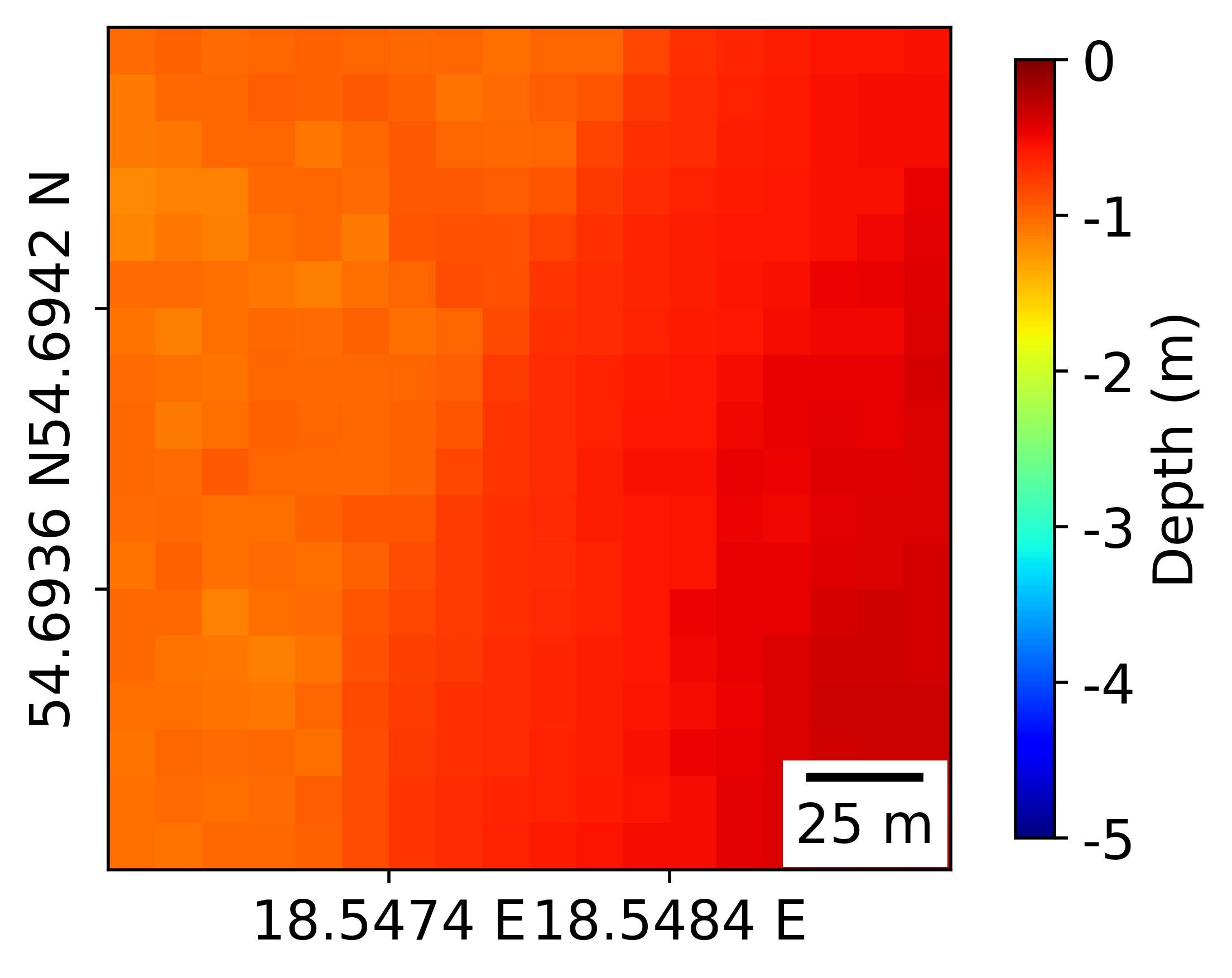}
    \end{minipage}\\
    
     \begin{minipage}[c]{0.425\columnwidth}
 \centering
    \includegraphics[width=\linewidth]{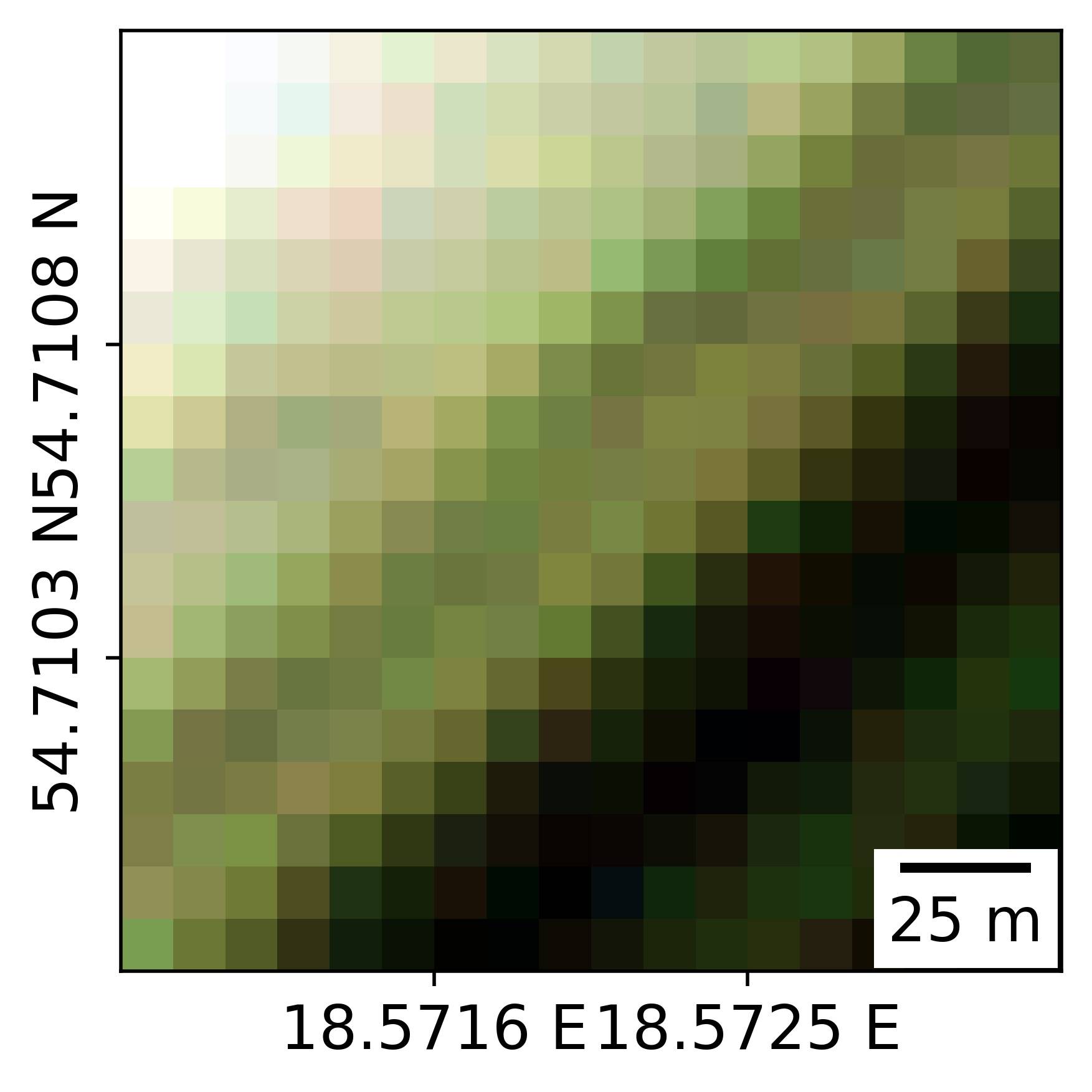}
    
        \textbf{(a)}
    \end{minipage} 
    \hfill
    \begin{minipage}[c]{0.425\columnwidth}
        \centering
        \includegraphics[width=\linewidth]{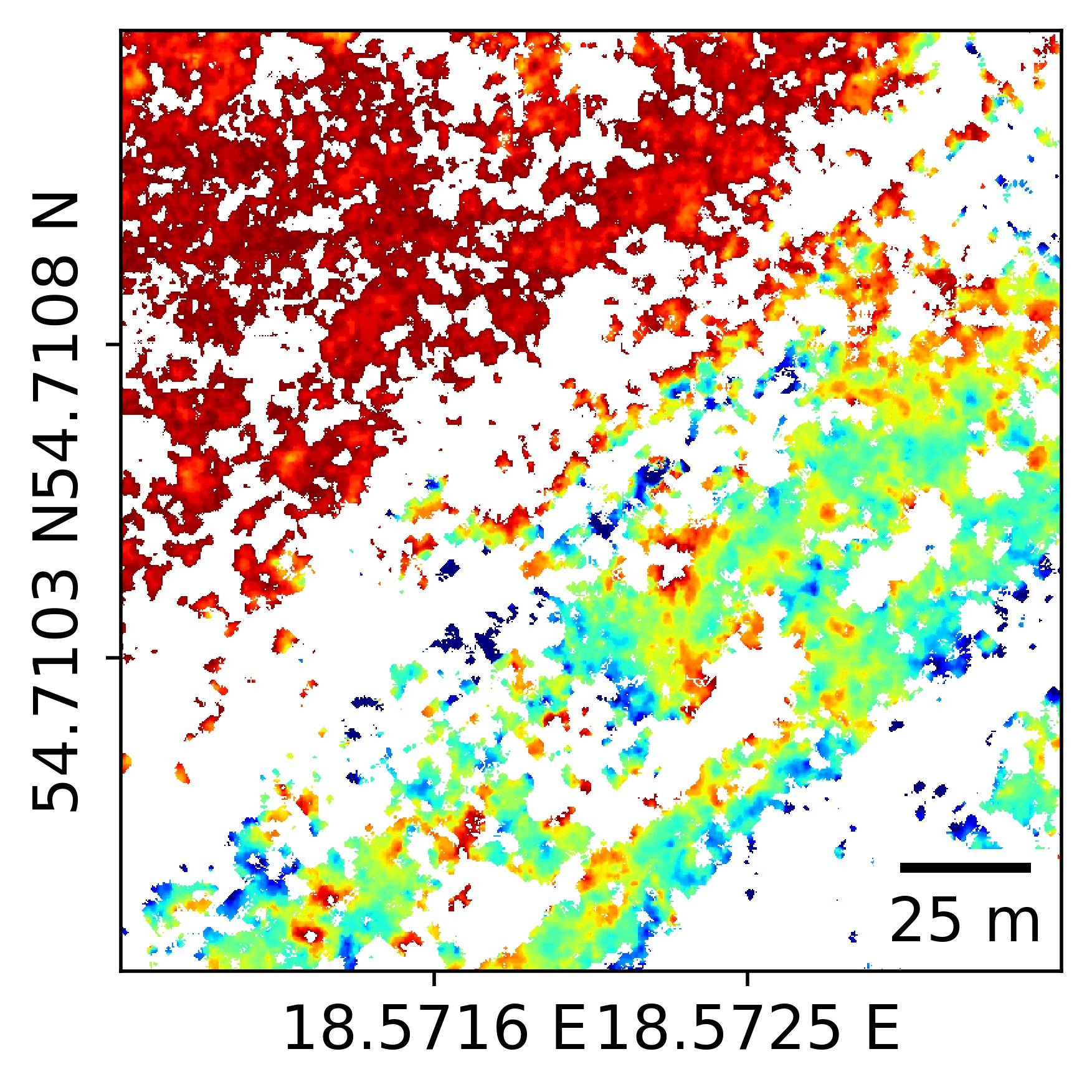}
        
        \textbf{(b)}
    \end{minipage}
    \vspace{2pt}
    \hfill
    \begin{minipage}[c]{0.425\columnwidth}
        \centering
        \includegraphics[width=\linewidth]{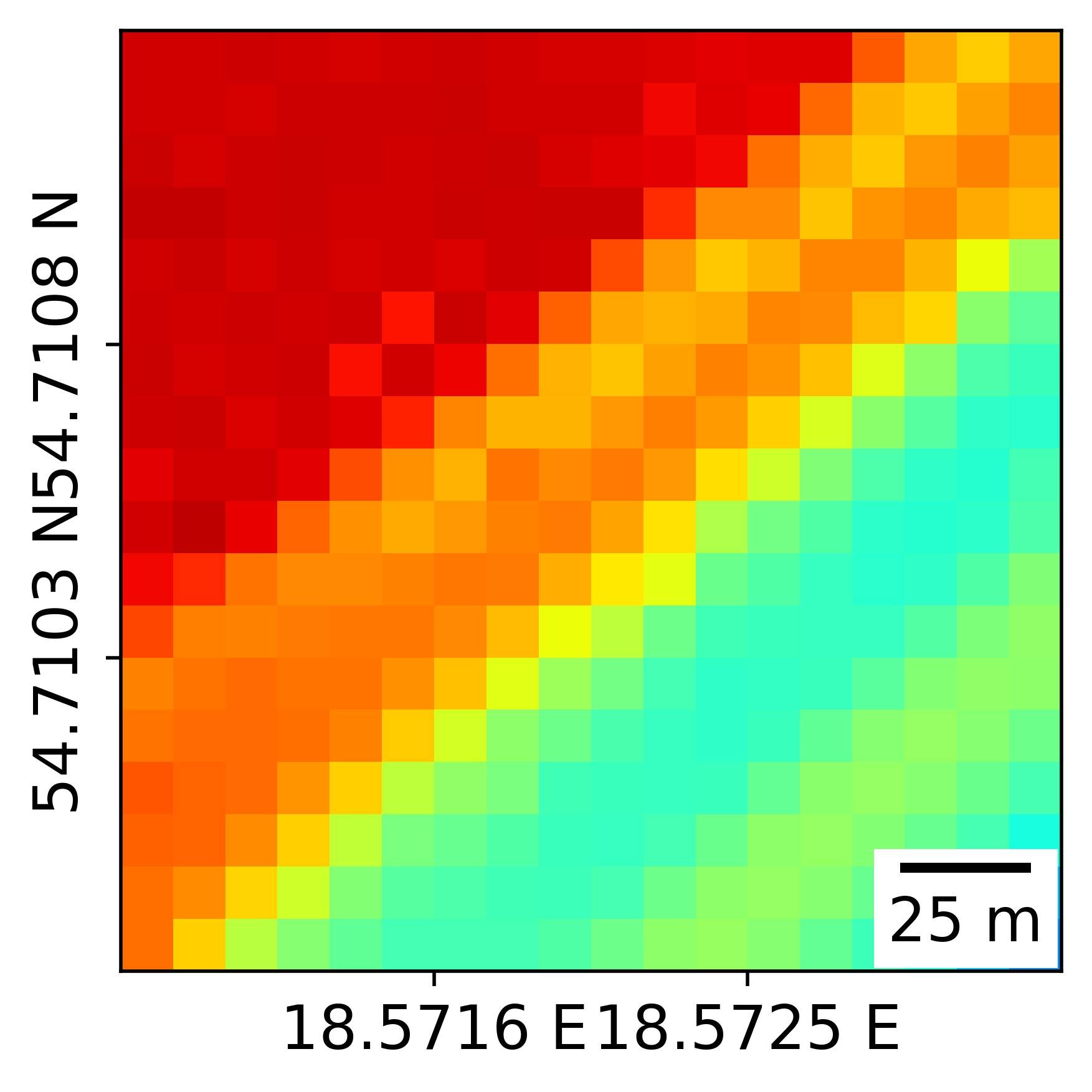}
        
        \textbf{(c)}
    \end{minipage}
    \hfill
    \begin{minipage}[c]{0.525\columnwidth}
        \centering
        \includegraphics[width=\linewidth]{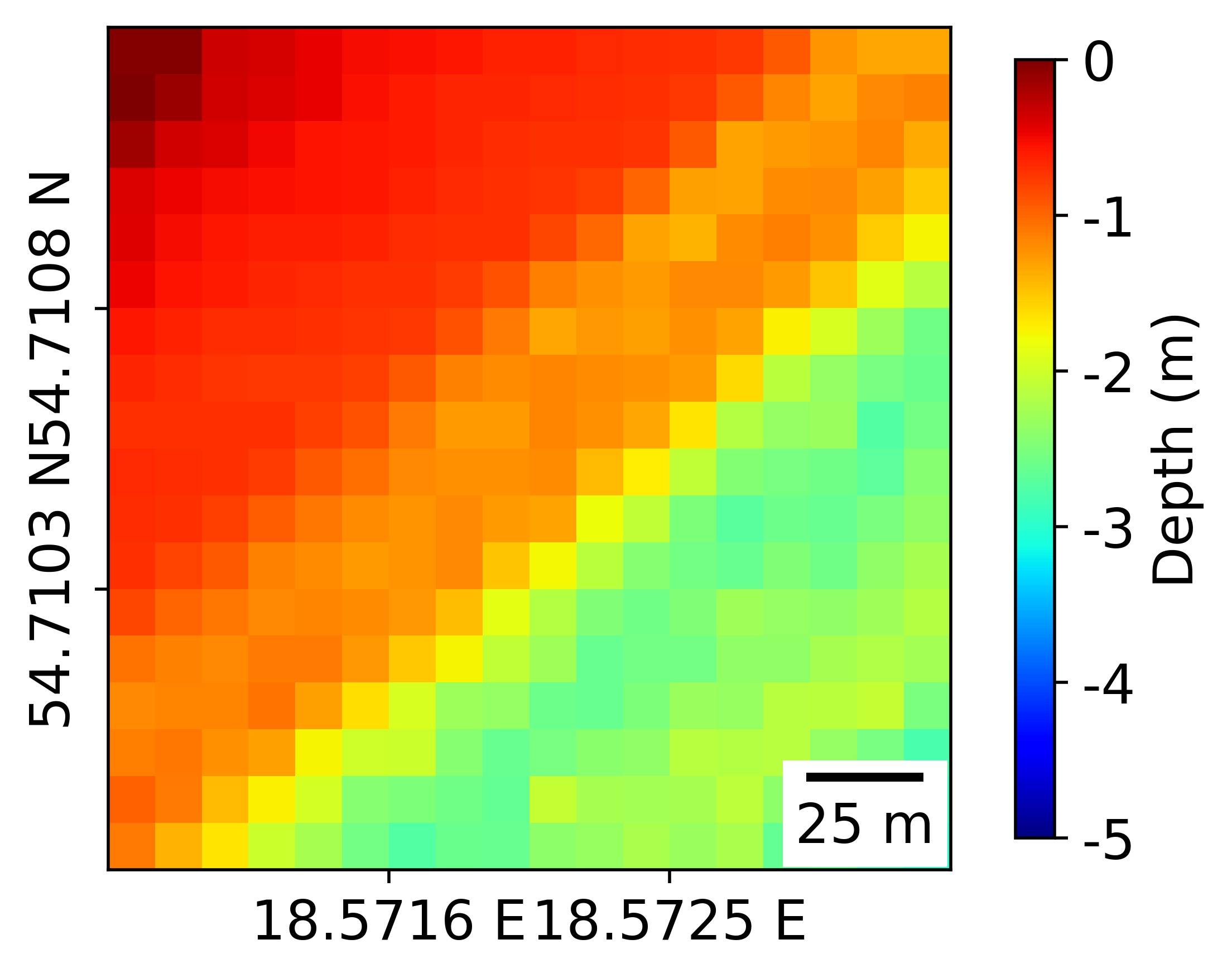}
        \textbf{(d)}
    \end{minipage}
  \end{tabular}
      \vspace{-0.09in}
  \caption{(a) True color composite of example patches acquired over Puck Lagoon from Sentinel-2, (b) the SfM-MVS refraction corrected depths, (c) the reference depths, and (d) the predicted depths using (a) and (b) for training, relative to WGS ’84.}
  \label{fig:15}
  \vspace{-0.1in}
\end{figure*}

\section{Discussion}
\label{section:Discussion}
The proposed approach increased the accuracy of the derived DSMs of the seabed, reduced the noise introduced by the SfM-MVS process, and increased the coverage and detail of the bathymetry. However, like any approach, it has some limitations, which are mostly related to the accuracy of the SfM-MVS-derived DSMs used for training and are discussed in this Section. Additionally, a more in-depth analysis of the remaining errors is provided.

\subsection{Increased accuracy and noise reduction}
The comparison between the results of the "Corrected SfM-MVS" and the "Swin-BathyUNet + BSW loss" in Tables \ref{table:table1.2} and \ref{table:table2.2} reveals significant reduction in large depth errors, exceeding both 1m and 0.50m. In Agia Napa, the number of predictions with depth error $>$ 1m decreased by 9\%, while the number of predictions with depth error $>$ 0.50m by 51.8\%. Similarly, in Puck Lagoon, the number of predictions with depth error $>$ 1m decreased by 95.3\%, while the the number of predictions with depth error $>$ 0.50m by 76.7\%. When comparing only the "non-gaps" areas in both sites, the "Swin-BathyUNet + BSW loss" method shows also a slight improvement in RMSE and MAE.

This improvement is due to the proposed Swin-BathyUNet model, which effectively predicts depths in areas where noisy measurements from the SfM-MVS process were not adequately addressed by the refraction correction. The refraction correction method relies solely on apparent depths as input, leading to persistent noise in the corrected SfM-MVS DSM. This noise remains because the SVR model processes these noisy input depths just like the other apparent depths. In contrast, the proposed Swin-BathyUNet utilizes the correlation of spectral values and the spatial relationships of neighboring pixels, predicting the correct depths and resulting in a significant reduction in large depth errors and noise. This reduction in noise is also evident in Figure \ref{fig:11} depicting the histograms of the depth differences calculated from the reference data. The results show a significant reduction in the width of the histograms for the predicted bathymetry which are depicted in green, compared to the ones of the refraction-corrected SfM-MVS bathymetry depicted in blue. This difference is more evident at higher frequency values. The narrower bell-shaped histograms indicate that the depth differences for the predicted bathymetry are more tightly clustered around the mean. This suggests that the predicted bathymetry has less variation in the depth differences and that the proposed approach effectively minimizes the noise typically present in the SfM-MVS depths in both sites. This reduction is also evident in the predicted depths of the indicative patches in Figures \ref{fig:fig10} and \ref{fig:fig9}. There, an obvious noise reduction can be noticed between the corrected SfM-MVS depths (Figures \ref{fig:fig9}b and \ref{fig:fig10}b) and the predicted depths in Figures \ref{fig:fig9}c and \ref{fig:fig10}c. Further evidence of noise reduction can be seen in the two bottom rows of Table \ref{table:table1} and Table \ref{table:table2}, where the combined prediction results in higher RMSE than the whole prediction. This reveals the remaining noise in the "non-gaps" areas, where the corrected SfM-MVS bathymetry remains unchanged.

\subsection{Enhanced coverage over SfM-MVS}
Tables \ref{table:table1.2} and \ref{table:table2.2} compare the bathymetric coverage between the SfM-MVS (both refracted and corrected) and the predicted depths using the proposed Swin-BathyUNet model. The results clearly show an increase in coverage of 43.51\% in the Agia Napa area and 12.66\% in Puck Lagoon. This is mainly because the learning-based SDB method can successfully estimate depths in uniform, featureless regions, where the feature-matching aspect of SfM-MVS methods often fails. It also performs well near the edges of the area of interest, where stereo coverage is limited. This advantage arises because the SDB method does not rely on viewing geometry or seafloor texture to generate accurate depth estimations. The difference in the increase is explained by the significantly greater depths in the Agia Napa area, which, combined with the presence of deep textureless regions and then poor viewing geometry, initially limit the ability of SfM-MVS to accurately estimate depths. 

\subsection{Enhanced detail over reference data}
By comparing Figure \ref{fig:fig9}a with \ref{fig:fig9}c and Figure \ref{fig:fig10}a with \ref{fig:fig10}c, a notable increase in the detail of the bathymetry becomes evident. This enhancement is attributed to the significantly smaller pixel size of the orthoimagery, which provides a higher-resolution prediction compared to the point spacing of the reference data. In the Agia Napa area, the point spacing ranges from 0.80m to 2m, which limits the granularity of the depth information. In contrast, while the point spacing in Puck Lagoon measures 0.20m, it is important to note that this value is derived after an interpolation step.

\subsection{Limitations}

\begin{figure*}[h!]
  \setlength{\tabcolsep}{1.5pt}
  \renewcommand{\arraystretch}{1}
  \footnotesize
  \centering
  \begin{tabular}{cc}
    \begin{minipage}[c]{0.95\columnwidth}
        \centering
        \includegraphics[width=\linewidth]{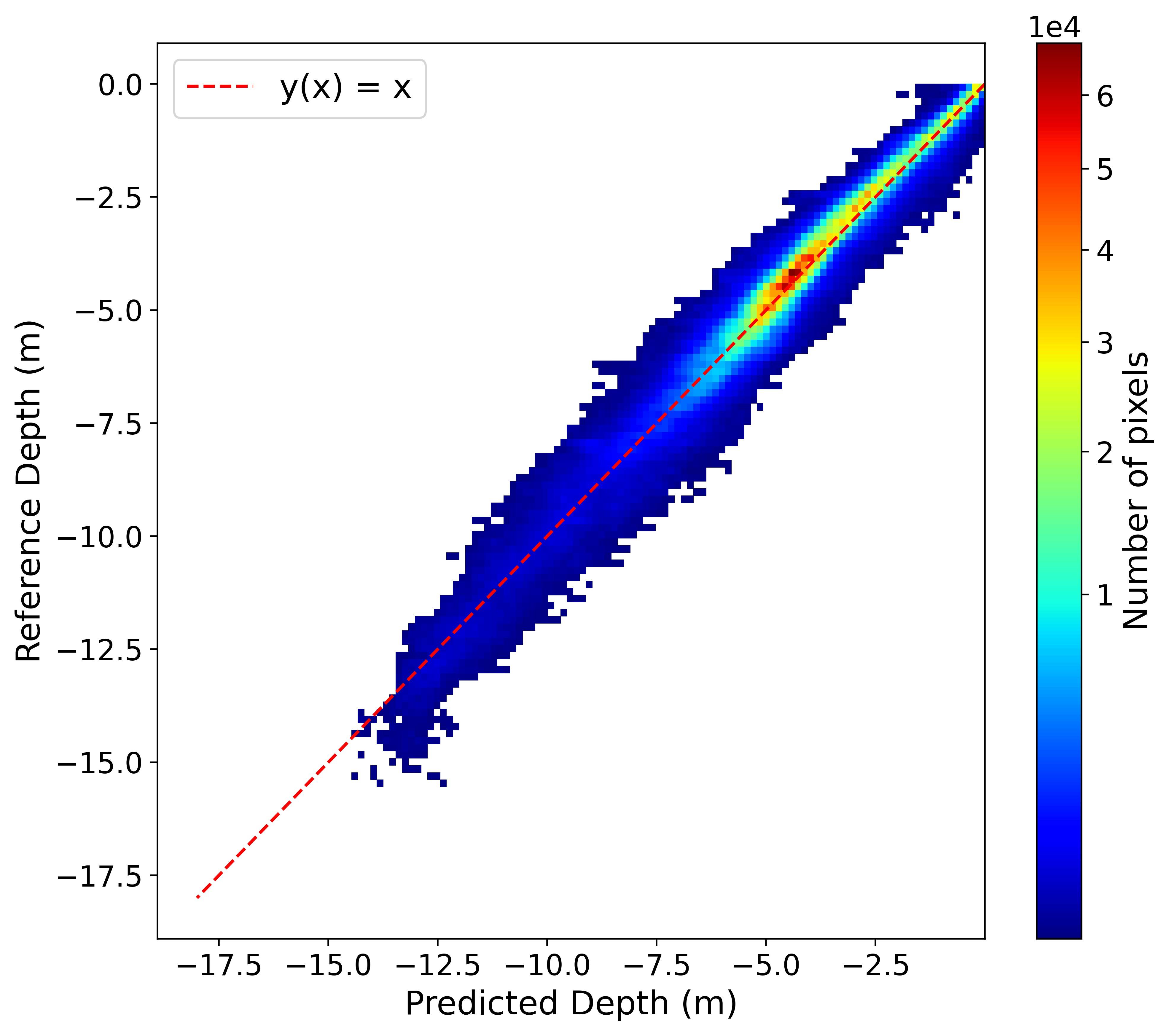}
    
        \textbf{(a)}
    \end{minipage} &
    \hfill
    \begin{minipage}[c]{0.9\columnwidth}
        \centering
        \includegraphics[width=\linewidth]{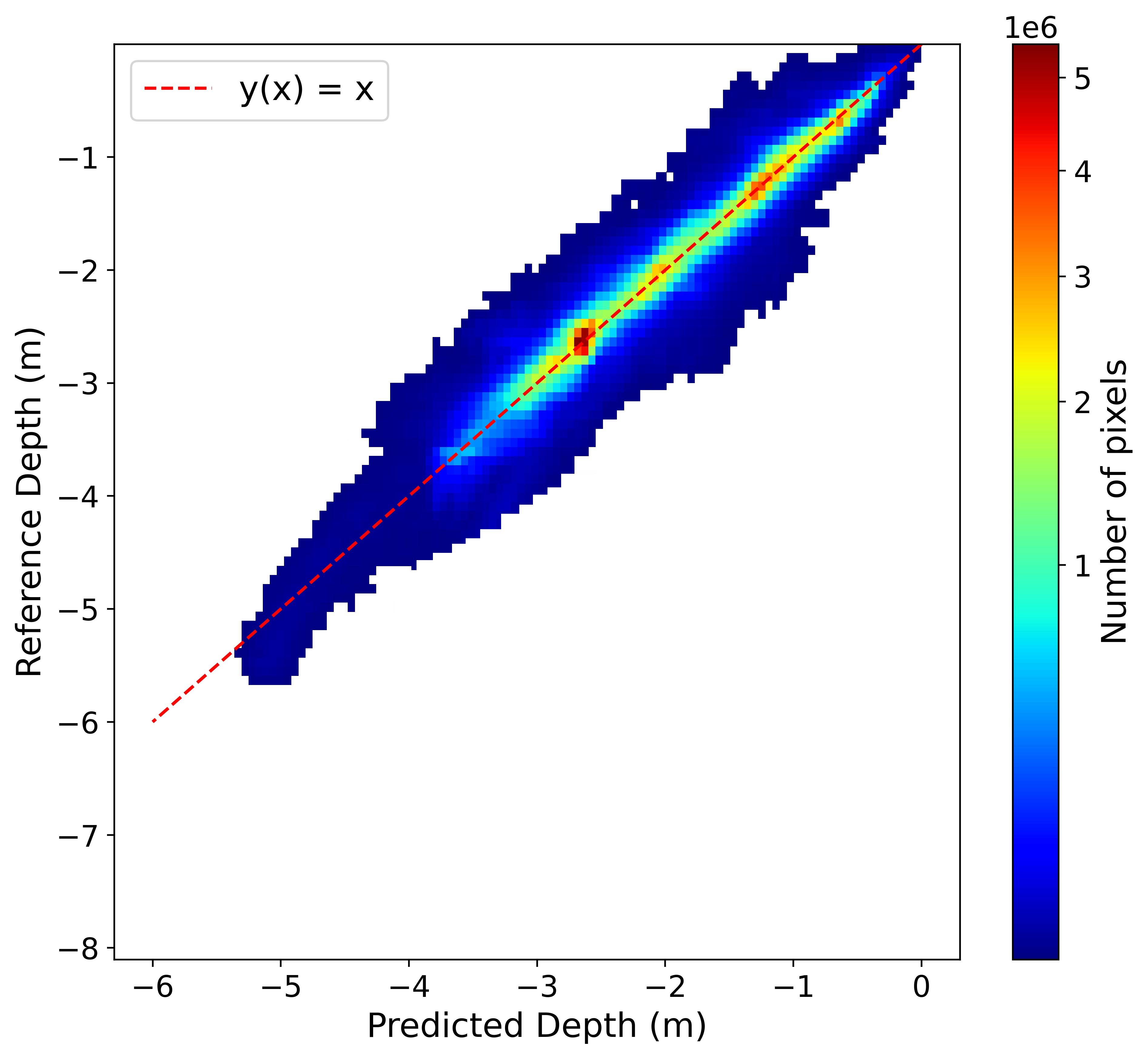}
        
        \textbf{(b)}
    \end{minipage}
  \end{tabular}
  \caption{Comparison between estimated shallow water bathymetry and
reference data for Agia Napa (a) and Puck Lagoon (b) areas. Red dashed line is the 1:1 line. Histograms are for the aerial modality.}
  \label{fig:13}
    \vspace{-0.15in}
\end{figure*}

\begin{figure*}[h!]
  \setlength{\tabcolsep}{1.5pt}
  \renewcommand{\arraystretch}{1}
  \footnotesize
  \centering
  \begin{tabular}{cc}
    \begin{minipage}[c]{0.9\columnwidth}
        \centering
        \includegraphics[width=\linewidth]{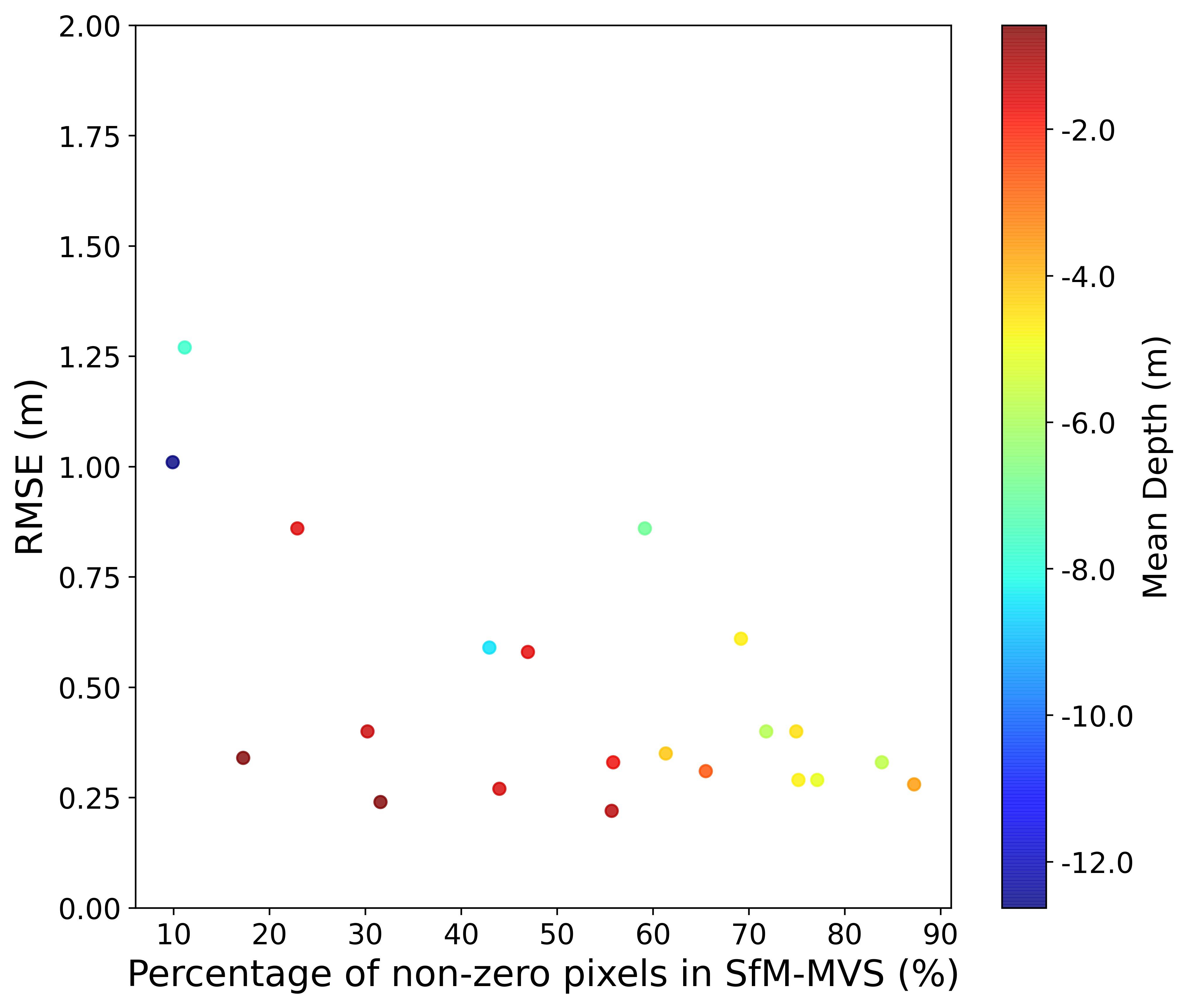}
    
        \textbf{(a)}
    \end{minipage} &
    \hfill
    \begin{minipage}[c]{0.9\columnwidth}
        \centering
        \includegraphics[width=\linewidth]{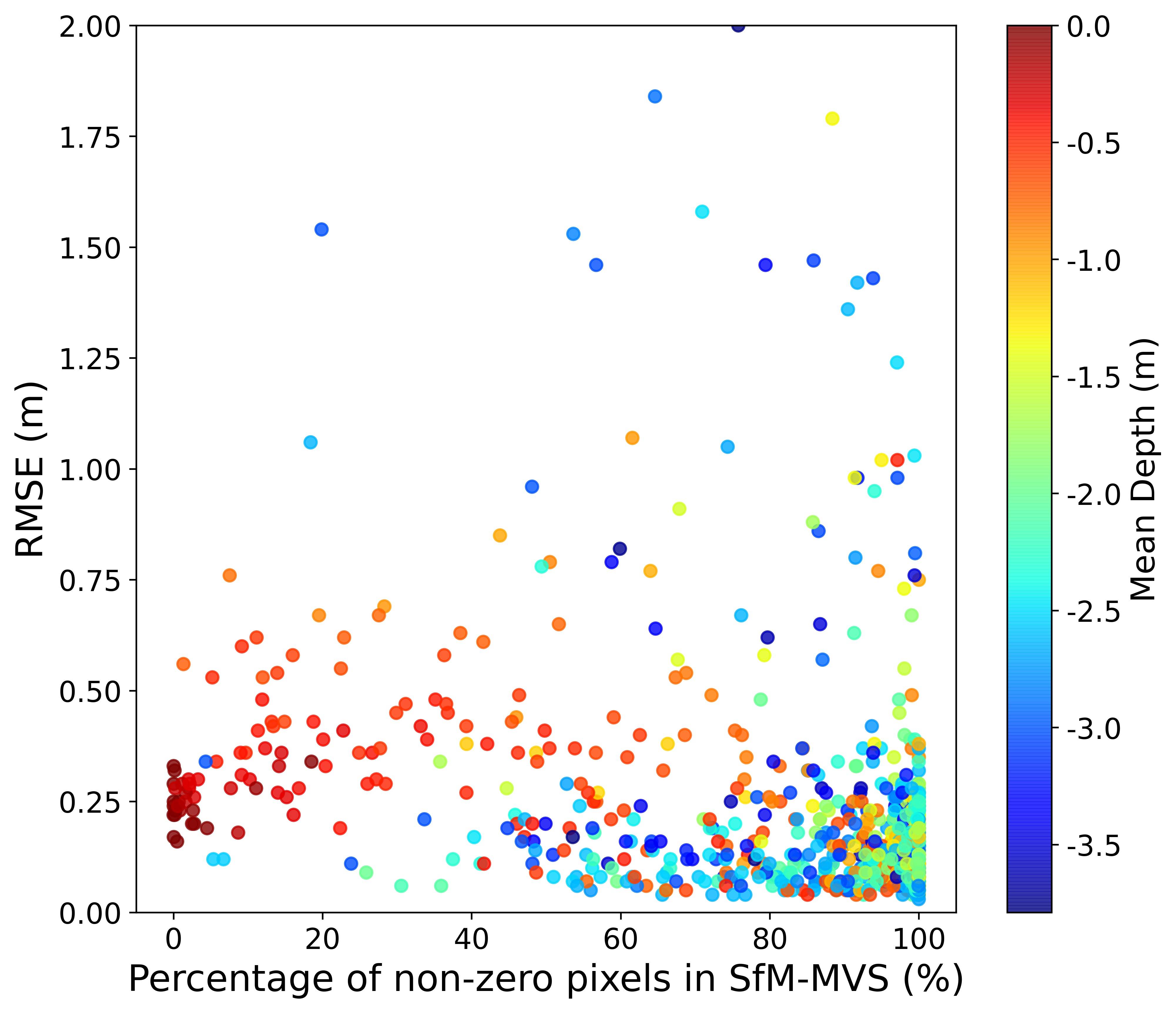}
        
        \textbf{(b)}
    \end{minipage}
  \end{tabular}
  \caption{Average RMSE of the estimated shallow water bathymetry in relation to the percentage of the non-zero pixels in SfM and the mean depth of the patch for Agia Napa (a) and Puck Lagoon (b) areas.}
  \label{fig:nonzero}
  \vspace{-0.15in}
\end{figure*}

Limitations in the presented approach are mostly related to the accuracy of the SfM-MVS-derived DSM. Figure \ref{fig:13} shows the 2D histograms of the comparisons between the predicted depths and the reference ones, not used for training, for the aerial modality. There, the red dashed line stands for the 1:1 line i.e. the optimal relation between reference and predictions. From the 2D histograms it is evident that these models are able to achieve a high performance in depth prediction till the depth of 13m in Agia Napa (Figure \ref{fig:13}a) and the depth of 5m in Puck Lagoon (Figure \ref{fig:13}b). The slightly deviating trend from the 1:1 line observed in the 2D histograms after those depths is related to the limitations imposed by the SfM-MVS discussed in Section \ref{III.B}. These limitations affect the depths used for training, particularly because depth data for those deeper areas were unavailable for training due to these constraints. In Agia Napa, the observed trend may also be attributed to the uneven distribution of GCPs, with deeper areas located 400-600m from shore-based GCPs. This distance can result in random $Z$-axis errors of up to 1m in the SfM process, which may either increase or decrease the apparent depths \citep{agrafiotis2019}. To eliminate the $Z$-axis errors, using drones equipped with built-in RTK sensors for precise camera positioning is suggested. Low image quality caused by suspended particles and turbidity during image acquisition can still be an additional source of error in the SfM-MVS data used for training, adding to the errors already discussed. In such cases, numerous outliers may arise, particularly in extremely shallow nearshore waters i.e. the wave braking zone. Removing those outliers can significantly enhance the accuracy of SfM-MVS bathymetry, however it is a time demanding process. Overall, the impact of these errors is minimal, as they occur in only a few evaluated pixels, as shown by the dark blue color in Figure \ref{fig:13}, which represents low density.

\subsection{Impact of SfM-MVS data distribution on depth estimation accuracy} 
In this subsection, we discuss the potential limitations of our method in areas with sparse SfM-MVS coverage, particularly in deeper or turbid regions where SfM-MVS may fail to capture sufficient depth information. As illustrated in Figure \ref{fig:nonzero}, a correlation between the percentage of non-zero pixels (i.e. annotated with SfM-MVS depth) and the RMSE is observed, where more annotated pixels generally result in lower RMSE, like in any statistical method. However, many cases of patches with 0-40\% of non-zero pixels having low RMSE exist too. Additionally, a strong correlation exists between the average depth of the patch and the number of annotated pixels: shallower regions tend to have more annotated pixels. This is most noticeable in the deeper sections of the study sites, where sand and seagrass are the dominant seabed types. These areas are especially susceptible to SfM failures and matching difficulties, as discussed in Section \ref{III.B}. While this correlation was expected, the results remain promising. In conjunction with Section \ref{Accuracy analysis by depth intervals and seabed habitats}, they suggest that depth and seabed class, have a greater influence on accuracy than the percentage of annotated pixels. However, it is important to note that depth and seabed characteristics also impact the percentage of pixels that can be successfully annotated. This interdependence suggests that, while annotation density alone does not dictate accuracy, it is indirectly shaped by the underlying environmental conditions. Despite these limitations, the proposed method shows strong performance even when annotations are sparse. As seen in Figure \ref{fig:nonzero}, the method continues to deliver depth estimates with reasonable accuracy, even if SfM-MVS coverage is extremely sparse. This shows that Swin-BathyUNet remains effective under a variety of conditions, including those with limited valid depth data, though its performance can vary depending on site-specific challenges.

\subsection{Residual errors} 
Several unrelated discrepancies between predicted and reference depths contribute to residual errors, as discussed below.

\subsubsection{Inherent errors in reference bathymetry} 
Residual errors, aside from the SfM-MVS noise resulting from refraction, water surface, and water visibility, can also be attributed to the precision of the reference data. In the Agia Napa area, based on \cite{agrafiotis2019}, the average standard deviation error of the LiDAR data (which is considered the reference bathymetry here) for the entire flight was estimated to be of the order of ±0.15m for depths up to 1.5 Secchi depth. For the Puck Lagoon, a much shallower site, based on \cite{puck}, the average standard deviation error of the LiDAR data for the scan lines of the entire flight was calculated to be 0.07m. 

\subsubsection{Asynchronous data and sea variability } 
In this study, image and bathymetry data were acquired with a small temporal difference of a few months. However, highly asynchronous data can introduce large residual errors if specific processing steps to detect and/or filter changes are not applied. Depending on the extent of these changes, deep learning models may account for them statistically; however, residual errors may still persist when comparing the results. While time differences in data acquisition may not significantly impact rocky seabeds, they can affect areas covered by seagrass, where changes in coverage, height, or distribution may have occurred. Additionally, shore-adjacent areas are prone to errors caused by nearshore erosion and/or accretion \citep{slocum2020combined}. Both refraction correction and learning-based SDB methods may exhibit elevation errors near the shoreline. These errors may be prominent in regions with active wave breaking and a highly variable water surface profile, as well as on sandy shore areas where wave run-up causes wetting and drying, altering the sand's color, making it resemble very shallow areas. Suspended sediment in the water column due to the above effect may also contribute to these errors affecting the predicted bathymetry.

\subsubsection{Varying data resolution} 
The varying levels of detail of the data introduced further discrepancies in depth comparison, potentially compromising the results presented. This issue is particularly evident when examining the upper left and lower left patches in Figure \ref{fig:fig9}. The predicted bathymetry captures seagrass depths and boundaries with significantly greater detail than the interpolated reference depths. It is important to note that in Agia Napa the reference depths have a point spacing ranging from 0.80m to 2m, while the spatial resolution of the predictions is 0.25m.

\subsection{Model complexity, inference time and real-time applicability}
To evaluate the applicability of the proposed model for real-time shallow water bathymetry retrieval applications, we assess the inference times for both small-scale (Agia Napa) and large-scale (Puck Lagoon) areas, as well as the model complexity and resulting accuracy, comparing them to the baseline (UNet). The results presented in Table \ref{time_table} indicate that, despite Swin-BathyUNet having a higher parameter count, thus requiring more offline time for training, the online inference time is almost the same for both models in both test sites respectively, suitable for near-real-time SDB applications. In more detail, for the Agia Napa test site, the time per image is 0.27 seconds for Swin-BathyUNet and 0.25 seconds for UNet. In Puck Lagoon, both models have similar high performance, needing just 0.14 seconds per image to predict the corresponding bathymetry. This shows that Swin-BathyUNet can efficiently handle large-scale datasets and still meet the near-real-time demands of operational SDB tasks. The relatively larger per-image times for the Agia Napa data, are attributed to overheads, such as model loading, which is constant across both sites and data preprocessing. These overheads are typically more noticeable in smaller datasets with fewer images. Notably, Swin-BathyUNet not only achieves higher accuracy, offsetting its larger parameter size and longer training time, but also maintains comparable inference times, making it suitable for near-real-time applications. To further improve near-real-time performance, optimizations such as efficient patch sampling, parallel processing, or cloud-based deployment of the model can be implemented.

\begin{table}[h!]
    \centering
    \caption{Inference time in seconds (s), number (\#) of trainable parameters in millions (M), and RMSE in meters (m) for Agia Napa (AN) and Puck Lagoon (PL) areas.}
    \begin{tabular}{@{}lcccc@{}}
    \toprule 
        {Metric} & \multicolumn{2}{c}{Swin-BathyUNet} & \multicolumn{2}{c}{UNet} \\ \cmidrule{2-5}
        & AN & PL & AN & PL\\
        \midrule 
        Inference Time (s) & 5.58 & 275.82 & 5.15 & 288.15 \\

        \# of Parameters (M) & 395 & 395 & 31 & 31\\
        RMSE (m) & 0.49 & 0.16 & 0.67 & 0.22\\
        \bottomrule 
    \end{tabular}
    \label{time_table}
    \vspace{-0.1in}
\end{table}

\section{Conclusion}
\label{section:Conclusion}

In this work, we presented a new approach that integrates the high-precision 3D reconstruction capabilities of SfM-MVS with advanced refraction correction techniques and the extensive coverage and spectral analysis features of a deep learning-based model to predict bathymetry. The proposed synergistic approach first relies on a state-of-the-art refraction correction method based on a linear SVR model trained on synthetic data and then exploits the proposed Swin-BathyUNet architecture, which integrates Swin Transformer blocks with both window-based self-attention and cross-attention mechanisms with U-Net. By this combination, the network can effectively capture both local fine details, such as small seabed structures, and global contextual information, such as large-scale depth variations and their relation to image colour values. The BSW loss function improved the model's ability to focus on challenging areas during training, leading to improved accuracy.

The experimental results in two different sites demonstrate the effectiveness of the proposed Swin-BathyUNet model and the proposed approach as a whole. While, the corrected SfM-MVS method yields RMSE, MAE, and Std. metrics that are comparable to those of Swin-BathyUNet; largely due to refraction correction bias, our approach achieves lower error values overall. More importantly, it significantly reduces large depth errors exceeding 0.50m and 1m,  with reductions ranging from 9\% to 35.3\%, compared to the SfM-MVS depths corrected for refraction used as input. Additionally, the vast majority of the predicted bathymetry meets the highest vertical accuracy standards of both S-57/S-101 and S-44, demonstrating Swin-BathyUNet’s suitability for various scientific and operational applications. Furthermore, Swin-BathyUNet enhances bathymetric coverage by estimating depths in featureless regions and near edges, where SfM-MVS struggles. The coverage increased by 43.51\% in Agia Napa and 12.66\% in Puck Lagoon. The method also delivers finer bathymetric details aided by the high resolution orthoimagery used for prediction and reduces noise compared to SfM-MVS. Swin-BathyUNet has also proven to work robustly with other types of training depth data, i.e. complete LiDAR and MBES depth data, highlighting the extended applicability of the model to standard bathymetric approaches utilizing full reference data. Despite its more parameters and longer training time, Swin-BathyUNet not only maintains comparable inference times, enabling close-to-real-time applications, but also delivers lower RMSEs, effectively counterbalancing this drawback.

One potential limitation of this approach is the quality of the initial SfM-MVS processing. This is mainly affected by the seabed's coverage and type, visibility restrictions due to depth, especially in deeper areas, and errors introduced by the wavy surface, the wave braking affects near the shore, the changing illumination conditions, the water clarity, and sun glint. However, in the experiments performed, the incomplete refraction-corrected SfM-MVS DSMs proved sufficient to train accurate learning-based bathymetric models and get reliable depth predictions. The initial SfM-MVS processing successfully resolved depths across the various bottom types. For example, both sites included small and large patches of sand. For some of the smaller ones, bathymetric data was provided by the SfM-MVS processing because the surrounding rocks and seagrass borders provided enough texture to produce valid but noisy depth information. This allowed the learning-based SDB model to learn the radiometric signature of each seafloor substrate at different depths, as well as to extrapolate and predict depths for areas that were not resolved in the initial SfM-MVS processing and are absent from the training dataset. While in this study, we focused on predicting the missing depth values using training data (SfM-MVS DSMs) from the same site, in future work, we plan to extend the approach to completely disjoint areas, which will require domain adaptation techniques to ensure scalability across diverse conditions.

By addressing key limitations of traditional SfM-MVS and SDB approaches, the proposed method advances seabed mapping for scientific and operational use. It effectively addresses the primary challenge in SfM-MVS-based bathymetry, namely its difficulty in environments with homogeneous visual textures, such as sandy or heavily seagrass-covered bottoms. Additionally, it provides essential, hard-to-obtain reference bathymetric data for SDB methods, through refraction-corrected SfM-MVS bathymetry. This eliminates the need for extensive manual fieldwork and costly external bathymetry data for training learning-based SDB models, expanding their applicability to reference-data-free areas and different modalities, such as Sentinel-2 Level 2A data. 

As a final remark, we would like to highlight that the proposed method is promising since it can significantly enhance coastal management and engineering, marine habitat mapping and monitoring, underwater archaeology and cultural heritage protection as well as navigation safety by providing high-precision and complete bathymetric data. This highlights the growing importance of deep learning in high resolution seabed mapping through photogrammetry and remote sensing.

\section*{Acknowledgment}
This work is part of MagicBathy: Multimodal multitAsk learninG for MultIsCale BATHYmetric mapping in shallow waters project funded by the European Union’s HORIZON Europe research and innovation programme under Marie Skłodowska-Curie Actions with Grant Agreement No. 101063294. Views and opinions expressed are, however, those of the authors only and do not necessarily reflect those of the European Union and the granting authority. Neither the European Union nor the granting authority can be held responsible for them. NVIDIA Corporation is acknowledged for supporting the initial experiments of this work through the NVIDIA Academic Hardware Grant Program. Prof. Dimitrios Skarlatos is acknowledged for the aerial data and GCP acquisition in Agia Napa. The authors thank Prof. Giorgos Tolias for the initial discussions on this topic.

\printcredits

\bibliographystyle{cas-model2-names}
\balance
\bibliography{cas-refs}



\end{document}